\documentclass[twoside,11pt]{article}

\usepackage{blindtext}

\usepackage{jmlr2e}

\usepackage{lastpage}
\jmlrheading{?}{2026}{1-\pageref{LastPage}}{6/2024; Revised 10/2026}{?}{24-0951}{Jia Li and Lin Lin}

\ShortHeadings{Wasserstein Canonical Variates}{Li and Lin}
\firstpageno{1}

\usepackage{amsmath,mathtools}
\DeclareMathOperator{\tr}{tr}
\usepackage{graphicx,psfrag,epsf}
\usepackage{enumerate}
\usepackage{enumitem}
\usepackage{natbib}
\usepackage{bbm}
\usepackage{caption}
\usepackage{subcaption}
\usepackage{booktabs}     %
\usepackage{placeins}
\usepackage{listings}

\usepackage{makecell}
\newsavebox{\otafnumcell}
\newsavebox{\otafnumwidth}
\newcolumntype{R}[1]{>{\begin{lrbox}{\otafnumcell}}c<{\end{lrbox}%
  \sbox{\otafnumwidth}{#1}\makebox[\wd\otafnumwidth][r]{\usebox{\otafnumcell}}}}
\usepackage{amssymb}
\usepackage{multirow}
\usepackage{xcolor}
\usepackage[scr=boondox]{mathalpha}

\usepackage[linesnumbered,ruled,vlined]{algorithm2e} 
\usepackage{algpseudocode}

\begin{document}

\def\spacingset#1{\renewcommand{\baselinestretch}%
{#1}\small\normalsize} \spacingset{1}

\newcommand{\revision}{\color{black}}
\newcommand{\revisionii}{\color{black}}

\title{Canonical Variates in Wasserstein Metric Space}

\author{\name Jia Li \email jiali@psu.edu \\
       \addr Department of Statistics\\
       The Pennsylvania State University\\
       University Park, PA 16802, USA
       \AND
       \name Lin Lin \email l.lin@duke.edu \\
       \addr Department of Biostatistics \& Bioinformatics \\
       Duke University \\
       Durham, NC 27710, USA}

\editor{Action Editor}

\maketitle

\begin{abstract}%
We introduce supervised canonical-variate analysis for distribution-valued observations, where each subject is represented by a probability distribution rather than a single feature vector. We propose the Optimal Transport And Fisher-ratio (OTAF) objective to learn a low-dimensional projection of the ambient feature space by maximizing a Fisher ratio of between-class to within-class variation based on Wasserstein distances. We consider both empirical measures with the 2-Wasserstein distance and fitted Gaussian mixture models (GMMs) with the Minimized Aggregated Wasserstein (MAW) distance. Using the envelope theorem, we derive the objective’s gradient and develop a safeguarded gradient-ascent method with a conditional first-order convergence guarantee. We also investigate spectral methods for finding promising starting points by constructing upper bounds on squared Gaussian Wasserstein distances that are quadratic in the projection matrix. We establish bounds relating distribution-estimation error to errors in the Fisher ratio and, under additional assumptions, its maximizing subspace. The analysis distinguishes these estimation effects from optimization error and differences between the Wasserstein and MAW objectives. Controlled simulations and three single-cell applications assess objective values, runtime, initialization sensitivity, and downstream classification performance. The applications demonstrate that OTAF can improve classification in fewer dimensions, with benefits depending on the representation, retained dimension, and downstream classifier.
\end{abstract}

\begin{keywords}
    Optimal transport, Wasserstein distance, Canonical variates, Dimension reduction, Distributional data
\end{keywords}

\section{Introduction}
\label{sec:intro}

In many biomedical studies, each subject is represented by a sample of measurements rather than a single feature vector. We call this unordered sample a \emph{data cloud}: each row is an observation, each column is a measured feature, and row order carries no information. The number of observations may also differ across subjects. Following our single-cell applications, we use the term \emph{cell} for an individual observation
within a cloud, even for observations that need not correspond to biological cells.
Flow cytometry, mass cytometry, and single-cell RNA sequencing, for example, measure many features on hundreds or thousands of cells per subject and are widely used in studies of immune response and disease. For subject-level prediction, the goal is to relate each subject's collection of measurements to an outcome, such as treatment response or disease risk. For example, \citet{Lin2015} analyze antigen-specific
T-cell responses using cytometry. We may use the observations directly as the
support of an empirical measure, with equal or unequal masses, or approximate
the underlying continuous distribution by a Gaussian mixture model (GMM).

We organize existing approaches to subject-level prediction into two broad strategies, according to how each subject is represented. Vector approaches reduce each data cloud to a fixed-length feature vector, which may be pre-specified or learned. 
Hierarchical mixture alignment was developed to match populations across
flow-cytometry samples~\citep{10.1371/journal.pcbi.1003130}, whereas scAlign learns a
shared representation for aligning single-cell data and identifying rare populations
\citep{Johansen2019}.  Population proportions and marker summaries can then feed a
standard classifier~\citep{doi:10.1126/scitranslmed.aax1880}; COMPASS is a prominent outcome-oriented example based on
antigen-specific T-cell subsets~\citep{Lin2015}, and scCURE combines response-cell
identification with outcome prediction~\citep{zou2023}.  
Alignment is not indispensable: permutation-invariant mixture moments and characteristic-function evaluations can encode full covariance and, for the latter, multimodal mixture structure. We have experimented with such vector baselines (Section~\ref{sec:leakage_clean_classification}).

Distributional approaches instead compare whole subject distributions.
Related single-cell methods learn cytometry distances or use optimal transport (OT) to improve cell--cell similarity~\citep{https://doi.org/10.1002/cyto.a.23792,
10.1093/bioinformatics/btac084}.
CytoEMD uses earth mover's distance to represent phenotype-related between-sample
variation \citep{10.1145/3535508.3545525}. PILOT builds patient-level OT
distances and trajectories \citep{https://doi.org/10.1038/s44320-023-00003-8}. QOT
constructs quantized OT sample distances for single-cell omics
\citep{10.1093/bib/bbae713}.  

The distributional methods motivate the central question of this work: \emph{How can a low-dimensional projection of the ambient feature space be learned to separate classes of subject-level distributions?}

We address that question by maximizing a pairwise Fisher ratio: the
average squared transport distance between selected subjects from different classes is
divided by the corresponding average within classes. This defines a supervised
ambient-space canonical-variate problem for distribution-valued observations. We
develop Optimal Transport And Fisher-ratio (OTAF) dimension reduction for both
empirical-measure and fitted-GMM representations. Its exact-gradient
iteration has a conditional first-order convergence guarantee.

Our problem differs from Wasserstein discriminant analysis (WDA), which
starts with vector observations and uses optimal transport between class-level
distributions to learn a discriminant metric~\citep{flamary2018wasserstein}.  Here, each
observation is already a distribution, and projected pairwise distribution distances
form the Fisher ratio.
Other related works have aimed to reduce the computational cost of the Wasserstein distance. For instance, the sliced-Wasserstein distance~\citep{bonneel2015sliced,tanguy2023properties} has been proposed based on linear slicing of probability distributions. Further extensions, such as the max-sliced-Wasserstein~\citep{deshpande2019max}, the multidimensional subspace-robust Wasserstein distance \citep{paty2019subspace}, and generalized sliced-Wasserstein~\citep{kolouri2019generalized}, have also been explored. These methods define or facilitate distribution comparison through projections, but their objectives differ from our supervised between-to-within Fisher ratio. Gromov--Wasserstein distance addresses the
different problem of comparing measures supported on different metric spaces
\citep{memoli2011gromov,delon2022gromov}.

Our contributions can be summarized as follows.
\begin{enumerate}[leftmargin=*,label=(\roman*),itemsep=1pt,topsep=3pt]
\item We introduce the supervised canonical-variate problem for observations that are
themselves distributions. The same framework accepts empirical distributions or fitted
GMMs as complementary finite representations; neither is treated as population truth.
\item 
We develop a principled gradient optimizer with an ascent safeguard and a conditional convergence guarantee. To reduce sensitivity to initialization, the optimizer screens several informed and random starting directions and attempts an initial refinement based on a generalized eigenvalue problem before proceeding with gradient ascent.
\item
  We establish bounds showing how errors in empirical and fitted-GMM representations affect the Fisher ratio and, under additional assumptions, the maximizing subspace. We distinguish these estimation effects from the gap between the objective value attained by the algorithm and the global maximum, and from the difference between the Wasserstein and MAW objectives. We also compare the computational costs of the two representations and identify conditions under which either may be advantageous.
\item 
Through controlled simulations and three single-cell applications, we evaluate objective values, runtime, and downstream classification performance, and investigate how these outcomes depend on initialization, sample size, ambient dimension, and distribution representation.
\end{enumerate}

Section~\ref{sec:preliminary} reviews optimal transport and MAW.
Section~\ref{sec:alg} defines the Fisher-ratio objective and presents the
optimization methods, complexity analysis, and theory.
Section~\ref{sec:exp} evaluates optimization, representation, and classification.
Section~\ref{sec:discuss} concludes.

\section{Preliminaries}
\label{sec:preliminary}

Consider two probability distributions (discrete or continuous) $\mathcal{P}_i$, $i=1, 2$, on the $d$-dimensional Euclidean space $\mathbb{R}^d$. For clarity of presentation, suppose that random vectors $X_i$ follow distribution $\mathcal{P}_i$, $i=1, 2$. A coupling between $\mathcal{P}_1$ and $\mathcal{P}_2$ is defined as a joint distribution of $(X_1, X_2)$ whose marginals are fixed at $\mathcal{P}_1$ and $\mathcal{P}_2$, respectively. 
A coupling is also commonly called a \emph{transport plan}; we use \emph{coupling} as the primary term below.
Denote the set of all such couplings by $\Pi$, with a particular coupling written as $\boldsymbol{\pi}$. When necessary, we stress the dependence of $\Pi$ on the marginal constraints by writing $\Pi_{\mathcal{P}_1, \mathcal{P}_2}$, or more simply $\Pi_{k_1, k_2}$ when $\mathcal{P}_{k_1}$ and $\mathcal{P}_{k_2}$ are under discussion. The $\mathtt{p}$-Wasserstein distance between $\mathcal{P}_1$ and $\mathcal{P}_2$ is defined through its $\mathtt{p}$th power as
\begin{eqnarray*}
W^{\mathtt{p}}(\mathcal{P}_1, \mathcal{P}_2):=\min_{\boldsymbol{\pi}\in \Pi} E_{\boldsymbol{\pi}}\|X_1-X_2\|^{\mathtt{p}} \; .
\end{eqnarray*}
In this work, we use the $2$-Wasserstein distance, denoted by $W_2$.
For brevity, we denote its square by $W^2$ rather than $W_2^2$. 

Let $\mathcal{Q}_k = \{(p^{(k)}_j, x^{(k)}_j), j=1, \dots, m_k\}$, $k=1, \dots, n$, denote a discrete distribution, where $m_k$ is the number of support points, $p^{(k)}_j$ is the probability assigned to support point $x^{(k)}_j \in \mathbb{R}^d$, and $\sum_j p^{(k)}_j = 1$. A coupling $\boldsymbol{\pi}$ is then characterized by joint probabilities $\pi_{ij}$ assigned to the Cartesian product $(x^{(k_1)}_i, x^{(k_2)}_j)$, subject to the marginal constraints $\sum_{j=1}^{m_{k_2}}\pi_{ij}=p^{(k_1)}_i$, $i=1,\ldots,m_{k_1}$, and
$\sum_{i=1}^{m_{k_1}}\pi_{ij}=p^{(k_2)}_j$, $j=1,\ldots,m_{k_2}$. Let $\Pi_{k_1, k_2}$ denote the set of all couplings for $\mathcal{Q}_{k_1}$ and $\mathcal{Q}_{k_2}$. The squared $2$-Wasserstein distance is then
\begin{eqnarray}
    W^2(\mathcal{Q}_{k_1}, \mathcal{Q}_{k_2}) =
\min_{\boldsymbol{\pi}\in \Pi_{k_1, k_2}}\sum_{i=1}^{m_{k_1}}\sum_{j=1}^{m_{k_2}}\pi_{ij}\|x^{(k_1)}_i-x^{(k_2)}_j\|^2\; .
\label{eq:Wdist}
\end{eqnarray}
From the perspective of optimization, this formulation is precisely the classical optimal transport (OT) problem~\citep{villani2003topics}.

For continuous distributions, closed-form Wasserstein distances are
generally unavailable except in special cases such as Gaussian distributions. Replacing
a distribution by a large empirical support makes exact transport increasingly costly.
Computational responses include accelerated earth mover's distance~\citep{5459199},
entropy-regularized transport~\citep{NIPS2013_af21d0c9}, and support
discretization~\citep{e25060839}. As an alternative finite representation,
\citet{chen2019aggregated} introduced the Minimized Aggregated Wasserstein (MAW)
distance for GMMs.

Consider two Gaussian distributions $\phi_{1}$ and $\phi_{2}$ in $\mathbb{R}^d$, parameterized by $(\mu_i, \mathbf{\Sigma}_i)$, $i=1,2$, where $\mu_i$ is the mean vector and $\mathbf{\Sigma}_i$ is the covariance matrix. Suppose $Z_1 \sim \phi_1$ and $Z_2 \sim \phi_2$. Let $\Pi_{\phi_1, \phi_2}$ be the set of couplings of $(Z_1, Z_2)$ with fixed marginals $\phi_1$ and $\phi_2$. Then
the squared Wasserstein distance $W^2(\phi_1,\phi_2)$ is given by the closed formula:
\begin{equation}
W^2(\phi_1, \phi_2) = \|\mathbf\mu_1 - \mathbf\mu_2\|^2 + \tr\left[
\mathbf\Sigma_1 + \mathbf\Sigma_2 -2\left(\mathbf\Sigma_1^{\frac 12}\mathbf
\Sigma_2\mathbf\Sigma_1^{\frac 12}\right)^{\frac 12}\right].
\label{eq:w_gaussian}
\end{equation}

Now consider GMMs $\mathcal{G}_k = \{(p_i^{(k)}, \phi_i^{(k)}), i=1, \dots, m_k\}$, $k=1, \dots, n$, where $m_k$ is the number of Gaussian components and $p_i^{(k)}$ is the prior weight of component $\phi_i^{(k)}$. Similar to the discrete case, let $\Pi_{k_1, k_2}$ be the set of couplings between the priors of $\mathcal{G}_{k_1}$ and $\mathcal{G}_{k_2}$. The squared MAW distance between $\mathcal{G}_{k_1}$ and $\mathcal{G}_{k_2}$ is defined as
\begin{equation*}
\widetilde{W}^2(\mathcal{G}_{k_1},\mathcal{G}_{k_2})=\min_{\boldsymbol{\pi} \in \Pi_{k_1,k_2}}
\sum_{i=1}^{m_{k_1}}\sum_{j=1}^{m_{k_2}} \pi_{ij} W^2(\phi^{(k_1)}_i,\phi^{(k_2)}_j). 
\end{equation*}

MAW treats a GMM as a discrete measure over Gaussian components, using
Gaussian $W_2$ costs in place of Euclidean costs between atoms (the support points of a
discrete measure); the remaining problem is a finite OT linear program (LP).
\citet{delon2020wasserstein} study the resulting
Wasserstein-type distance theoretically, \citet{delon2023optimal} use GMM transport for
multiscale texture synthesis, and \citet{https://doi.org/10.1111/biom.13630} use MAW
barycenters to integrate distributional observations. Thus the representation can
avoid transporting a large generated sample.
This reduction in transport support size begins after fitting; it
does not remove the statistical cost of estimating mixture parameters. In a favorable
well-separated spherical-mixture regime, \citet{kwon2020em} obtain a locally initialized
EM guarantee with sample size \(\widetilde O(Kd/\epsilon^2)\) under their stated separation
and accuracy conditions. For general full-covariance \(K\)-component mixtures,
\citet{ashtiani2018nearly} establish total-variation density-learning complexity
\(\widetilde\Theta(Kd^2/\epsilon^2)\), while their axis-aligned upper bound is
\(\widetilde O(Kd/\epsilon^2)\). Here \(\widetilde O\) denotes an asymptotic upper
bound and \(\widetilde\Theta\) denotes matching upper and lower orders; in both
notations, logarithmic factors are suppressed. Thus MAW can reduce the post-fit OT
support size, but
stable fitted-GMM inputs in high dimension still require a sample size appropriate to the
covariance model, separation, and target accuracy.

A discrete measure can be represented as a mixture of degenerate Gaussians with zero covariance. In this case, Gaussian squared Wasserstein costs reduce to squared Euclidean distances between support points, and MAW equals discrete $W_2$. This computational embedding does not imply that a discrete measure has a continuous GMM density. For nondegenerate mixtures, MAW is a metric on fitted mixture representations and need not equal the Wasserstein distance between the underlying populations or their empirical measures.

\section{Method}
\label{sec:alg}

\subsection{The OTAF Objective}
\label{sec:objective_hierarchy}

We use the following notations throughout.
\begin{enumerate}
\item
$d$: the original ambient feature dimension.
\item
$q$ ($q<d$): the retained dimension after projection.
\item
$n$: the number of subject-level distributions, or simply, the number of subjects. 
\item
$N_k$: the number of observed points (cells) for subject $k$, $k=1, ..., n$. 
\item
$m_k$: the number of empirical support points or fitted mixture components used by the relevant representation for subject $k$, $k=1, ..., n$. In the two types of distributions, $m_k$'s generally differ and are not being equated;
the common notation keeps the subsequent transport formulas applicable to either distribution representation.
\item 
$\mathrm{m_c}$: the number of different classes, e.g. $\mathrm{m_c}=2$ for binary classification.
\item
$A\in\mathbb{R}^{d\times q}$: a multi-direction projection.
\item
$a\in\mathbb{R}^d$: the projection $A$ in the special case $q=1$.
\item 
$I_q$: the $q\times q$ identity matrix.
\end{enumerate}

\subsubsection{Distribution Representations and Projections}

We distinguish the data-generating distribution from the two finite representations used by
OTAF. For subject \(k\), let \(\mathcal{P}_k\) denote the unknown population distribution and
let \(x_1^{(k)},\ldots,x_{N_k}^{(k)}\) be the observed sample. OTAF is applied
either to an empirical measure supported on \(m_k\leq N_k\) of these observations
(renumbered if necessary),
\[
\mathcal{Q}^{\mathrm{emp}}_k=\sum_{i=1}^{m_k}p_i^{(k)}\delta_{x_i^{(k)}},
\qquad \sum_{i=1}^{m_k}p_i^{(k)}=1,
\]
where \(\delta_{x_i^{(k)}}\) denotes the Dirac probability measure concentrated at
\(x_i^{(k)}\),
or to a fitted GMM
\[
\mathcal{G}^{\mathrm{fit}}_k=\sum_{i=1}^{m_k}p_i^{(k)}
\phi_i^{(k)},\qquad
\phi_i^{(k)}=N(\mu_i^{(k)},\mathbf{\Sigma}_i^{(k)}).
\]
In either representation, \(p_i^{(k)}\) denotes the nonnegative mass of the
\(i\)th support point or mixture component, with masses summing to one.
These weights belong to the chosen representation and need not be the same
for the empirical measure and fitted GMM.
The empirical support and the fitted GMM parameters are fixed inputs to the dimension-reduction
iteration; in particular, OTAF does not refit the GMM after changing the projection. We suppress
the superscripts ``emp'' and ``fit'' below when the representation is clear. The discrepancies
\(\mathcal{P}_k\to\mathcal{Q}^{\mathrm{emp}}_k\) and
\(\mathcal{P}_k\to\mathcal{G}^{\mathrm{fit}}_k\) are representation or estimation errors,
not errors produced by the OTAF optimization.

For a target
dimension \(q<d\), OTAF searches over orthonormal matrices
\[
\operatorname{St}(q,d)=\{A\in\mathbb R^{d\times q}:A^\top A=I_q\}.
\]
Only the span of \(A\) matters: \(A\) and \(AQ\), for any \(q\times q\) orthogonal
matrix \(Q\), represent the same subspace and give projected coordinates differing
only by an orthogonal rotation. The Grassmann manifold
\(\operatorname{Gr}(q,d)\) is the collection of all \(q\)-dimensional linear
subspaces of \(\mathbb R^d\), so it is the natural parameter space for OTAF
\citep{AbsilEtAl2008}. Appendix~\ref{ap:grassmann} defines the principal angles and
Grassmann distance used to compare two such subspaces. Projection sends a point \(x\)
to \(A^\top x\) and a Gaussian \(N(\mu,\Sigma)\) to
\(N(A^\top\mu,A^\top\Sigma A)\). More generally, let
\(T_A(x)=A^\top x\). For a probability distribution \(\mathcal Q\) on
\(\mathbb R^d\), we use the shorthand
\(A_\#\mathcal Q:=(T_A)_\#\mathcal Q\) for its \emph{pushforward} through this
projection (the symbol \(\#\) is not an index). Thus, for every measurable \(B\subseteq\mathbb R^q\),
\[
A_\#\mathcal Q(B)=\mathcal Q\{x:A^\top x\in B\}.
\]
Equivalently, if \(X\sim\mathcal Q\), then \(A_\#\mathcal Q\) is the distribution
of \(A^\top X\). Accordingly, for
\(\phi=N(\mu,\Sigma)\), we write
\[
P(\phi,A):=N(A^\top\mu,A^\top\Sigma A)
\]
for the Gaussian distribution obtained by projecting \(\phi\) through \(A\). For
\(q=1\), we write \(a\) for the unit
direction; \(a\) and \(-a\) represent the same projection.

\subsubsection{Objectives with Different Couplings}

Let \(Y_k\in\{1,\ldots,\mathrm{m_c}\}\) be the class label of subject \(k\), where $\mathrm{m_c}$ is the number of classes.
Let \(\mathcal I_B\) contain selected subject pairs from different classes and
\(\mathcal I_W\) selected pairs within the same class. These may be all eligible pairs
or subsets selected using training data only.

We next introduce notations for a feasible coupling set, an optimal coupling, a collection of pairwise
couplings, and the objectives evaluated with those couplings. For an indexed pair \((k_1,k_2)\), define the projected cost matrices
\[
\begin{split}
\big[C_{\mathsf{emp}}^{(k_1,k_2)}(A)\big]_{ij}
&=\left\|A^\top(x_i^{(k_1)}-x_j^{(k_2)})\right\|^2,\\
\big[C_{\mathsf{gmm}}^{(k_1,k_2)}(A)\big]_{ij}
&=W^2\!\left(P(\phi_i^{(k_1)},A),P(\phi_j^{(k_2)},A)\right).
\end{split}
\]
Here \(\Pi_{k_1,k_2}\) is the feasible transport polytope determined by the two vectors of
support or mixture weights. 

For compact
notation, index a selected
subject pair \((k_1,k_2)\) by \(\vartheta=(k_1,k_2)\), and let
\(s\in\{\mathsf{emp},\mathsf{gmm}\}\) identify the empirical and fitted-GMM representations,
respectively. Then
\(C_s^{(\vartheta)}(A):=C_s^{(k_1,k_2)}(A)\) and
\(\Pi_\vartheta:=\Pi_{k_1,k_2}\). The dependence of \(\Pi_\vartheta\) on the
chosen representation through its support counts and marginal weights is suppressed.
Denote a finite coupling
matrix by $\boldsymbol\lambda$, the discrete form of the coupling \(\boldsymbol\pi\) defined in Section~\ref{sec:preliminary}. 
For any feasible coupling
\(\boldsymbol\lambda^{(\vartheta)}\in\Pi_\vartheta\), define its fixed-coupling pair cost by
\begin{equation}
\begin{aligned}
\varphi_s^{(\vartheta)}(A,\boldsymbol\lambda^{(\vartheta)})
=\left\langle\boldsymbol\lambda^{(\vartheta)},C_s^{(\vartheta)}(A)\right\rangle
=\sum_{ij}\lambda_{ij}^{(\vartheta)}[C_s^{(\vartheta)}(A)]_{ij}.
\end{aligned}
\label{eq:pair-envelope}
\end{equation}
Here the brackets denote the Frobenius inner product of two matrices.
For GMM input, every entry of
\(C_{\mathsf{gmm}}^{(\vartheta)}(A)\) is the exact projected Gaussian
\(W^2\) cost. 
The minimum pairwise transport cost and one corresponding
optimal coupling are
\[
D_s^{(\vartheta)}(A)
=\min_{\boldsymbol\lambda^{(\vartheta)}\in\Pi_\vartheta}
\varphi_s^{(\vartheta)}(A,\boldsymbol\lambda^{(\vartheta)}),
\qquad
\boldsymbol\lambda_s^{(\vartheta),*}(A)
\in\underset{\boldsymbol\lambda^{(\vartheta)}\in\Pi_\vartheta}
{\operatorname{arg\,min}}\,
\varphi_s^{(\vartheta)}(A,\boldsymbol\lambda^{(\vartheta)}).
\]
Evaluating \(D_s^{(\vartheta)}(A)\) at a new projection therefore requires
solving the pairwise transport problem again.  
Denote the collection containing one selected optimal coupling for every indexed pair by
\[
\displaystyle \boldsymbol{\Gamma}^{\mathrm{opt}}_s(A)
=\left\{\boldsymbol{\lambda}^{(\vartheta),*}_s(A):
\vartheta \in\mathcal{I}_B\cup\mathcal{I}_W\right\}\, .
\]
Note that this
collection is different from the feasible set \(\Pi_{\vartheta}\) for a single pair.

The between- and within-class variations are defined as
\begin{align}
V^{\mathrm{opt}}_{B,s}(A)
&=\frac{1}{|\mathcal{I}_B|}
\sum_{\vartheta\in\mathcal{I}_B}D_s^{(\vartheta)}(A),
\label{eq:betweenvar}\\
V^{\mathrm{opt}}_{W,s}(A)
&=\frac{1}{|\mathcal{I}_W|}
\sum_{\vartheta\in\mathcal{I}_W}D_s^{(\vartheta)}(A).
\label{eq:withinvar}
\end{align}
Define the Fisher ratio
\begin{equation}
r^{\mathrm{opt}}_s(A)
=\frac{V^{\mathrm{opt}}_{B,s}(A)}
{V^{\mathrm{opt}}_{W,s}(A)}\, .
\label{eq:fisher}
\end{equation}
The OTAF canonical-variate problem is to solve
\begin{equation}
A_s^*\in\underset{A\in\operatorname{St}(q,d):
V^{\mathrm{opt}}_{W,s}(A)>0}{\operatorname{arg\,max}}
r_s^{\mathrm{opt}}(A).
\label{eq:obj}
\end{equation}
The terminology follows Fisher's criterion in linear discriminant analysis
(LDA)~\citep{hastie2009elements}. Our
criterion uses pairwise distances between subject distributions, so it does not require
a Wasserstein barycenter. In Euclidean data, variance is proportional to an average of
pairwise squared distances; this motivates Eq.~\eqref{eq:fisher} without making it
identical to multiclass LDA. The pairwise form also permits deliberate training-only
selection of confusing subject pairs.

Equation~\eqref{eq:obj} is the OTAF learning objective throughout this paper. The
fixed-coupling quantities introduced next do not define a competing learning criterion;
they are auxiliary functions used inside the optimizer when couplings computed at a
current projection are temporarily held fixed. For this purpose, for any fixed
admissible coupling collection
\[
\boldsymbol{\Gamma}
=\left\{\boldsymbol{\lambda}^{(\vartheta)}\in\Pi_{\vartheta}:
\vartheta\in\mathcal I_B\cup\mathcal I_W\right\},
\]
define
\begin{align}
V^{\mathrm{fix}}_{B,s}(A;\boldsymbol{\Gamma})
&=\frac{1}{|\mathcal{I}_B|}
\sum_{\vartheta\in\mathcal{I}_B} \varphi_s^{(\vartheta)}(A,\boldsymbol\lambda^{(\vartheta)}),\\
V^{\mathrm{fix}}_{W,s}(A;\boldsymbol{\Gamma})
&=\frac{1}{|\mathcal{I}_W|}
\sum_{\vartheta\in\mathcal{I}_W}
\varphi_s^{(\vartheta)}(A,\boldsymbol\lambda^{(\vartheta)}).
\end{align}
The corresponding auxiliary ratio is
\begin{equation}
r^{\mathrm{fix}}_s(A;\boldsymbol{\Gamma})
=\frac{V^{\mathrm{fix}}_{B,s}(A;\boldsymbol{\Gamma})}
{V^{\mathrm{fix}}_{W,s}(A;\boldsymbol{\Gamma})}.
\end{equation}

We call \(r_s^{\mathrm{fix}}(A;\boldsymbol\Gamma)\) the
\emph{fixed-coupling objective}. It is an auxiliary objective in which
the supplied collection of couplings \(\boldsymbol\Gamma\) is held fixed
while \(A\) varies.
By contrast, we call the true OTAF objective,
\(r_s^{\mathrm{opt}}(A)\), the \emph{transport-reoptimized objective},
emphasizing that every pairwise transport problem is solved anew at each
candidate projection \(A\). Correspondingly, we refer to
\(V^{\mathrm{opt}}_{B,s}(A)\) and \(V^{\mathrm{opt}}_{W,s}(A)\) as the
\emph{transport-reoptimized between- and within-class variations}, respectively.
The superscript ``opt'' refers to optimization of the pairwise couplings;
it does not mean that \(A\) globally maximizes the Fisher ratio.
Consequently,
\begin{eqnarray}
r^{\mathrm{opt}}_s(A)
=r^{\mathrm{fix}}_s\!\left(A;\boldsymbol{\Gamma}^{\mathrm{opt}}_s(A)\right),
\label{eq:pretend-fixed-point}
\end{eqnarray}
whereas \(r^{\mathrm{fix}}_s(A;\boldsymbol{\Gamma})\) with a coupling collection obtained at
another projection is a different function of \(A\).

When contrasting these objectives with quadratic approximations, ``exact''
refers to the stated Gaussian or empirical transport cost and to reoptimizing
the couplings, rather than retaining an approximate cost or a previously
computed coupling. Numerical evaluations still use finite-precision arithmetic
and transport-solver tolerances; the term does not mean symbolic computation.
An exact gradient additionally requires the differentiability conditions
stated in Section~\ref{sec:gradient_otaf}.

The fixed-coupling objective is useful for constructing and analyzing OTAF updates.
Let \(\tau\) denote the iteration index
and let \(A_\tau\) be the current projection at iteration \(\tau\). OTAF computes
\(\boldsymbol\Gamma_\tau=\boldsymbol\Gamma_s^{\mathrm{opt}}(A_\tau)\).
Under the regularity conditions in Section~\ref{sec:gradient_otaf}, Danskin's theorem~\citep{Danskin1967}
permits the derivative of the transport-reoptimized objective to be calculated by
differentiating the exact fixed-coupling costs at
\((A_\tau,\boldsymbol\Gamma_\tau)\). Later sections introduce optional update
mechanisms intended to broaden the search for a favorable basin; those mechanisms do
not change the objective definitions. In the remainder of
the paper, an unadorned \(r(A)\) denotes
\(r_{\mathsf{emp}}^{\mathrm{opt}}(A)\) for the discrete distributions and
\(r_{\mathsf{gmm}}^{\mathrm{opt}}(A)\) for the GMMs.

\subsubsection{Selection of Subject Pairs}
The index sets may contain all eligible pairs or a supervised subset. For each subject
\(k\), the implemented ``confusing-pair'' option first computes, in the original
ambient space, the following averages and their ratio:
\[
\xi_k^W=\operatorname{mean}_{\ell\ne k:Y_\ell=Y_k}D(k,\ell),
\qquad
\xi_k^B=\operatorname{mean}_{\ell:Y_\ell\ne Y_k}D(k,\ell),
\qquad
\gamma_k=\xi_k^B/\xi_k^W,
\]
where \(D\) is squared \(W_2\) for empirical inputs and squared MAW for fitted GMMs.
A small \(\gamma_k\) identifies a subject whose between-class separation is weak
relative to its within-class dispersion. The implementation selects
$n_{\mathrm{sel}}=\min\{n,\max(\lfloor0.30n\rfloor,4)\}$
subjects with the smallest ratios and calls their index set
\(\mathcal K_{\mathrm{sel}}\). It then uses the directed indexed pairs
\[
\mathcal I_B=\{(k,\ell):k\in\mathcal K_{\mathrm{sel}},\ Y_k\ne Y_\ell\},
\qquad
\mathcal I_W=\{(k,\ell):k\in\mathcal K_{\mathrm{sel}},\ k\ne\ell,\ Y_k=Y_\ell\}.
\]
If both endpoints are selected, both directed pairs are included. Thus their
undirected distance intentionally receives twice the weight of a pair with only
one selected endpoint, before normalization by the number of directed pairs.
The fraction and minimum count are user options; \(0.30\) and \(4\) are the defaults
used here. Because labels and intersubject distances enter this selection, it is
recomputed from the training subjects inside every validation or test split. The
all-pair option bypasses this selection and uses all eligible subject pairs.

\subsection{Exact-Gradient OTAF}
\label{sec:gradient_otaf}

\subsubsection{Challenges in Optimization}
Three features of the objective motivate the optimizer. First, the Fisher ratio
is nonconvex on the space of projection subspaces, so different initializations
can lead to different local optima. Second, the minimizing transport couplings
change with the projection; at ties with different active derivatives, the
objective can be nonsmooth. Third, nearly singular projected Gaussian covariances
can make derivative calculations ill-conditioned, while a small within-class
variation can amplify errors in the ratio. The gradient construction below
addresses local ascent where the envelope assumptions hold. Initialization
screening addresses sensitivity empirically; it does not remove nonconvexity.

The difficulty in maximizing the transport-reoptimized objective in
Eq.~\eqref{eq:obj} is that every projected pairwise distance is itself defined by an
optimal transport problem whose minimizing coupling can change with \(A\). 
For a fixed representation \(s\), use the following shorthand at iteration \(\tau\):
\[
\boldsymbol\lambda_{\tau}^{(\vartheta),*}
:=\boldsymbol\lambda_s^{(\vartheta),*}(A_\tau),
\qquad
\boldsymbol\Gamma_\tau
:=\boldsymbol\Gamma_s^{\mathrm{opt}}(A_\tau).
\]
For notational brevity in the gradient formulas, define
\begin{equation}
F_s(A):=r_s^{\mathrm{opt}}(A), \qquad
R_{s,\tau}(A):=r_s^{\mathrm{fix}}(A;\boldsymbol\Gamma_\tau)
=\frac{V_{B,s}^{\mathrm{fix}}(A;\boldsymbol\Gamma_\tau)}
{V_{W,s}^{\mathrm{fix}}(A;\boldsymbol\Gamma_\tau)},
\label{eq:exact-frozen-ratio}
\end{equation}
which are, respectively, the transport-reoptimized objective and the fixed-coupling objective formed from the couplings at iteration \(\tau\).
At \(A_\tau\), the collection \(\boldsymbol\Gamma_\tau\) is optimal, so
\begin{eqnarray}
F_s(A_\tau)=R_{s,\tau}(A_\tau) \, ,
\label{eq:pretend-fixed-point2}
\end{eqnarray}
which is essentially Eq.~(\ref{eq:pretend-fixed-point}). 

Eq.~(\ref{eq:pretend-fixed-point2}) naturally suggests an
alternating procedure: compute the optimal couplings at the current projection,
hold those couplings fixed while choosing a new projection by optimizing
\(R_{s,\tau}\), and then recompute the optimal couplings at the new projection. However, choosing a projection that maximizes the fixed-coupling objective
\(R_{s,\tau}\) does not guarantee that \(F_s\)
will increase after the couplings are recomputed. Thus, the equality at \(A_\tau\) does not by itself make this alternation a coordinate-ascent method or guarantee convergence to a fixed point. 
The reason is that the equality holds only at the current projection. At another projection
\(A\),
\(F_s(A)\) recomputes every pairwise coupling at \(A\), whereas
\(R_{s,\tau}(A)\) retains the couplings obtained at \(A_\tau\). Reoptimization cannot
increase an individual pairwise transport cost, so it can reduce both the between-class
and within-class variations. Because the numerator and denominator can decrease by
different relative amounts, \(F_s(A)\) may be either larger or smaller than
\(R_{s,\tau}(A)\).

Nevertheless, the fixed-coupling objective remains
useful for a different, local reason: under the regularity conditions below,
\(R_{s,\tau}\) and \(F_s\) have the same first-order derivative at \(A_\tau\).
Danskin's envelope theorem~\citep{Danskin1967} gives the precise identity.

For a differentiable scalar function \(h\), the notation
\(\mathbf{d} h(A)[H]\) means the differential of \(h\) at \(A\) applied to the
direction \(H\); equivalently, it is the directional derivative along a smooth
curve through \(A\) with tangent \(H\). For the Grassmann manifold, we represent a
tangent direction by a matrix \(H\) satisfying \(A^\top H=0\).

\begin{proposition}[Envelope identity]
\label{prop:envelope-gradient-rev3}
Suppose that each \(\Pi_{\vartheta}\) is compact and independent of \(A\), that
\(\varphi_s^{(\vartheta)}\) and its \(A\)-derivative are continuous near
\((A_\tau,\boldsymbol\lambda_\tau^{(\vartheta),*})\), and that the pairwise optimizer is locally unique or all
active optimizers have the same \(A\)-derivative. If
\(V_{W,s}^{\mathrm{opt}}(A_\tau)>0\), then for every Grassmann tangent direction \(H\),
\begin{equation}
\mathbf{d}D_s^{(\vartheta)}(A_\tau)[H]
=\mathbf{d}_A\varphi_s^{(\vartheta)}(A_\tau,\boldsymbol\lambda_\tau^{(\vartheta),*})[H],
\qquad
\operatorname{grad}F_s(A_\tau)
=\left.\operatorname{grad}_A R_{s,\tau}(A)\right|_{A=A_\tau}.
\label{eq:envelope-gradient-rev3}
\end{equation}
Here \(\operatorname{grad}\) denotes the Riemannian gradient on
\(\operatorname{Gr}(q,d)\), also called the Grassmann gradient. The subscript \(A\)
emphasizes that the projection is differentiated while the optimal coupling found at
\(A_\tau\) is held fixed.
\end{proposition}

Appendix~\ref{ap:envelope-gradient} proves
Proposition~\ref{prop:envelope-gradient-rev3} by stating the minimum-form version
of Danskin's theorem and verifying its assumptions for the OTAF
transport problems. It also derives the gradient formulas below.

\subsubsection{Computing the Exact Gradient}
The gradient can be formed directly after the pairwise optimal couplings
\(\boldsymbol\Gamma_\tau\) have been computed. For an empirical cost-matrix entry,
write \(z=x_i^{(k_1)}-x_j^{(k_2)}\); then
\[
\nabla_A\|A^\top z\|^2=2zz^\top A.
\]
For entry \((i,j)\) of the GMM cost matrix for pair
\(\vartheta=(k_1,k_2)\) (that is, $C_{\mathsf{gmm}}^{(\vartheta)}(A)$), two Gaussian components are involved: 
\(\phi_i^{(k_1)}=N(\mu_i^{(k_1)},\Sigma_i^{(k_1)})\) and 
\(\phi_j^{(k_2)}=N(\mu_j^{(k_2)},\Sigma_j^{(k_2)})\). We denote the reversed pair by $\vartheta^{r}=(k_2,k_1)$.
Let $(k_1, i)$ and $(k_2, j)$
label the two
Gaussians in this component pair. Set
\(S_{k_1, i,\tau}=A_\tau^\top\Sigma_i^{(k_1)} A_\tau\), and likewise define $S_{k_2, j,\tau}$. When
\(S_{k_1,i,\tau}\) and \(S_{k_2,j, \tau}\) are positive definite, define
\[
T_{i,j,\tau}^{(\vartheta)}=S_{k_1,i,\tau}^{-1/2}
(S_{k_1,i,\tau}^{1/2}S_{k_2,j,\tau}S_{k_1,i,\tau}^{1/2})^{1/2}
S_{k_1,i,\tau}^{-1/2},
\qquad T_{j,i,\tau}^{(\vartheta^r)}=\bigl (T_{i,j,\tau}^{(\vartheta)}\bigr )^{-1}.
\]
With \(\Delta\mu_{ij}^{(\vartheta)}=\mu_i^{(k_1)}-\mu_j^{(k_2)}\), the gradient of their exact projected
Gaussian \(W^2\) cost at the current projection is
\begin{equation}
\begin{aligned}
& \left.
\nabla_A W^2\!\left(
P(\phi_i^{(k_1)},A),P(\phi_j^{(k_2)},A)\right)
\right|_{A=A_\tau} \\
={}&2
\Delta\mu_{ij}^{(\vartheta)}
\bigl(\Delta\mu_{ij}^{(\vartheta)}\bigr)^\top A_\tau
+2\Sigma_i^{(k_1)}A_\tau(I_q-T_{i,j,\tau}^{(\vartheta)})
+2\Sigma_j^{(k_2)}A_\tau(I_q-T_{j,i,\tau}^{(\vartheta^r)}).
\end{aligned}
\label{eq:exact-component-gradient-rev3}
\end{equation}
Equation~\eqref{eq:exact-component-gradient-rev3} is precisely
\(\left.\nabla_A[C_{\mathsf{gmm}}^{(\vartheta)}(A)]_{ij}\right|_{A=A_\tau}\). For the empirical
representation, with
\(z_{ij}^{(\vartheta)}=x_i^{(k_1)}-x_j^{(k_2)}\),
\[ \left .
\nabla_A[C_{\mathsf{emp}}^{(\vartheta)}(A)]_{ij}\right|_{A=A_\tau} 
=2z_{ij}^{(\vartheta)}(z_{ij}^{(\vartheta)})^\top A_\tau,
\]
as shown above.
Thus, for each selected subject pair \(\vartheta\), for $s=\mathsf{emp}, \mathsf{gmm}$, form
\[
J_{s,\tau}^{(\vartheta)}
=\sum_{ij}\lambda_{\tau,ij}^{(\vartheta),*}
\left.\nabla_A[C_s^{(\vartheta)}(A)]_{ij}\right|_{A=A_\tau}.
\]
The two variation gradients are then
\[
\nabla_A V_{B,s}^{\mathrm{fix}}(A_\tau;\boldsymbol\Gamma_\tau)
=\frac{1}{|\mathcal I_B|}\sum_{\vartheta\in\mathcal I_B}J_{s,\tau}^{(\vartheta)},
\qquad
\nabla_A V_{W,s}^{\mathrm{fix}}(A_\tau;\boldsymbol\Gamma_\tau)
=\frac{1}{|\mathcal I_W|}\sum_{\vartheta\in\mathcal I_W}J_{s,\tau}^{(\vartheta)}.
\]
Apply the quotient rule to obtain the exact Euclidean gradient
\begin{equation}
\begin{aligned}
Z_\tau
&=\frac{1}{\bigl(V_{W,s}^{\mathrm{opt}}(A_\tau)\bigr)^2}
\Big [
V_{W,s}^{\mathrm{opt}}(A_\tau)
\nabla_A V_{B,s}^{\mathrm{fix}}(A_\tau;\boldsymbol\Gamma_\tau)
-V_{B,s}^{\mathrm{opt}}(A_\tau)
\nabla_A V_{W,s}^{\mathrm{fix}}(A_\tau;\boldsymbol\Gamma_\tau)
\Big],\\
G_\tau
&=(I_d-A_\tau A_\tau^\top)Z_\tau
=\operatorname{grad}F_s(A_\tau).
\end{aligned}
\label{eq:exact-grassmann-gradient}
\end{equation}
Here \(Z_\tau=\nabla_A F_s(A_\tau)\) is the ordinary Euclidean gradient when
\(A\) is viewed as an unconstrained \(d\times q\) matrix, whereas
\(G_\tau=\operatorname{grad}F_s(A_\tau)\) is the Riemannian gradient on
\(\operatorname{Gr}(q,d)\), also called the Grassmann gradient.
Appendix~\ref{ap:envelope-gradient} derives the tangent-space projection.
No derivative of the optimal coupling is required.
Proposition~\ref{prop:envelope-gradient-rev3} also implies that
a smooth local maximizer of \(F_s\) is
stationary for its own fixed-coupling objective, although it need not globally
maximize that fixed-coupling objective.

\subsubsection{Exact-Gradient Iteration}
At outer iteration \(\tau\), the method performs the following steps.
\begin{enumerate}
\item \textbf{Update the couplings:}
At the current projection \(A_\tau\), solve all projected transport problems and
evaluate the transport-reoptimized objective \(F_s(A_\tau)\). These optimal
couplings are held fixed only while computing the gradient in the next step.

\item \textbf{Compute the gradient:}
Form the Grassmann gradient \(G_\tau\) using
Eq.~\eqref{eq:exact-grassmann-gradient}. If
\(\|G_\tau\|_F\leq\epsilon_{\mathrm{grad}}\), stop; otherwise, continue to the
line search. Here \(\|\cdot\|_F\) denotes the Frobenius norm,
\(\|G\|_F=\{\operatorname{tr}(G^\top G)\}^{1/2}\). The implementation uses \(\epsilon_{\mathrm{grad}}=10^{-7}\).

\item \textbf{Construct a trial projection:}
Set
\(
\alpha_0=\min\{1,\|G_\tau\|_F^{-1}\}
\)
and consider successively halved step sizes
$\alpha_j=2^{-j}\alpha_0$, $j=0,1,\ldots$.
For each \(\alpha_j\), form
\[
A_\tau(\alpha_j)=\operatorname{qf}(A_\tau+\alpha_jG_\tau),
\]
where \(\operatorname{qf}(M)\) denotes the orthonormal \(Q\)-factor in the thin
QR factorization of \(M\). This QR retraction converts the trial step into an
orthonormal representative of a candidate projection subspace.

\item \textbf{Test the trial projection:}
Recompute all optimal transport couplings at \(A_\tau(\alpha_j)\) and the
corresponding between- and within-class variations. A trial is admissible only
if \(V_{W,s}^{\mathrm{opt}}\{A_\tau(\alpha_j)\}>0\); a zero denominator makes
the Fisher ratio undefined, so such a trial cannot be accepted.
For an admissible trial, evaluate its exact transport-reoptimized objective.
Accept the first step size satisfying
\begin{equation}
F_s\{A_\tau(\alpha_j)\}
\geq F_s(A_\tau)+\omega\alpha_j\|G_\tau\|_F^2.
\label{eq:gradient-armijo}
\end{equation}
Here \(\omega\in(0,1)\) is the Armijo sufficient-increase parameter; the
implementation uses \(\omega=10^{-4}\). If the condition fails, halve the step
size and repeat this step. Equation~\eqref{eq:gradient-armijo} states the ideal
test. To tolerate numerical roundoff, the finite implementation subtracts
\(
10^{-10}\max\{1,|F_s(A_\tau)|,|F_s(A_\tau(\alpha_j))|\}
\)
from its right-hand side. It permits at most 50 trials
and reports a numerical failure if none passes this test.

\item \textbf{Update the projection:}
Denote the first accepted step size at iteration \(\tau\) by
\(\alpha_\tau\), and set
\(A_{\tau+1}=A_\tau(\alpha_\tau)\).
Then begin the next outer iteration.
Thus, the couplings used to form \(G_\tau\)
are not used to decide whether the trial projection is accepted; acceptance is
based on newly optimized couplings and the corresponding exact objective.
\end{enumerate}

The outer iteration continues until the gradient tolerance or the prescribed
iteration cap is reached. The implementation stores each visited projection
\(A_\tau\) and its exact transport-reoptimized objective value \(F_s(A_\tau)\).
Upon termination, it returns the visited projection having the largest stored
objective value. The ideal Armijo test gives nondecreasing accepted values;
the implemented tolerance can permit decreases of numerical size. The stored
best value is nondecreasing in either case. 

\begin{theorem}[Monotonicity and first-order convergence]
\label{thm:gradient-convergence-rev3}
\mbox{}\par\noindent
Consider the exact-gradient iteration that continues whenever its Grassmann
gradient is nonzero and does not impose the practical outer-iteration or backtracking
caps. As in Steps 3 and 4 above, use the QR retraction and evaluate the exact
transport-reoptimized objective in the Armijo test
Eq.~\eqref{eq:gradient-armijo}. Define the initial superlevel set
\[
\mathcal L_0
=\{A\in\operatorname{Gr}(q,d):F_s(A)\geq F_s(A_0)\}.
\]
Assume that \(\mathcal L_0\) is compact and has an open neighborhood
\(\mathcal U\) on which the following conditions hold:
\begin{enumerate}[label=(\roman*),leftmargin=2.5em]
\item \(V_{W,s}^{\mathrm{opt}}(A)\geq\nu>0\);
\item the conditions of Proposition~\ref{prop:envelope-gradient-rev3} hold for
every selected transport problem;
\item projected GMM component covariances, when applicable, are uniformly positive
definite; and
\item \(F_s\) has a Lipschitz-continuous Riemannian gradient.
\end{enumerate}
Then backtracking terminates after finitely many trials at every nonstationary iterate,
\[
F_s(A_{\tau+1})-F_s(A_\tau)
\geq\omega\alpha_\tau\|\operatorname{grad}F_s(A_\tau)\|_F^2>0,
\]
the objective values converge, and
\[
\|\operatorname{grad}F_s(A_\tau)\|_F\longrightarrow0.
\]
Every accumulation point is therefore first-order stationary for the
transport-reoptimized objective.
\end{theorem}

Appendix~\ref{ap:envelope-gradient} provides the proof. The conditions exclude
specific failures of smooth gradient analysis, but differ in how restrictive they
are for OTAF. Condition (i) keeps the average projected within-class variation
bounded away from zero: a zero denominator makes the Fisher ratio undefined,
while a near-zero denominator can make the ratio and its derivatives unstable.
This condition excludes projections that eliminate all within-class differences,
even if they preserve between-class separation.

Condition (ii) makes the derivative of each optimized transport cost unambiguous.
The compactness and projection-independence requirements in
Proposition~\ref{prop:envelope-gradient-rev3} hold automatically for fixed finite
marginal weights. The ground costs are smooth for empirical measures and, under
condition (iii), for GMMs. The substantive requirement concerns optimal couplings: if two
equally good couplings give different directional derivatives along the same
feasible projection direction, the optimized pairwise cost is nonsmooth and has
no single gradient.
Multiple optimal couplings are nevertheless allowed when their derivatives agree.
Such ties can occur naturally as the optimal matching changes with the projection,
so the excluded cases are not necessarily pathological data configurations.

Condition (iii) excludes singular projected Gaussian covariances. At singularity,
a component has zero variance in some projected direction, and the inverse
matrices in the Gaussian derivative formula cease to exist, although the
Wasserstein distance itself remains defined. Such singularity can reflect
legitimate lower-dimensional data. This condition is satisfied whenever
all supplied ambient component covariances are positive definite: orthonormal
projection preserves positive definiteness, and finitely many components provide
a common positive lower bound on directional variance. 

Condition (iv) controls how rapidly the gradient changes with the projection,
ensuring a uniformly accurate first-order prediction for the backtracking
argument. It excludes abrupt gradient changes or unbounded curvature. On compact
regions inside a neighborhood where every selected transport problem has a
locally unique optimal coupling, the within-class variation is positive, and the
GMM covariances are positive definite when applicable, the objective is smooth
with bounded second derivatives. This gives a sufficient setting for the
Lipschitz-gradient condition.

In summary, several requirements follow from the formulation or positive-definite
covariance inputs, whereas the denominator and transport-smoothness requirements
remain genuine restrictions of
the convergence result. 

Theorem~\ref{thm:gradient-convergence-rev3} establishes first-order convergence
of the idealized exact-gradient iteration under the stated regularity conditions.
The implementation allows tolerance-level objective decreases and, for GMM
derivatives, uses a small covariance ridge near singularity
(Appendix~\ref{ap:envelope-gradient}). These numerical safeguards address
finite precision and ill-conditioning, rather than simplify the optimization
problem, and are not covered by the theorem.
The implementation returns the best evaluated projection and its gradient norm.

\subsection{Fixed-Coupling Spectral Proposals}
\label{sec:spectral_proposals}

We also consider optional \emph{spectral proposals}: candidate projection subspaces obtained from generalized eigenvectors. In particular, we introduce the independent-coupling proposal (Eq.~\eqref{eq:W2P}) and the geometric-coupling proposal (Eq.~\eqref{eq:geometric-bound-cost}), each derived from an upper bound on the squared Wasserstein distance between Gaussian distributions. These bounds are quadratic in the projection matrix. We further examine a \emph{double-bound update} that backtracks along each proposal separately and selects the successful trial with the larger transport-reoptimized ratio.

Relative to maximizing the transport-reoptimized objective, these proposals involve one level of approximation for empirical representations and an additional level for GMM representations. Both replace the transport-reoptimized objective with the fixed-coupling objective when constructing a candidate projection. The motivation is to obtain a computationally convenient candidate that may lead to a better basin before gradient refinement. The spectral proposals therefore serve as initialization strategies, while the gradient-ascent iteration in Section~\ref{sec:gradient_otaf} remains the main optimization method. Section~\ref{sec:algorithm_experiments} compares these spectral proposals with gradient ascent.

First, at the current projection \(A_\tau\), we compute
\(\boldsymbol\Gamma_\tau
=\boldsymbol\Gamma_s^{\mathrm{opt}}(A_\tau)\)
and consider maximizing the fixed-coupling objective
\(R_{s,\tau}(A)=r_s^{\mathrm{fix}}(A;\boldsymbol\Gamma_\tau)\)
instead of the transport-reoptimized objective \(F_s(A)\).
This substitution has a local mathematical motivation:
the two objectives agree at \(A_\tau\) and, under the conditions of
Proposition~\ref{prop:envelope-gradient-rev3}, have the same first-order
derivative there. However, they can differ substantially away from
\(A_\tau\). Thus, maximizing \(R_{s,\tau}\) provides only a candidate
projection, not a guaranteed ascent step for \(F_s\).

\subsubsection{Fixed-Coupling Objective for Empirical Distributions}
For empirical measures, the fixed-coupling costs are quadratic in \(A\).
For a fixed collection $\boldsymbol{\Gamma}
=\left\{\boldsymbol{\lambda}^{(\vartheta)}\in\Pi_{\vartheta}:
\vartheta\in\mathcal I_B\cup\mathcal I_W\right\}$ of support-point couplings,
define the between- and within-class scatter matrices by
\begin{equation}
\begin{aligned}
C_{\iota,\mathsf{emp}}(\boldsymbol\Gamma)
=\frac{1}{|\mathcal I_\iota|}
\sum_{\vartheta=(k_1,k_2)\in\mathcal I_\iota}
\sum_{i=1}^{m_{k_1}}\sum_{j=1}^{m_{k_2}}
\lambda_{ij}^{(\vartheta)}
\bigl(x_i^{(k_1)}-x_j^{(k_2)}\bigr)
\bigl(x_i^{(k_1)}-x_j^{(k_2)}\bigr)^\top,
\; \iota\in\{\mathrm{B},\mathrm{W}\}.
\end{aligned}
\label{eq:empirical-spectral-scatter}
\end{equation}
Here \(\mathrm{B}\) denotes between-class and \(\mathrm{W}\) within-class.
These are positive-semidefinite \(d\times d\) matrices, distinct from the
pairwise projected cost matrices \(C_{\mathsf{emp}}^{(\vartheta)}(A)\)
in Section~\ref{sec:objective_hierarchy}.
The identity
\[
\bigl\|A^\top(x_i^{(k_1)}-x_j^{(k_2)})\bigr\|^2
=\operatorname{tr}\!\left\{
A^\top\bigl(x_i^{(k_1)}-x_j^{(k_2)}\bigr)
\bigl(x_i^{(k_1)}-x_j^{(k_2)}\bigr)^\top A
\right\}
\]
therefore gives the exact fixed-coupling variations
\begin{equation}
V_{\iota,\mathsf{emp}}^{\mathrm{fix}}(A;\boldsymbol\Gamma)
=\operatorname{tr}\!\left\{
A^\top C_{\iota,\mathsf{emp}}(\boldsymbol\Gamma)A\right\},
\qquad \iota\in\{\mathrm{B},\mathrm{W}\}.
\label{eq:empirical-spectral-variation}
\end{equation}
Appendix~\ref{ap1} details the aggregation. For \(q=1\), the resulting
fixed-coupling objective is the generalized Rayleigh quotient
\begin{equation}
r_{\mathsf{emp}}^{\mathrm{fix}}(a;\boldsymbol\Gamma)
=\frac{a^\top C_{B,\mathsf{emp}}(\boldsymbol\Gamma)a}
{a^\top C_{W,\mathsf{emp}}(\boldsymbol\Gamma)a},
\qquad \|a\|=1.
\label{eq:empirical-spectral-rayleigh}
\end{equation}
If \(C_{W,\mathsf{emp}}(\boldsymbol\Gamma)\) is positive definite, this
quotient is maximized by a leading generalized eigenvector, normalized
to unit Euclidean norm:
\[
C_{B,\mathsf{emp}}(\boldsymbol\Gamma)a
=\beta_{\max}C_{W,\mathsf{emp}}(\boldsymbol\Gamma)a,
\]
where \(\beta_{\max}\) is the largest generalized eigenvalue and equals
the maximum fixed-coupling ratio. Taking
\(\boldsymbol\Gamma=\boldsymbol\Gamma_\tau\) yields a candidate from
\(R_{\mathsf{emp},\tau}\), not a guaranteed maximizer or ascent step for
\(F_{\mathsf{emp}}\). For \(q>1\), the fixed-coupling objective is a ratio
of traces; using the leading generalized eigenvectors then gives only a
candidate subspace. Section~\ref{sec:geometric_proposal} describes its
orthonormalization, diagonal loading, and exact-objective screening.

\subsubsection{Two Quadratic Gaussian-Coupling Bounds}
\label{sec:alg_gmm}

For GMMs, the exact fixed-coupling costs are nonquadratic in \(A\).
Replacing them by quadratic upper bounds yields a generalized eigenvalue problem for one
direction. These cost bounds, however, do not imply a bound on the
Fisher ratio. Since
neither fixing the couplings nor approximating the Gaussian costs
guarantees ascent in \(F_s\), we safeguard candidate moves
through backtracking, reoptimizing all pairwise transports at each trial
and determining acceptance from \(F_s\).
The practical benefit of these optional spectral proposals is assessed
empirically.

Consider two Gaussian components drawn from a pair of subject
distributions indexed by \(\vartheta=(k_1,k_2)\).
For the componentwise discussion, write these components as
\(\phi_1=N(\mu_1,\Sigma_1)\) and
\(\phi_2=N(\mu_2,\Sigma_2)\), suppressing their subject and
component indices for brevity. We restore these indices when
aggregating the component costs below.
Let
\(\Delta\mu=\mu_1-\mu_2\). Their exact projected squared distance is
\begin{align}
&W^2\!\left(P(\phi_1,A),P(\phi_2,A)\right) 
\nonumber\\
&=\|A^\top\Delta\mu\|^2+
\operatorname{tr}\!\left[
S_1+S_2-2\left(S_1^{1/2}S_2S_1^{1/2}\right)^{1/2}\right],
\quad S_\ell=A^\top\Sigma_\ell A.
\label{eq:w2phi}
\end{align}
The covariance term depends nonlinearly on \(A\). Thus, even after the mixture-component
couplings are fixed, directly maximizing the exact fixed-coupling Fisher ratio is not a
generalized eigenvalue problem. We seek quadratic component costs that upper-bound
Eq.~\eqref{eq:w2phi}; after aggregation, either choice yields a generalized Rayleigh
quotient.

Both constructions use fixed joint Gaussian couplings. Let \(U_{12}\)
be a cross-covariance for which
\[
\begin{pmatrix}
\Sigma_1&U_{12}\\ U_{12}^\top&\Sigma_2
\end{pmatrix}\succeq0.
\]
This block matrix specifies a joint Gaussian vector \((Z_1,Z_2)\) with the required
marginals. Projecting that same coupling gives the quadratic expected cost
\begin{equation}
c_{U_{12}}(A)
=\|A^\top\Delta\mu\|^2+
\operatorname{tr}\!\left[A^\top
\{\Sigma_1+\Sigma_2-U_{12}-U_{12}^\top\}A\right].
\label{eq:fixed-gaussian-coupling-cost}
\end{equation}
Because the squared Wasserstein distance is the minimum expected squared cost over all
couplings, every feasible \(U_{12}\) makes Eq.~\eqref{eq:fixed-gaussian-coupling-cost}
an upper bound on the exact projected cost.

\paragraph{Independent coupling:}
Taking \(U_{12}^{\mathrm{ind}}=0\) couples \(Z_1\) and \(Z_2\) independently and gives
\begin{align}
\widehat W_{\mathrm{ind}}^2(\phi_1,\phi_2)
&=\|\Delta\mu\|^2+\operatorname{tr}(\Sigma_1+\Sigma_2),
\label{eq:W2}\\
c_{12}^{\mathrm{ind}}(A)
&=\widehat W_{\mathrm{ind}}^2\!\left(P(\phi_1,A),P(\phi_2,A)\right)\nonumber\\
&=\operatorname{tr}\!\left[A^\top
\{\Delta\mu\Delta\mu^\top+\Sigma_1+\Sigma_2\}A\right].
\label{eq:W2P}
\end{align}
Appendix~\ref{ap2} derives both formulas as the independent special case of a
common fixed-Gaussian-coupling identity.
This is the simplest quadratic bound, but it discards all cross-covariance between the
two Gaussian random vectors.

\paragraph{Geometric coupling:}
For \(\Sigma_1,\Sigma_2\succ0\), define
\begin{equation}
G_{12}=\Sigma_1^{1/2}
\left(\Sigma_1^{-1/2}\Sigma_2\Sigma_1^{-1/2}\right)^{1/2}
\Sigma_1^{1/2}.
\label{eq:geometric-cross-covariance}
\end{equation}
Here \(G_{12}\) is the matrix geometric mean of \(\Sigma_1\) and \(\Sigma_2\),
extending the usual geometric mean of two positive scalars. We call the resulting
construction a \emph{geometric-mean coupling}, shortened to \emph{geometric coupling}.
This matrix is positive definite and satisfies
\(G_{12}\Sigma_1^{-1}G_{12}=\Sigma_2\). Consequently,
\[
\begin{pmatrix}\Sigma_1&G_{12}\\G_{12}&\Sigma_2\end{pmatrix}
\]
is positive semidefinite and singular and therefore defines a valid degenerate joint
Gaussian coupling. Appendix~\ref{ap1} proves these claims and gives an explicit joint
Gaussian construction. Substituting \(U_{12}=G_{12}\) in
Eq.~\eqref{eq:fixed-gaussian-coupling-cost} gives
\begin{equation}
c_{12}^{\mathrm{geo}}(A)=\|A^\top\Delta\mu\|^2
+\operatorname{tr}\!\left[A^\top
\{\Sigma_1+\Sigma_2-2G_{12}\}A\right].
\label{eq:geometric-bound-cost}
\end{equation}

The geometric coupling yields the tighter of the two upper bounds:
\begin{equation}
\begin{aligned}
W^2\!\left(P(\phi_1,A),P(\phi_2,A)\right)
&\leq c_{12}^{\mathrm{geo}}(A)
\leq c_{12}^{\mathrm{ind}}(A),\\
c_{12}^{\mathrm{ind}}(A)-c_{12}^{\mathrm{geo}}(A)
&=2\operatorname{tr}(A^\top G_{12}A)\geq0.
\end{aligned}
\label{eq:geometric-bound-ordering}
\end{equation}
The first inequality follows from feasibility of the projected geometric
coupling. The second line follows by subtracting
Eq.~\eqref{eq:geometric-bound-cost} from Eq.~\eqref{eq:W2P}
and using \(G_{12}\succeq0\).
Appendix~\ref{ap1} derives these cost formulas and proves the
feasibility and positive definiteness claims.
This ordering is preserved when component costs are aggregated
with the same nonnegative mixture-component coupling weights.

We now restore the subject-pair and component indices.
For \(\vartheta=(k_1,k_2)\), let
\(c_{ij}^{(\vartheta,\eta)}(A)\),
\(\eta\in\{\mathrm{geo},\mathrm{ind}\}\), denote the cost obtained
from \(c_{12}^{\eta}(A)\) by substituting
\[
(\mu_1,\Sigma_1)=(\mu_i^{(k_1)},\Sigma_i^{(k_1)}),
\qquad
(\mu_2,\Sigma_2)=(\mu_j^{(k_2)},\Sigma_j^{(k_2)}).
\]
For any fixed admissible mixture-component coupling
\(\boldsymbol\lambda^{(\vartheta)}\in\Pi_{k_1,k_2}\), define
\[
D_{\mathsf{gmm}}^{(\vartheta),\mathrm{fix}}
(A;\boldsymbol\lambda^{(\vartheta)})
=\sum_{ij}\lambda_{ij}^{(\vartheta)}
W^2\!\left(P(\phi_i^{(k_1)},A),P(\phi_j^{(k_2)},A)\right),
\]
and
\[
\widetilde D_{\eta}^{(\vartheta),\mathrm{fix}}
(A;\boldsymbol\lambda^{(\vartheta)})
=\sum_{ij}\lambda_{ij}^{(\vartheta)}
c_{ij}^{(\vartheta,\eta)}(A).
\]
Then
\begin{equation}
D_{\mathsf{gmm}}^{(\vartheta)}(A)
\leq D_{\mathsf{gmm}}^{(\vartheta),\mathrm{fix}}
(A;\boldsymbol\lambda^{(\vartheta)})
\leq\widetilde D_{\mathrm{geo}}^{(\vartheta),\mathrm{fix}}
(A;\boldsymbol\lambda^{(\vartheta)})
\leq\widetilde D_{\mathrm{ind}}^{(\vartheta),\mathrm{fix}}
(A;\boldsymbol\lambda^{(\vartheta)}).
\label{eq:maw-spectral-bound-ordering}
\end{equation}
The first inequality reflects reoptimization of the mixture-component coupling in MAW; the
remaining inequalities follow componentwise from
Eq.~\eqref{eq:geometric-bound-ordering}. Importantly, these cost inequalities do
\emph{not} order the resulting Fisher ratios, because both the between-class numerator
and the within-class denominator are replaced.

The geometric bound is exact when the two component covariances
are positive scalar multiples of a common covariance matrix.
Specifically, if
\(\Sigma_1=s_1\Sigma_0\) and \(\Sigma_2=s_2\Sigma_0\),
where \(s_1,s_2>0\) and \(\Sigma_0\succ0\), then
\(G_{12}=\sqrt{s_1s_2}\,\Sigma_0\), and
Eq.~\eqref{eq:geometric-bound-cost} equals the exact projected
Gaussian \(W^2\) cost for every orthonormal projection matrix \(A\).
Appendix~\ref{ap1} gives the proof.
Identical covariances correspond to \(s_1=s_2\).
This exactness and the tighter-bound ordering motivate considering
the geometric construction for spectral proposals.

\subsubsection{Generalized-Eigenvector Construction for GMM}
\label{sec:geometric_proposal}

For \(\eta\in\{\mathrm{ind},\mathrm{geo}\}\), define the matrix for each component pair $(i, j)$ from a pair of subjects indexed by $\vartheta=(k_1, k_2)$:
\begin{equation}
H_{ij}^{(\vartheta,\eta)}
=\Delta\mu_{ij}^{(\vartheta)}\bigl(\Delta\mu_{ij}^{(\vartheta)}\bigr)^\top
+\Sigma_i^{(k_1)}+\Sigma_j^{(k_2)}
-U_{ij}^{(\vartheta,\eta)}-\bigl(U_{ij}^{(\vartheta,\eta)}\bigr)^\top,
\label{eq:spectral-component-matrix}
\end{equation}
where \(U_{ij}^{(\vartheta,\mathrm{ind})}=0\) and
\(U_{ij}^{(\vartheta,\mathrm{geo})}=G_{ij}^{(\vartheta)}\). Holding $\boldsymbol{\Gamma}
=\left\{\boldsymbol{\lambda}^{(\vartheta)}\in\Pi_{\vartheta}:
\vartheta\in\mathcal I_B\cup\mathcal I_W\right\}$
fixed, aggregate these matrices as
\begin{equation}
C_{\iota,\eta}(\boldsymbol\Gamma)
=\frac{1}{|\mathcal I_\iota|}
\sum_{\vartheta\in\mathcal I_\iota}\sum_{ij}
\lambda_{ij}^{(\vartheta)}H_{ij}^{(\vartheta,\eta)},
\qquad \iota\in\{\mathrm{B},\mathrm{W}\}.
\label{eq:spectral-scatter-matrices}
\end{equation}
The corresponding bound-based fixed-coupling variations are
\[
\widetilde V_{\iota,\mathsf{gmm}}^{\eta,\mathrm{fix}}
(A;\boldsymbol\Gamma)
=\operatorname{tr}\!\left\{A^\top C_{\iota,\eta}(\boldsymbol\Gamma)A\right\}.
\]
Appendix~\ref{ap2} derives this identity, including the mixture-component and
subject-pair aggregation, for both the independent and geometric couplings.
For one direction, their between-to-within ratio is the generalized Rayleigh quotient
\begin{equation}
\widetilde r_{\mathsf{gmm}}^{\eta,\mathrm{fix}}
(a;\boldsymbol\Gamma)
=\frac{a^\top C_{B,\eta}(\boldsymbol\Gamma)a}
{a^\top C_{W,\eta}(\boldsymbol\Gamma)a}.
\label{eq:spectral-rayleigh-quotient}
\end{equation}
When \(C_{W,\eta}\) is positive definite, Eq.~\eqref{eq:spectral-rayleigh-quotient}
is maximized by a leading generalized eigenvector of the pair
\((C_{B,\eta},C_{W,\eta})\). The
implementation uses the diagonal loading
\[
C_{W,\eta,\rho_{\mathrm{load}}}=C_{W,\eta}+\rho_{\mathrm{load}}I_d,
\qquad
\rho_{\mathrm{load}}=0.001\,\operatorname{mean}\{\operatorname{diag}(C_{W,\eta})\},
\]
so it maximizes the loaded quotient instead. This small diagonal ridge prevents a
singular or nearly singular within-class matrix from making the generalized eigenvalue
problem numerically unstable. It regularizes only the construction of the spectral
candidate; acceptance is still determined by the unregularized, exact
transport-reoptimized objective. For \(q>1\), the leading generalized eigenvectors
define a candidate span, which is orthonormalized before exact screening. 

For empirical measures, the same diagonal-loading and multi-direction construction
apply with \(C_{\iota,\mathsf{emp}}(\boldsymbol\Gamma)\) from
Eq.~\eqref{eq:empirical-spectral-scatter} in place of \(C_{\iota,\eta}(\boldsymbol\Gamma)\).
For GMMs, both spectral constructions are approximations used only to generate candidates. We consider both because they yield quadratic costs—and hence generalized eigenvalue problems.
Independence is the simplest choice, whereas
the geometric coupling gives the tighter componentwise upper bound in
Eq.~\eqref{eq:geometric-bound-ordering}. That ordering alone cannot establish which
proposal gives the better Fisher ratio, because both its numerator and denominator are
approximated. We therefore treat the two couplings as alternative algorithm-design
proposals and compare them empirically in Section~\ref{sec:algorithm_experiments}; those
experiments favor the geometric proposal. The independent construction is retained to
make that design comparison explicit and for sensitivity analysis. Appendix~\ref{ap4} gives its detailed fixed-coupling
ratio-discrepancy and maximizer bounds.

\subsection{The Screened GeoGrad Algorithm}
\label{sec:initialization_modes}
The gradient ascent optimizer starts with
an initialization of the \(q\)-dimensional projection \(A_0\). For every proposed \(A_0\), OTAF projects the subject
representations, solves all selected transport problems, and thereby obtains the
corresponding couplings \(\boldsymbol\Gamma_s^{\mathrm{opt}}(A_0)\). We use several
complementary, inexpensive candidates because the objective is nonconcave.

For a GMM subject \(k\), define its distribution mean and covariance by
\begin{equation}
\bar{\mu}^{(k)}=\sum_i p_i^{(k)}\mu_i^{(k)},
\qquad
\bar\Sigma^{(k)}=\sum_i p_i^{(k)}\left\{\Sigma_i^{(k)}+
(\mu_i^{(k)}-\bar{\mu}^{(k)})
(\mu_i^{(k)}-\bar{\mu}^{(k)})^\top\right\}.
\label{eq:subject-moments-initialization}
\end{equation}
The latter
includes both within-component variation and variation between component means.
The corresponding empirical moments are used for empirical input.
We employ the following four approaches to construct candidate initial subspaces
\(A_0\):
\begin{enumerate}[leftmargin=*,label=(\roman*),itemsep=4pt,topsep=4pt]
\item \textbf{Ambient-coupling initialization:}
Solve the pairwise transports in the original \(d\)-dimensional representation,
hold those couplings fixed, and form the independent-coupling spectral candidate
of Section~\ref{sec:spectral_proposals}. This candidate uses class information
through the between- and within-class pair sets. We abbreviate it as ``Ambient''
in the experimental tables.

\item \textbf{Pooled-moment principal component analysis (PCA):}
Give each subject equal weight, put
\(\bar{\mu}=n^{-1}\sum_{k=1}^n\bar{\mu}^{(k)}\), and form
\begin{equation}
S_{\mathrm{pool}}=\frac1n\sum_{k=1}^n
\left\{\bar\Sigma^{(k)}+
(\bar{\mu}^{(k)}-\bar{\mu})(\bar{\mu}^{(k)}-\bar{\mu})^\top\right\}.
\label{eq:pooled-pca-initialization}
\end{equation}
Its leading \(q\) eigenvectors give the PCA candidate. Thus PCA uses both
within-subject distributional variability and between-subject mean variability; it does
not simply pool cells, which would overweight subjects containing more observed cells.
Subject-level class labels are not used. The motivation is to begin in a subspace containing dominant
distributional variation.

\item \textbf{Subject-mean LDA:}
Represent each subject by \(\bar{\mu}^{(k)}\), calculate class means and the ordinary
between- and within-class scatter matrices from these \(n\) subject vectors, and regularize the
within-class matrix by \(10^{-3}cI_d\), where \(c\) is the maximum of 1 and the mean
absolute diagonal entries of the within- and between-class matrices. The leading
generalized eigenvector is the LDA direction. This is a label-aware summary candidate;
it is not cell-level
LDA and intentionally ignores \(\bar\Sigma^{(k)}\) when constructing its first direction. For a
binary outcome and \(q>1\), LDA supplies only one nonzero discriminant direction. We
therefore form an LDA-enriched candidate by orthogonally completing that direction with
leading pooled-PCA directions, rather than labeling arbitrary null-space vectors as
additional LDA directions.

\item \textbf{Random initialization:}
For each random candidate, we generate a \(d\times q\) matrix by sampling each entry
independently from the standard Gaussian distribution \(N(0,1)\), and then compute its
QR factorization.  The first
\(q\) orthonormal columns of the \(Q\) factor define a candidate \(q\)-dimensional
subspace.  Because the Gaussian distribution is rotationally invariant, the resulting
subspace is uniformly distributed over \(\operatorname{Gr}(q,d)\); equivalently, the random
initialization does not favor any particular orientation.  The random matrices or their
generating seeds are recorded to make the initialization reproducible.
\end{enumerate}

Ambient-coupling, PCA, LDA, and random subspaces are candidates only.
Each is evaluated by the exact transport-reoptimized objective \(F_s\), and the largest
screened value supplies \(A_0\). All label-aware construction and screening must be
repeated inside each training fold.

The \emph{Screened GeoGrad} algorithm is presented in 
Algorithm~\ref{alg:OTAF}.
It screens the candidate initial subspaces by their transport-reoptimized
objective values, attempts one safeguarded geometric proposal from the selected
subspace, and then uses the exact-gradient iteration in
Section~\ref{sec:gradient_otaf}. Our experiments use 
all the four initialization schemes described above;
the number of random candidates (eight by default) is a user option.

We include the geometric step in Screened GeoGrad because the
experiments in Section~\ref{sec:algorithm_experiments} show that it
can improve the terminal objective when only a few optimization
updates are affordable. The additional implementation
is limited to one safeguarded proposal, which users can disable
to obtain screened gradient ascent with the same conditional
convergence guarantee.

\LinesNotNumbered
\begin{algorithm}[t]
\color{black}
\caption{Screened GeoGrad}
\label{alg:OTAF}
\centering
\begin{minipage}{0.90\linewidth}
\begin{algorithmic}[1]
\footnotesize
\Require{Fixed empirical or GMM subject representations and labels; selected pair sets
\(\mathcal I_B,\mathcal I_W\); dimension \(q\); number \(R_{\rm init}\) of random
starts (default 8); gradient tolerance and iteration cap.}
\State{Construct ambient-coupling, pooled-moment PCA, subject-mean LDA, and
\(R_{\rm init}\) seeded random candidates using
training subjects only.}
\State{For each candidate \(A\), reoptimize all projected transports and evaluate
\(F_s(A)\); select the candidate with largest exact value and denote it by \(A_0\).}
\State{At the selected initialization \(A_0\), use its optimized component couplings to
construct the geometric proposal \(A_{\mathrm{geo}}\).}
\State{Test the full move from \(A_0\) to \(A_{\mathrm{geo}}\), followed if needed by
moves covering \(1/2\), \(1/4\), \(1/8\), and \(1/16\) of a shortest
Grassmann geodesic between the two subspaces, constructed as described in
Appendix~\ref{ap:grassmann-geodesic}.
At every trial, reoptimize all transports and evaluate the exact \(F_s\).}
\State{Accept the first trial whose exact objective is no smaller than \(F_s(A_0)\)
within the relative tolerance described below; if none succeeds, retain \(A_0\). Then begin
gradient ascent.}
\State{\textbf{while} the gradient norm exceeds its tolerance and the iteration cap has not been reached
\textbf{do}}
\State{Reoptimize projected transports and form the exact Riemannian gradient using
Eq.~\eqref{eq:exact-grassmann-gradient}.}
\State{Apply QR-retraction Armijo backtracking using the transport-reoptimized acceptance
test in Eq.~\eqref{eq:gradient-armijo}.}
\State{\textbf{end while}}
\State{Return the visited subspace with largest exact objective and report its gradient
norm and stopping reason.}
\end{algorithmic}
\end{minipage}
\end{algorithm}

The default gradient tolerance is \(\epsilon_{\mathrm{grad}}=10^{-7}\).
For a geometric trial with ratio \(F_{\mathrm{trial}}\), the acceptance tolerance
is \(10^{-10}\max\{1,|F_s(A_0)|,|F_{\mathrm{trial}}|\}\): a decrease smaller
than this numerical scale is treated as a tie. Setting this tolerance to zero
requires literal nondecrease. The best visited projection is retained in either case.

The initialization screen and the single safeguarded geometric proposal can be viewed
together as selecting the starting subspace for the gradient phase.
Theorem~\ref{thm:gradient-convergence-rev3} allows an arbitrary initial subspace and
therefore does not depend on how that starting point is constructed. Consequently, if
the theorem's regularity assumptions hold for the subspace produced by this preliminary
stage, its first-order convergence conclusion applies to the subsequent idealized
gradient iteration. 

\subsection{Computational Comparison}
\label{sec:optimizer_complexity}

We compare computation after the empirical or GMM representation has been formed; GMM
fitting and empirical subsampling are excluded. Let 
\(n_{\mathrm{pair}}=|\mathcal I_B\cup\mathcal I_W|\), \(\bar{m}\) be an upper bound on empirical atoms per subject, and
\(\bar{k}\) be an upper bound on GMM components. When all subjects have the same
representation size, these bounds equal the counts \(m\) and \(K\), respectively,
used in the subsequent statistical analysis. 
Let \(\mathcal T_{\mathrm{LP}}(u,\epsilon_{\mathrm{LP}})\) denote
the computational cost of solving one optimal transport linear
program between two distributions, each with at most \(u\)
support points or Gaussian components.
Here
\(0<\epsilon_{\mathrm{LP}}<1\) is the requested numerical accuracy of the linear-program
solve, as specified by the solver's optimality and feasibility tolerances. This
solver-dependent term is kept explicit. For
orientation, a path-following interior-point convention with transportation-structured
normal equations gives
\begin{equation}
\mathcal T_{\mathrm{LP}}(u,\epsilon_{\mathrm{LP}})
=O\{u^4\log(1/\epsilon_{\mathrm{LP}})\},
\label{eq:transport_ip_complexity}
\end{equation}
using \(O(u)\) Newton steps and \(O(u^3)\) structured linear algebra per step; practical
sparse solvers need not follow this power law~\citep{wright2023direct}.

\subsubsection{Three Fitted-GMM Optimizer Paths}

For dense full covariances, one exact transport-reoptimized evaluation costs
\begin{equation}
\mathcal T_{\mathrm{eval},\mathsf{gmm}}
=O\!\left[n\bar{k}(dq+d^2q+dq^2)+n_{\mathrm{pair}}\bar{k}^2q^3
+n_{\mathrm{pair}}\mathcal T_{\mathrm{LP}}(\bar{k},\epsilon_{\mathrm{LP}})\right],
\label{eq:exact-evaluation-complexity-rev3}
\end{equation}
and one analytical exact-gradient formation has the conservative cost
\begin{equation}
\mathcal T_{\mathrm{grad},\mathsf{gmm}}
=O\!\left[n\bar{k}d^2q+n_{\mathrm{pair}}\bar{k}^2(q^3+dq^2+dq)\right].
\label{eq:gradient-complexity-rev3}
\end{equation}
The independent- and geometric-coupling spectral proposals each have per-formation cost
\begin{equation}
\mathcal T_{\mathrm{prop},\mathrm{ind}}
=O(n_{\mathrm{pair}}\bar{k}^2d^2+d^3),\qquad
\mathcal T_{\mathrm{prop},\mathrm{geo}}
=O(n_{\mathrm{pair}}\bar{k}^2d^2+d^3).
\label{eq:spectral-complexity-rev3}
\end{equation}
The geometric matrices additionally require a one-time cache with
\(\mathcal T_{\mathrm{cache},\mathrm{geo}}=O(n_{\mathrm{pair}}\bar{k}^2d^3)\) construction time and
\(O(n_{\mathrm{pair}}\bar{k}^2d^2)\) memory. This cache stores the geometric-coupling matrices, which depend only
on the supplied component covariances. They are computed once and
reused whenever a geometric proposal is constructed, avoiding
repeated matrix-square-root calculations across iterations.

To compare complete paths, assume the same \(t_{\mathrm{iter}}\) outer
updates and a common cap of \(\bar{b}\) exact safeguard evaluations per update. Starting
from the same fixed initial subspace, the three total costs are
\begin{align}
\mathcal T_{\nabla,\mathsf{gmm}}
&=O\{\mathcal T_{\mathrm{eval},\mathsf{gmm}}
+t_{\mathrm{iter}}(\mathcal T_{\mathrm{grad},\mathsf{gmm}}+\bar{b}\,\mathcal T_{\mathrm{eval},\mathsf{gmm}})\},\notag\\
\mathcal T_{\mathrm{ind},\mathsf{gmm}}
&=O\{\mathcal T_{\mathrm{eval},\mathsf{gmm}}
+t_{\mathrm{iter}}(\mathcal T_{\mathrm{prop},\mathrm{ind}}+\bar{b}\,\mathcal T_{\mathrm{eval},\mathsf{gmm}})\},\notag\\
\mathcal T_{\mathrm{geo},\mathsf{gmm}}
&=O\{\mathcal T_{\mathrm{cache},\mathrm{geo}}+\mathcal T_{\mathrm{eval},\mathsf{gmm}}
+t_{\mathrm{iter}}(\mathcal T_{\mathrm{prop},\mathrm{geo}}+\bar{b}\,\mathcal T_{\mathrm{eval},\mathsf{gmm}})\}.
\label{eq:three_optimizer_total_costs}
\end{align}
The three lines correspond to gradient-only, independent-coupling, and
geometric-coupling paths. Each spectral path uses its stated proposal at every update.
The formulas also show why a
spectral proposal is not automatically faster: all three paths pay for exact
safeguarding, and the geometric path additionally pays for its cache. Under the
conditions of Theorem~\ref{thm:gradient-convergence-rev3}, only the gradient path has the
stated first-order convergence guarantee.

\subsubsection{Empirical versus Fitted-GMM Gradient OTAF}

For an empirical representation, one exact evaluation and one directly contracted
analytical gradient cost
\begin{align}
\mathcal T_{\mathrm{eval},\mathsf{emp}}
&=O\!\left[n\bar{m}dq+n_{\mathrm{pair}}\bar{m}^2q
+n_{\mathrm{pair}}\mathcal T_{\mathrm{LP}}(\bar{m},\epsilon_{\mathrm{LP}})\right],\notag\\
\mathcal T_{\mathrm{grad},\mathsf{emp}}
&=O(n_{\mathrm{pair}}\bar{m}^2dq).
\label{eq:empirical_evaluation_gradient_costs}
\end{align}
The gradient cost in the second line follows by computing each
support-point pair's contribution as \(z(z^\top A)\), where
\(z=x_i^{(k_1)}-x_j^{(k_2)}\). This requires \(O(dq)\) operations
per pair and avoids constructing the \(d\times d\) matrix \(zz^\top\).
Under the same \(t_{\mathrm{iter}}\)-update and \(\bar{b}\)-trial budget as above,
\begin{equation}
\mathcal T_{\nabla,\mathsf{emp}}
=O\{\mathcal T_{\mathrm{eval},\mathsf{emp}}
+t_{\mathrm{iter}}(\mathcal T_{\mathrm{grad},\mathsf{emp}}+\bar{b}\,\mathcal T_{\mathrm{eval},\mathsf{emp}})\}.
\label{eq:empirical_gradient_total_cost}
\end{equation}
Table~\ref{tab:implemented_complexity} summarizes and compares
the computational costs of gradient-based OTAF for empirical
and fitted-GMM representations.
\begin{table}[t]
\centering
\color{black}
\small
\begin{tabular}{>{\raggedright\arraybackslash}p{0.24\textwidth}
>{\raggedright\arraybackslash}p{0.32\textwidth}
>{\raggedright\arraybackslash}p{0.34\textwidth}}
\toprule
Quantity & Empirical representation & Fitted-GMM representation\\
\midrule
Stored subject representation
& \(O(n\bar{m}d)\)
& \(O\{n\bar{k}(d^2+d)\}\)\\
One exact objective evaluation
& \(\mathcal T_{\mathrm{eval},\mathsf{emp}}\) in Eq.~\eqref{eq:empirical_evaluation_gradient_costs}
& \(\mathcal T_{\mathrm{eval},\mathsf{gmm}}\) in Eq.~\eqref{eq:exact-evaluation-complexity-rev3}\\
One exact-gradient formation
& \(O(n_{\mathrm{pair}}\bar{m}^2dq)\)
& \(\mathcal T_{\mathrm{grad},\mathsf{gmm}}\) in Eq.~\eqref{eq:gradient-complexity-rev3}\\
Complete \(t_{\mathrm{iter}}\)-update gradient run
& \(O\{\mathcal T_{\mathrm{eval},\mathsf{emp}}+t_{\mathrm{iter}}(\mathcal T_{\mathrm{grad},\mathsf{emp}}+\bar{b}\,\mathcal T_{\mathrm{eval},\mathsf{emp}})\}\)
& \(O\{\mathcal T_{\mathrm{eval},\mathsf{gmm}}+t_{\mathrm{iter}}(\mathcal T_{\mathrm{grad},\mathsf{gmm}}+\bar{b}\,\mathcal T_{\mathrm{eval},\mathsf{gmm}})\}\)\\
Generic LP-dominated term under Eq.~\eqref{eq:transport_ip_complexity}
& \(O(t_{\mathrm{iter}}\bar{b}\,n_{\mathrm{pair}}\bar{m}^4\log(1/\epsilon_{\mathrm{LP}}))\)
& \(O(t_{\mathrm{iter}}\bar{b}\,n_{\mathrm{pair}}\bar{k}^4\log(1/\epsilon_{\mathrm{LP}}))\)\\
\bottomrule
\end{tabular}
\caption{Post-representation cost comparison for exact-gradient OTAF.
The same subject-pair sets, outer-update budget, safeguard cap, and transport tolerance
are used in both columns. Representation construction is intentionally excluded.}
\label{tab:implemented_complexity}
\end{table}

At equal retained size \(\bar{m}=\bar{k}\), empirical OTAF is normally cheaper since it does not need to handle covariance. A possible GMM
advantage arises only when a compact mixture replaces a substantially larger empirical
support. If the cost of solving transport dominates and Eq.~\eqref{eq:transport_ip_complexity} is
used, the empirical-to-GMM LP cost ratio is approximately \((\bar{m}/\bar{k})^4\). For \(q=1\) with equal empirical weights, because sorting can replace the generic empirical LP,
this fourth power of the ratio does not apply.
If the cost of constructing the reduced-space cost matrices
dominates, comparing \(\bar{m}^2q\) with \(\bar{k}^2q^3\) favors GMM roughly when
\(\bar{m}>\bar{k}q\). If projection cost dominates, comparing \(n\bar{m}dq\) with \(n\bar{k}d^2q\) favors GMM
roughly when \(\bar{m}>\bar{k}d\). 
These thresholds indicate when GMM may be faster under the respective
dominant-cost assumptions; actual runtime also depends on the remaining
computational costs and implementation.

These cost formulas condition on the supplied representation sizes. How large those sizes should be is a separate question. Section~\ref{sec:representation_error_propagation} relates distribution-estimation accuracy to the Fisher ratio and selected subspace, then
uses sample-size benchmarks to put the computational comparison on a statistical footing.
Section~\ref{sec:representation_diagnostics_rev3} reports the corresponding
timing and sample-size experiments.

\subsection{Propagation of Finite-Representation Error}
\label{sec:representation_error_propagation}
\label{sec:gmm_estimation_perturbation}

The computational comparison above treats the representation sizes as given. We now
ask two statistical questions: 
\begin{enumerate}
\item
\emph{How does error in estimating each subject's
distribution affect the Fisher ratio and its maximizing subspace?}
\item
\emph{How many observations per subject suffice to attain a prescribed accuracy?}
\end{enumerate}
We answer these
questions separately for empirical measures and fitted GMMs, then explain what their
different sample-size requirements imply for computation. We retain the stability
results and their computational implications here; Appendix~\ref{ap5} contains the
proofs, detailed sample-size conditions, and derivations.

To isolate estimation error, the subspace comparisons below concern global maximizers
of the population and estimated objectives. They do not assert that OTAF finds these
maximizers: any remaining optimization gap is a separate source of error. Likewise,
ordinary \(W_2\) and MAW can define different population ratios even with unlimited
data. That difference is a target-metric discrepancy, not estimation error; it is
addressed at the end of this subsection.

We use the same fixed subject-pair sets \(\mathcal I_B,\mathcal I_W\) for the
population and estimated objectives. For \(\vartheta=(k_1,k_2)\), define the
population projected pair costs by
\[
\begin{aligned}
D_{\mathsf{emp}}^{(\vartheta),0}(A)
&=W^2(A_\#\mathcal P_{k_1},A_\#\mathcal P_{k_2}),\\
D_{\mathsf{gmm}}^{(\vartheta),0}(A)
&=\widetilde W^2(A_\#\mathcal G_{k_1}^0,A_\#\mathcal G_{k_2}^0).
\end{aligned}
\]
Here \(\mathcal P_k\) is the true subject distribution, and \(\mathcal G_k^0\)
is its true mixture representation in the correctly specified GMM setting described
below. The first cost uses ordinary Wasserstein distance; the second uses MAW on
the true mixture parameters. Both costs minimize the relevant pairwise transport
problem at each projection \(A\). For \(s\in\{\mathsf{emp},\mathsf{gmm}\}\), $\iota \in\{\mathrm{B}, \mathrm{W}\}$, the
population transport-reoptimized variations and ratio are defined as
\[
\begin{aligned}
V_{\iota,s}^0(A)&=\frac{1}{|\mathcal I_\iota|}
\sum_{\vartheta\in\mathcal I_\iota}D_s^{(\vartheta),0}(A),
\qquad r_s^0(A)=\frac{V_{B,s}^0(A)}{V_{W,s}^0(A)}.
\end{aligned}
\]
Here the superscript \(0\) denotes population inputs.
Replacing the population inputs by their estimated representations gives
\(\widehat V_{B,s},\widehat V_{W,s}\) and \(\widehat r_s\).
For those supplied inputs, \(\widehat r_s(A)=F_s(A)=r_s^{\mathrm{opt}}(A)\);
the hat emphasizes estimation from data in this subsection. All suprema below
are over orthonormal \(A\in\mathbb R^{d\times q}\).
We write \(\zeta_s\) for a squared-pair-cost error bound,
\(\zeta^{\mathrm{up}}_{B,s}\) for an upper bound on the population between-class
variation, and \(\nu_s>0\) for a lower bound on the population within-class variation.

\subsubsection{Empirical-Measure Stability}
We first quantify how distribution-estimation error affects the objective and its
maximizing subspace.
Let \(\widehat{\mathcal P}_k\) be the empirical measure based on \(m\) independent
observations from \(\mathcal P_k\). Here \(m\) is the retained support size, taken equal
across subjects for this rate calculation. Let \(\varepsilon_{\mathsf{emp}}\) bound
the distribution-estimation error, measured in \(W_2\):
\begin{equation}
\max_k W_2(\widehat{\mathcal P}_k,\mathcal P_k)\leq\varepsilon_{\mathsf{emp}}.
\label{eq:empirical_subject_error}
\end{equation}
If every selected population pair has \(W_2\) distance at most \(D_{\mathsf{emp}}^{\mathsf{up}}\),
let
\begin{eqnarray}
\zeta_{\mathsf{emp}}:=4\varepsilon_{\mathsf{emp}}
(D_{\mathsf{emp}}^{\mathsf{up}}+\varepsilon_{\mathsf{emp}}) \; .
\label{eq:zeta_emp}
\end{eqnarray}
This bounds the error in every projected squared pair cost and, consequently,
in each normalized variation. The following proposition translates this bound into
uniform accuracy of the ratio in Eq.~\eqref{eq:empirical_ratio_estimation_bound},
the loss in the population objective at the estimated maximizer in
Eq.~\eqref{eq:empirical_argmax_regret}, and the subspace distance in
Eq.~\eqref{eq:empirical_argmax_subspace_bound}.
Here \(d_{\mathrm{Gr}}\) denotes the Grassmann geodesic distance: the square root
of the sum of squared principal angles between two subspaces, as defined in
Appendix~\ref{ap:grassmann}.

\begin{proposition}[Empirical-measure objective stability]
\label{prop:empirical_estimation_chain}
Under Eq.~\eqref{eq:empirical_subject_error}, the empirical between- and within-class
variations differ from their population counterparts by at most \(\zeta_{\mathsf{emp}}\).
If, uniformly in \(A\),
\[
V_{W,\mathsf{emp}}^{0}(A)\geq\nu_{\mathsf{emp}}>\zeta_{\mathsf{emp}},
\qquad
V_{B,\mathsf{emp}}^{0}(A)\leq \zeta^{\mathrm{up}}_{B,\mathsf{emp}},
\]
then
\begin{equation}
\sup_A|\widehat r_{\mathsf{emp}}(A)-r_{\mathsf{emp}}^{0}(A)|
\leq\varepsilon_{r,\mathsf{emp}}:=
\frac{\zeta_{\mathsf{emp}}(\nu_{\mathsf{emp}}+\zeta^{\mathrm{up}}_{B,\mathsf{emp}})}
{\nu_{\mathsf{emp}}(\nu_{\mathsf{emp}}-\zeta_{\mathsf{emp}})}.
\label{eq:empirical_ratio_estimation_bound}
\end{equation}
If \(A_{\mathsf{emp}}^{0}\) and \(\widehat A_{\mathsf{emp}}\) globally maximize the
population and empirical ratios, respectively, then
\begin{equation}
0\leq r_{\mathsf{emp}}^{0}(A_{\mathsf{emp}}^{0})
-r_{\mathsf{emp}}^{0}(\widehat A_{\mathsf{emp}})
\leq2\varepsilon_{r,\mathsf{emp}}.
\label{eq:empirical_argmax_regret}
\end{equation}
If \(\operatorname{span}\widehat A_{\mathsf{emp}}\) lies in a Grassmann neighborhood
\(\mathcal N_{\mathsf{emp}}\) of \(\operatorname{span}A_{\mathsf{emp}}^{0}\) and there is
a constant \(\varrho_{\mathsf{emp}}>0\) such that, for every \(A\) with
\(\operatorname{span}A\in\mathcal N_{\mathsf{emp}}\),
\[
r_{\mathsf{emp}}^{0}(A_{\mathsf{emp}}^{0})-r_{\mathsf{emp}}^{0}(A)
\geq\frac{\varrho_{\mathsf{emp}}}2d_{\mathrm{Gr}}^2
(\operatorname{span}A,\operatorname{span}A_{\mathsf{emp}}^{0}),
\]
then
\begin{equation}
d_{\mathrm{Gr}}(\operatorname{span}\widehat A_{\mathsf{emp}},
\operatorname{span}A_{\mathsf{emp}}^{0})
\leq2\sqrt{\varepsilon_{r,\mathsf{emp}}/\varrho_{\mathsf{emp}}}.
\label{eq:empirical_argmax_subspace_bound}
\end{equation}
\end{proposition}

The constant
\(\varrho_{\mathsf{emp}}\) in the subspace conclusion measures how sharply the
population optimum is identified. This local quadratic-growth assumption is needed
only to bound subspace error. In contrast, a nearly flat objective can have a small value error
without identifying a unique nearby subspace. Appendix~\ref{ap:statistical-subspaces}
discusses this condition and proves the conclusions.

\subsubsection{Fitted-GMM Stability under MAW}

We first elaborate on the relationship between component-estimation error and objective and subspace error. 
Here the population comparison uses MAW on the true mixture parameters, not
ordinary Wasserstein distance between the mixture densities.
For subject \(k\), let
\[
\mathcal G_k^0=\sum_{i=1}^{K}p_i^{(k)}\phi_i^{(k)},\qquad
\widehat{\mathcal G}_k=\sum_{i=1}^{K}\widehat p_i^{(k)}
\widehat\phi_i^{(k)}
\]
be an oracle \(K\)-component GMM and its fit. Assume that the fitted components admit
a common relabeling such that, uniformly in \(k,i\),
\begin{equation}
\|\widehat p^{(k)}-p^{(k)}\|_1\leq\varepsilon_w,
\qquad
W_2(\widehat\phi_i^{(k)},\phi_i^{(k)})\leq\varepsilon_{\mathsf{gmm}}.
\label{eq:gmm_parameter_error}
\end{equation}
Component matching is an identifiability assumption, not a consequence of small
mixture-density error. If every relevant oracle component pair has \(W_2\) distance
at most \(D_{\mathsf{gmm}}\), define
\begin{equation}
\zeta_{\mathsf{gmm}}:=
4\varepsilon_{\mathsf{gmm}}(D_{\mathsf{gmm}}+\varepsilon_{\mathsf{gmm}})
+D_{\mathsf{gmm}}^2\varepsilon_w.
\label{eq:gmm_maw_error_constant}
\end{equation}

\begin{proposition}[Fitted-GMM objective stability]
\label{prop:gmm_estimation_chain}
Under Eq.~\eqref{eq:gmm_parameter_error}, each selected fitted squared MAW value and each
normalized fitted variation differs from its oracle counterpart by at most
\(\zeta_{\mathsf{gmm}}\), uniformly in \(A\). If
\[
V_{W,\mathsf{gmm}}^{0}(A)\geq\nu_{\mathsf{gmm}}>\zeta_{\mathsf{gmm}},
\qquad
V_{B,\mathsf{gmm}}^{0}(A)\leq \zeta^{\mathrm{up}}_{B,\mathsf{gmm}}
\]
uniformly in \(A\), then
\begin{equation}
\sup_A|\widehat r_{\mathsf{gmm}}(A)-r_{\mathsf{gmm}}^{0}(A)|
\leq\varepsilon_{r,\mathsf{gmm}}:=
\frac{\zeta_{\mathsf{gmm}}(\nu_{\mathsf{gmm}}+\zeta^{\mathrm{up}}_{B,\mathsf{gmm}})}
{\nu_{\mathsf{gmm}}(\nu_{\mathsf{gmm}}-\zeta_{\mathsf{gmm}})}.
\label{eq:gmm_ratio_estimation_bound}
\end{equation}
If \(A_{\mathsf{gmm}}^{0}\) and \(\widehat A_{\mathsf{gmm}}\) globally maximize the
oracle and fitted ratios, respectively, then the oracle-objective regret is at most
\(2\varepsilon_{r,\mathsf{gmm}}\). Under the analogous local quadratic-growth condition,
with \(\operatorname{span}\widehat A_{\mathsf{gmm}}\) in the relevant neighborhood and
curvature constant \(\varrho_{\mathsf{gmm}}\),
\begin{equation}
d_{\mathrm{Gr}}(\operatorname{span}\widehat A_{\mathsf{gmm}},
\operatorname{span}A_{\mathsf{gmm}}^{0})
\leq2\sqrt{\varepsilon_{r,\mathsf{gmm}}/\varrho_{\mathsf{gmm}}}.
\label{eq:gmm_argmax_subspace_bound}
\end{equation}
\end{proposition}

Appendix~\ref{ap5} proves Propositions~\ref{prop:empirical_estimation_chain}
and~\ref{prop:gmm_estimation_chain}. Neither proposition requires a unique optimal
transport plan because it controls optimal values, not the selected couplings or their
derivatives. In particular, the sample-size results in Appendix~\ref{ap5} do not establish stability of
the coupling matrices. Near a tie, small cost perturbations can change an optimal
coupling substantially while its value changes little. The envelope regularity and
smoothness assumptions in Theorem~\ref{thm:gradient-convergence-rev3} are additional
conditions; we do not prove a sample-size bound that ensures them throughout the
optimization path.

\subsubsection{Implications for Representation Choice and Computation}

For this comparison, suppose each subject has a true \(K\)-component GMM in ambient
dimension \(d\). Let \(m\) be the number of retained observations in its empirical
representation and \(N_{\mathrm{cell}}\) the number of cells used to fit its GMM.
Fix a confidence level \(1-\varepsilon_{\mathrm{fail}}\), where
\(0<\varepsilon_{\mathrm{fail}}<1\) is the allowed probability of failing to meet
the accuracy target, and let \(0<\varepsilon_{\mathrm{tar}}<1\) be the desired
uniform Fisher-ratio error tolerance. Under the respective assumptions below,
sufficiently large sample sizes of the stated orders ensure, separately for each representation
\(s\in\{\mathsf{emp},\mathsf{gmm}\}\), that
\[
\mathbb P\!\left\{
\sup_A|\widehat r_s(A)-r_s^0(A)|\leq\varepsilon_{\mathrm{tar}}
\right\}\geq1-\varepsilon_{\mathrm{fail}}.
\]
Here \(\widehat r_s\) is the ratio computed from the empirical measures or fitted
GMMs, and \(r_s^0\) is its corresponding population ratio. The probability is over
the observations sampled for all subjects, and the supremum is over all orthonormal
projections \(A\). Thus, with the prescribed confidence, the estimated ratio is
within \(\varepsilon_{\mathrm{tar}}\) of its own population counterpart
simultaneously at every admissible projection.

Combining Proposition~\ref{prop:empirical_estimation_chain} with
\citet{fournier2015rate} gives a sufficient empirical support size
\(m\geq c_{\mathrm{conf}}\varepsilon_{\mathrm{tar}}^{-d}\) when \(d>4\).
Here \(c_{\mathrm{conf}}>0\) is a sufficiently large constant depending on the
prescribed confidence, number of subjects, dimension, moment bounds, and the distance
and variation bounds in the proposition, but not on \(\varepsilon_{\mathrm{tar}}\).
For correctly specified, well-separated spherical mixtures with weights of order
\(1/K\), component standard deviations bounded above by a fixed constant, and
suitable initialization,
combining Proposition~\ref{prop:gmm_estimation_chain} with \citet{kwon2020em} gives
the sufficient fitting-sample order
\(N_{\mathrm{cell}}=\widetilde O(Kd/\varepsilon_{\mathrm{tar}}^2)\), where
\(\widetilde O\) suppresses logarithmic factors.
Both comparisons assume that the relevant population distances and between-class variations are bounded above, while the population within-class variations are bounded below by a positive constant, uniformly over the admissible projections. When comparing different ambient dimensions, these upper bounds must remain bounded and the positive lower bound must remain away from zero. 
Appendix~\ref{ap5} gives the full assumptions and derivations.

Empirical OTAF transports \(m\) support points per subject, whereas fitted-GMM
OTAF transports \(K\) components, independently of \(N_{\mathrm{cell}}\).
Combining the sufficient sample-size bounds above with the costs in
Section~\ref{sec:optimizer_complexity} illustrates how compact GMMs can
reduce computation after fitting. This comparison is indicative: the bounds
may be conservative, and their constants may depend on dimension.
For these comparisons, we assume that each estimated objective is
globally maximized, so that the resulting differences reflect
representation error rather than optimization error.

Another subtlety in the comparison is 
that the two methods optimize different population
ratios, both assumed to be acceptable. Empirical OTAF targets the ordinary-\(W_2\) ratio
\(r_{\mathsf{emp}}^{0}\), while GMM OTAF targets the MAW ratio
\(r_{\mathsf{gmm}}^{0}\). Define their structural discrepancy by
$\Delta_{\mathrm{MAW}}
=\sup_A|r_{\mathsf{gmm}}^{0}(A)-r_{\mathsf{emp}}^{0}(A)|$.
MAW restricts transport to Gaussian-component couplings and can differ from ordinary
mixture \(W_2\) even with infinite data~\citep{delon2020wasserstein}. If ordinary
\(W_2\) is declared the common target, the fitted-GMM maximizer has population regret
bounded by \(2(\varepsilon_{r,\mathsf{gmm}}+\Delta_{\mathrm{MAW}})\).

In practice, covariance arithmetic, fitting sample size, model misspecification,
solver behavior, and intrinsic dimension also affect the choice. We therefore do not
recommend GMMs as a universal solution in high dimensions. When both representations
are feasible, they should be compared within the training data using held-out
objectives or downstream performance, with runtime and memory reported separately.

\section{Experiments}
\label{sec:exp}

We evaluate OTAF in three stages: the optimization and initialization choices that
define Screened GeoGrad (Section~\ref{sec:algorithm_experiments}), the statistical and computational tradeoffs between
representations (Section~\ref{sec:representation_diagnostics_rev3}), and classification on simulated and real distribution-valued data (Section~\ref{sec:predictive_demonstrations}).
Unless otherwise stated, all OTAF experiments use Screened GeoGrad.
All Gaussian costs use Eq.~\eqref{eq:w_gaussian}. A larger
transport-reoptimized Fisher ratio is better, but does not necessarily imply better
classification. The OTAF computations use MATLAB R2024a, with MOSEK 11 for
general transport linear programs, on an Apple M2 Pro with 16\,GiB RAM.
Timed methods are run sequentially. In the representation-comparison and
support-size runtime studies, MATLAB is restricted to one computational thread,
while MOSEK retains its default thread settings. The operations included in each
timing are specified with the respective experiment.

\subsection{Optimization and Initialization}
\label{sec:algorithm_experiments}

\subsubsection{Controlled Two-Dimensional Comparisons}
\label{sec:two_dimensional_optimizer_study}
To separate optimization from estimation, we use known GMMs for six subjects,
three in each class, with two components per subject. All nine between-class and
six within-class pairs are used. For projections from two dimensions to one,
\(a(\theta)=(\cos\theta,\sin\theta)^\top\), a grid of \(0.5^\circ\) over
\([0,180^\circ)\) first evaluates the exact objective at 360 directions.
We then refine every grid-local maximum by bounded one-dimensional optimization
between its two neighboring grid angles, treating the angular domain periodically.
The largest value among the grid and refined candidates is denoted by
\(F_{\mathrm{ref}}^*\); Appendix~\ref{ap:optimization_references} gives the details.
We report terminal regret
\[
  100\,\frac{\max\{0,F_{\mathrm{ref}}^*-F(a_{\mathrm{out}})\}}
  {F_{\mathrm{ref}}^*},
\]
where \(a_{\mathrm{out}}\) is the returned direction. The reference is numerical,
not an analytical global optimum; clipping allows an off-grid solution to exceed
it slightly. The small subject count is intentional: the aim is to compare
the optimizers' ability to maximize the same objective, rather than to assess
classification performance.

The first comparison uses 1,440 simulated datasets.
Each dataset contains the six subject GMMs generated for one parameter
combination and one random seed. The design crosses three class-mean offset levels
\(\{0.35,1,2\}\), three covariance scales \(\{0.05,0.5,2\}\), two anisotropy
levels \(\{1,9\}\), two orientation shifts \(\{0,\pi/4\}\), two component
imbalance levels \(\{0,0.25\}\), and 20 seeds. The same 20 seeds are reused
across the 72 parameter settings; Appendix~\ref{ap:optimization_gmm_2d}
specifies how they generate the mixture weights, means, and covariances.
All paths start at \(a_0=(1,0)^\top\) and allow at most 20 updates.

We compare the following optimizers: (i) gradient ascent; (ii) repeated geometric proposals;
(iii) repeated double-bound updates; and (iv) one geometric step followed by
gradient ascent.
A double-bound update separately backtracks the independent and geometric
proposals and takes the better accepted projection. Every acceptance uses the
transport-reoptimized objective. Optimizer timing includes setup, including the
geometric covariance cache when needed, but excludes GMM model fitting,
reference search, verification, and downstream classification.

\begin{table}[t]
\centering\small
\setlength{\tabcolsep}{4pt}
\begin{tabular}{lcccc}
\toprule
Optimizer & \makecell{Mean regret\\(\%)} & \makecell{90th percentile\\regret (\%)}
 & \makecell{Median total\\time (s)} & \makecell{Median time per\\attempted update (s)}\\
\midrule
Gradient & 0.2967 & $<10^{-4}$ & 0.0440 & 0.0022\\
Repeated geometric & 3.7834 & 11.6537 & 0.0025 & 0.0012\\
Repeated double-bound & 2.2077 & 4.4104 & 0.0050 & 0.0022\\
\makecell[l]{One geometric step,\\then gradient} & 0.4109 & $<10^{-4}$ & 0.0440 & 0.0022\\
\bottomrule
\end{tabular}
\caption{Matched optimizer comparison on 1,440 two-dimensional simulated datasets.
An attempted update is one outer iteration that constructs a proposed move,
including all its backtracking trials.
A double-bound update includes computing both proposals, and a rejected initial geometric
proposal followed immediately by a gradient fallback counts as one update.
The last column is the median of the per-dataset total optimizer time divided by
the number of attempted updates (using divisor one if none is attempted), so setup
time is included. Initialization screening is excluded to isolate the update rule.}
\label{tab:optimizer-oracle-rev3}
\end{table}

Table~\ref{tab:optimizer-oracle-rev3} shows that gradient ascent alone
achieves the lowest mean terminal regret. Adding one geometric spectral
proposal before gradient ascent does not improve this average:
mean regret is \(0.4109\%\), compared with \(0.2967\%\) for gradient
ascent alone. Both methods achieve nearly zero regret on at least
90\% of the simulated datasets; a small fraction of poorer outcomes
accounts for their positive mean regret.
Using geometric or double-bound proposals throughout optimization
finishes faster but produces larger terminal regret. These shorter
runtimes partly reflect earlier termination, rather than uniformly
cheaper iterations.

To compare the two spectral proposals in one update of the projection, we start each simulated dataset
at the same direction \(a_0\) and compute its optimal component couplings.
Holding these couplings fixed, we construct the independent and geometric
proposals and evaluate both using the exact Gaussian Wasserstein costs.
Relative to the largest fixed-coupling ratio found on a dense angular grid,
the mean percentage shortfall is \(17.21\%\) for the geometric proposal
and \(33.76\%\) for the independent proposal.

We then reoptimize all component couplings at each proposed direction
and compare the resulting transport-reoptimized objective with its value at \(a_0\).
Taking the full proposed move would decrease this objective in
\(29.58\%\) of datasets for the geometric proposal and \(53.82\%\) for
the independent proposal. These comparisons precede backtracking and
safeguarding. They favor the geometric proposal, but also show why
neither proposal should be accepted without checking the
transport-reoptimized objective.

\subsubsection{Higher Dimensions and a Short Iteration Budget}
\label{sec:highdim_optimizer_study}
For higher dimensions, we use a matched-budget optimization comparison: pilot
runs set method-specific iteration caps to approximate a common runtime target
within each \((d,q)\) setting. Early stopping can leave part of a budget unused,
so this does not enforce equal realized runtimes. The evaluation comprises 192
simulated datasets: \(d\in\{8,32,64\}\), \(q\in\{1,3\}\), diagonal or full
covariances, and two class-mean offset levels \(\{1,2\}\), with eight random-seed
replicates per parameter setting. Each replicate generates six subject GMMs,
three per class and three components per subject. Seeds 1 and 2 are used only for
calibration; seeds \(3\sim10\) generate the evaluation datasets and are reused across
parameter settings. Within a dataset, the methods use the same GMMs and initial
subspace. Appendix~\ref{ap:optimization_gmm_highd} specifies this generator,
which differs from the two-dimensional design in Section~\ref{sec:two_dimensional_optimizer_study}; Appendix~\ref{ap:optimization_budgets}
gives the calibration procedure and iteration caps.

Gradient ascent attains a larger terminal objective than repeated double-bound
updates on 176 of the 192 datasets. Relative to the gradient optimizer's terminal objective in
each dataset, double-bound is lower by \(14.82\%\) on average.

We next examine whether an initial geometric step helps when only a few
optimization updates are permitted. We use 20 simulated datasets with ambient dimension
\(d=8\), described in Appendix~\ref{sec:sim}. For each dataset, we consider
three fitted GMM component counts, \(K\in\{3,4,6\}\), and three retained
dimensions, \(q\in\{1,2,3\}\). This gives \(20\times3\times3=180\)
combinations of dataset, component count, and retained dimension. For each dataset, all optimizers and all nine \((K,q)\)
combinations use the same preselected training subset: 45 of its 50 subjects,
obtained from one split of the ten-fold partition.

With at most four updates after initialization, the mean relative
gain in the terminal objective between one geometric step followed
by gradient ascent and gradient ascent alone is \(5.80\%\), using
the latter as a reference. The former achieves a larger
terminal objective in 158 of the 180 matched combinations.
Its median paired runtime ratio is
1.107. Thus the objective gain costs a median 10.7\% more optimization time.
Replacing the geometric step with one double-bound update gives the same
gain (\(5.80\%\)); two double-bound updates give \(5.91\%\),
at runtime ratios 1.207 and 1.691, respectively. The second update yields negligible extra gain at the cost of considerably longer runtime. 
We therefore retain only one geometric step in Screened GeoGrad. In the corresponding ten-fold
classification experiment, the geometric-step method has mean AUC 0.9879 versus
0.9911 for gradient alone. This shows that improving the training ratio does not guarantee an
improvement in prediction. 

\subsubsection{Initialization and Screened GeoGrad}
\label{sec:initialization_diagnostics_short}
This experiment assesses how the initial projection affects the terminal gradient
solution and whether screening several candidates reduces sensitivity to the
starting point. A separate 1,920-dataset study adds class-mean offset 0.1 to the
two-dimensional factorial design of data and uses new seeds, a \(0.5^\circ\) reference
grid without local refinement, and at most 50 updates. Its generation and
initialization details are in Appendix~\ref{ap:optimization_gmm_2d}.
It compares the initialization families in Section~\ref{sec:initialization_modes}:
(i) ambient-coupling; (ii) equal-subject pooled-moment PCA; (iii) subject-mean LDA;
and (iv) eight seeded random directions. The screening policy evaluates the exact
initial ratio at all 11 candidates and optimizes only from the candidate with the
largest initial value. For comparison, a \emph{full multistart} policy, a more expensive procedure, optimizes all 11
paths and selects the largest terminal value.

\begin{table}[t]
\centering\small
\begin{tabular}{lR{0.7299}R{100.00}}
\toprule
Initialization and optimizer & \multicolumn{1}{c}{Mean regret (\%)} & \multicolumn{1}{c}{Within 5\% regret (\%)}\\
\midrule
Ambient + gradient & 0.7299 & 96.61\\
PCA + gradient & 1.1191 & 95.89\\
LDA + gradient & 0.5614 & 97.34\\
Screen all 11 + gradient & 0.1790 & 98.91\\
Screened GeoGrad & 0.2705 & 98.39\\
Full multistart gradient (11 paths) & 0.0015 & 100.00\\
\bottomrule
\end{tabular}
\caption{Initialization and complete-algorithm validation on 1,920 simulated datasets.
Full multistart runs gradient ascent from all 11 initial candidates and retains
the largest terminal objective, using substantially more computation than the
single-path schemes.}
\label{tab:gradient-initialization-rev3}
\end{table}

With the ambient-coupling initialization, gradient ascent reaches a
final objective within \(5\%\) of the numerical reference in \(96.61\%\)
of datasets. This proportion rises to \(98.91\%\) when we first screen
11 initial directions and run gradient ascent only from the one with
the largest initial Fisher ratio
(Table~\ref{tab:gradient-initialization-rev3}).
This improvement supports screening as a way to reduce sensitivity
to initialization without optimizing multiple complete paths.

Full multistart further improves
the terminal objectives, with all 1,920 solutions within \(5\%\) regret,
at the cost of optimizing all 11 paths rather than one.
Screened GeoGrad attains regret below \(5\%\) in \(98.39\%\) of simulated datasets.
Of the 1,920 datasets, its screening step selects the ambient-coupling
initialization in 484 cases, PCA in 269, LDA in 334, and one of the eight random
candidates in 833. These counts identify the candidate with the largest initial
transport-reoptimized Fisher ratio, before any geometric or gradient update;
they do not indicate which candidate would give the largest final objective
after optimization. The geometric step is rejected in \(23.44\%\) of datasets.

Taken together, these experiments support initialization screening and a gradient
optimization phase. 
The benefit of one safeguarded geometric proposal step is more specific: it can
improve the final objective when only a few gradient updates are allowed. Thus these results
do not establish a consistent advantage of Screened GeoGrad over screened gradient
ascent. We nevertheless use Screened GeoGrad in the subsequent experiments for
the practical reasons discussed in Section~\ref{sec:initialization_modes}; its
geometric step remains optional.

\subsection{Representation Accuracy and Computational Tradeoffs}
\label{sec:representation_diagnostics_rev3}
We compare GMM and subsampled empirical inputs in terms of total computation
time and the population Fisher ratios (defined in Eq.~\eqref{eq:representation_population_ratio}) attained by their learned projections.
Each dataset contains six subjects, three per class, whose true distributions
are three-component GMMs. We use the higher-dimensional generating family from
Section~\ref{sec:highdim_optimizer_study}, but place the mean signal in the first
two coordinates: the two classes receive opposite shifts
\(\pm(e_1+0.25e_2)\), where \(e_j\) is the \(j\)th coordinate vector.
Each subject supplies \(N_{\mathrm{cell}}=1536\) sampled cells. We consider
\(d\in\{16,64,128,256,512\}\) and diagonal or full generating covariances.
For each dimension and covariance structure, we generate three datasets using
different random seeds. All six representation choices below are evaluated on
these same three datasets. Appendix~\ref{ap:representation_simulation} gives the data
generation and model fitting details. This is a representation and optimization
comparison, not a classification experiment.

For each dataset, we construct six representations. The three GMM choices
use \(K=2,3,6\) components, respectively, and all available cells. For each
subject, we apply \(k\)-means with five restarts, then estimate component
weights, means, and regularized covariances from the resulting clusters. The diagonal or
full covariance structure is supplied to match the generating setting.
The three empirical choices retain \(m=32,64,128\) cells, respectively, with
equal masses, using nested subsets of the same cell sample. Under all six representations, we
use Screened GeoGrad with \(q=1\), all nine between-class and six within-class
pairs, the same set of random initialization seeds, and at most nine updates after
initialization. 

Table~\ref{tab:geograd_repr_time} reports representation-construction time
(Rep.), the complete optimizer time (Opt.), and their sum (Total).
Representation construction includes GMM fitting or empirical subsampling,
together with input data formatting. Optimization includes initialization,
screening, the geometric step, gradient
updates, and objective checks. Synthetic-data generation, evaluation on the
true distributions, classification, and file operations are excluded.

\begin{table}[p]
\centering\small
\setlength{\tabcolsep}{3pt}
\begin{tabular}{R{512}llR{0.0774}R{0.135}R{0.162}R{0.000271}R{0.000183}R{0.000291}}
\toprule
\multicolumn{1}{c}{$d$} & Covariance & Stage & \multicolumn{1}{c}{\makecell{GMM\\$K=2$}} & \multicolumn{1}{c}{\makecell{GMM\\$K=3$}} & \multicolumn{1}{c}{\makecell{GMM\\$K=6$}} & \multicolumn{1}{c}{\makecell{Empirical\\$m=32$}} & \multicolumn{1}{c}{\makecell{Empirical\\$m=64$}} & \multicolumn{1}{c}{\makecell{Empirical\\$m=128$}}\\
\midrule
16 & Diagonal & Rep. & 0.0774 & 0.135 & 0.162 & 0.000271 & 0.000183 & 0.00018\\
 &  & Opt. & 0.276 & 0.434 & 0.533 & 2.53 & 9.46 & 29.1\\
 &  & Total & 0.353 & 0.568 & 0.695 & 2.53 & 9.46 & 29.1\\
\midrule
16 & Full & Rep. & 0.0698 & 0.102 & 0.155 & 0.000264 & 0.000234 & 0.000291\\
 &  & Opt. & 0.167 & 0.323 & 0.523 & 2.61 & 8.27 & 29.7\\
 &  & Total & 0.237 & 0.425 & 0.678 & 2.61 & 8.27 & 29.7\\
\midrule
64 & Diagonal & Rep. & 0.19 & 0.245 & 0.363 & 0.000673 & 0.000134 & 0.000255\\
 &  & Opt. & 0.345 & 0.536 & 0.904 & 3.66 & 11.3 & 38.3\\
 &  & Total & 0.535 & 0.781 & 1.27 & 3.66 & 11.3 & 38.3\\
\midrule
64 & Full & Rep. & 0.206 & 0.269 & 0.385 & 0.000973 & 0.000121 & 0.000416\\
 &  & Opt. & 0.319 & 0.624 & 1.63 & 3.67 & 11.1 & 40.2\\
 &  & Total & 0.525 & 0.893 & 2.01 & 3.67 & 11.1 & 40.2\\
\midrule
128 & Diagonal & Rep. & 0.337 & 0.471 & 0.683 & 0.000214 & 0.000665 & 0.00117\\
 &  & Opt. & 0.462 & 0.929 & 2.36 & 7.07 & 17.5 & 55.8\\
 &  & Total & 0.799 & 1.4 & 3.05 & 7.07 & 17.5 & 55.8\\
\midrule
128 & Full & Rep. & 0.375 & 0.507 & 0.738 & 0.00064 & 0.00096 & 0.000767\\
 &  & Opt. & 0.786 & 1.72 & 5.57 & 6.95 & 16.9 & 54.5\\
 &  & Total & 1.16 & 2.22 & 6.31 & 6.95 & 16.9 & 54.5\\
\midrule
256 & Diagonal & Rep. & 0.72 & 1.16 & 1.27 & 0.00277 & 0.0014 & 0.00412\\
 &  & Opt. & 1.66 & 3.62 & 12 & 5.3 & 37.2 & 105\\
 &  & Total & 2.38 & 4.78 & 13.2 & 5.31 & 37.2 & 105\\
\midrule
256 & Full & Rep. & 0.759 & 1.05 & 1.61 & 0.00752 & 0.000741 & 0.00479\\
 &  & Opt. & 4.11 & 8.91 & 34.4 & 5.13 & 41.1 & 102\\
 &  & Total & 4.87 & 9.95 & 36 & 5.14 & 41.1 & 102\\
\midrule
512 & Diagonal & Rep. & 1.39 & 1.96 & 2.85 & 0.000645 & 0.0112 & 0.0054\\
 &  & Opt. & 10.4 & 22.5 & 81.9 & 16.6 & 52.6 & 329\\
 &  & Total & 11.8 & 24.5 & 84.7 & 16.6 & 52.6 & 329\\
\midrule
512 & Full & Rep. & 2.02 & 2.76 & 4.27 & 0.00149 & 0.0097 & 0.00606\\
 &  & Opt. & 27.1 & 60.3 & 215 & 18.2 & 56.4 & 338\\
 &  & Total & 29.1 & 63.1 & 219 & 18.2 & 56.4 & 338\\
\bottomrule
\end{tabular}
\caption{Representation and optimization time at $N_{\mathrm{cell}}=1536$ cells per subject. Rep., Opt., and Total are mean representation-construction time, complete Screened GeoGrad runtime, and their sum (seconds per six-subject dataset; three generated datasets per setting). Construction includes GMM fitting or empirical subsampling and arranging the resulting representations for the optimizer. The retained dimension is $q=1$ with at most ten recorded projections.}
\label{tab:geograd_repr_time}
\end{table}

Compact GMMs can substantially reduce total time despite the fitting cost.
At \(d=128\) with full covariances, \(K=3\) takes a mean 2.22 seconds per
six-subject dataset, compared with 54.5 seconds for \(m=128\).
At \(d=512\), the corresponding times are 63.1 and 338 seconds, but retaining
only \(m=32\) empirical supports takes 18.2 seconds. GMMs are therefore not
uniformly faster. A separate diagnostic in Appendix~\ref{ap:gaussian_kernel_timing}
examines which individual operations become expensive as the ambient dimension
increases. Table~\ref{tab:high_dimension_kernels} in the appendix shows that, at \(d=512\),
constructing Gaussian costs takes a median 6.81 seconds, versus 0.00531 seconds
for the corresponding transport LPs. This explains why a small component count
does not by itself make high-dimensional covariance calculations inexpensive.

To compare the learned projections, we evaluate them on the same true
distributions. Let \(\mathcal G_k^0\) be subject \(k\)'s generating GMM (true distribution).
For any unit direction \(a\), define its population Wasserstein Fisher ratio by
\begin{equation}
F_{\mathrm W}^{0}(a)=
\frac{|\mathcal I_B|^{-1}\displaystyle\sum_{(k,\ell)\in\mathcal I_B}
 W^2(a_{\#}\mathcal G_k^0,a_{\#}\mathcal G_\ell^0)}
{|\mathcal I_W|^{-1}\displaystyle\sum_{(k,\ell)\in\mathcal I_W}
 W^2(a_{\#}\mathcal G_k^0,a_{\#}\mathcal G_\ell^0)}.
\label{eq:representation_population_ratio}
\end{equation}
Here \(a_{\#}\) projects by \(x\mapsto a^\top x\). Define
\(F_{\mathsf{gmm}}^{0}(a)\) by replacing each squared Wasserstein distance in
the above ratio with squared MAW between the same projected generating GMMs.
These are the population objectives \(r_{\mathsf{emp}}^0\) and
\(r_{\mathsf{gmm}}^0\), respectively, from
Section~\ref{sec:gmm_estimation_perturbation}.

The objective values attained during optimization are not directly comparable across representations: they depend on different estimated distributions and different distances, namely $W_2$ for empirical measures and MAW for fitted GMMs. We therefore evaluate every learned direction using both population ratios defined above, each providing a common criterion for comparing the two representations.
Because every learned direction projects
the distributions to one dimension, their ordinary Wasserstein distances can be
evaluated numerically from the known generating mixtures;
Appendix~\ref{ap:population_direction_evaluation} gives the formula and numerical
procedure. 

\begin{table}[t]
\centering
\small
\setlength{\tabcolsep}{3pt}
\begin{tabular}{R{512}llR{126.6}R{148.1}R{140.8}R{127.6}R{152.1}R{154.2}}
\toprule
\multicolumn{1}{c}{$d$} & Covariance & Target & \multicolumn{1}{c}{\makecell{GMM\\$K=2$}} & \multicolumn{1}{c}{\makecell{GMM\\$K=3$}} & \multicolumn{1}{c}{\makecell{GMM\\$K=6$}} & \multicolumn{1}{c}{\makecell{Empirical\\$m=32$}} & \multicolumn{1}{c}{\makecell{Empirical\\$m=64$}} & \multicolumn{1}{c}{\makecell{Empirical\\$m=128$}}\\
\midrule
16 & Diagonal & $F_{\mathsf{gmm}}^{0}$ & 126.6 & 148.1 & 140.8 & 127.6 & 152.1 & 154.2\\
 &  & $F_{\mathrm W}^{0}$ & 434.8 & 446.5 & 285.4 & 327.6 & 254.3 & 372.1\\
\midrule
16 & Full & $F_{\mathsf{gmm}}^{0}$ & 85.13 & 80.45 & 85.26 & 84.48 & 86.18 & 83.77\\
 &  & $F_{\mathrm W}^{0}$ & 131.2 & 125.7 & 128.8 & 126 & 128.4 & 125.1\\
\midrule
64 & Diagonal & $F_{\mathsf{gmm}}^{0}$ & 144.9 & 149.6 & 148.3 & 153.4 & 153.6 & 155.3\\
 &  & $F_{\mathrm W}^{0}$ & 400.4 & 397.3 & 390.6 & 278.4 & 279.3 & 292.6\\
\midrule
64 & Full & $F_{\mathsf{gmm}}^{0}$ & 166.5 & 160 & 149.8 & 142.1 & 161.3 & 161.2\\
 &  & $F_{\mathrm W}^{0}$ & 509.3 & 464.6 & 387.2 & 348.1 & 434.7 & 417.7\\
\midrule
128 & Diagonal & $F_{\mathsf{gmm}}^{0}$ & 149.1 & 152.3 & 155.2 & 153.6 & 154.3 & 154.6\\
 &  & $F_{\mathrm W}^{0}$ & 334.4 & 358.6 & 376.1 & 285.5 & 284.1 & 347.8\\
\midrule
128 & Full & $F_{\mathsf{gmm}}^{0}$ & 140.4 & 144.6 & 144.3 & 117 & 133.7 & 140.6\\
 &  & $F_{\mathrm W}^{0}$ & 324.6 & 323.3 & 327.4 & 235.9 & 288.4 & 295.7\\
\midrule
256 & Diagonal & $F_{\mathsf{gmm}}^{0}$ & 135.2 & 135.8 & 152.3 & 142.1 & 147 & 150.5\\
 &  & $F_{\mathrm W}^{0}$ & 339.3 & 395.4 & 366.9 & 262.3 & 297.3 & 288.9\\
\midrule
256 & Full & $F_{\mathsf{gmm}}^{0}$ & 244 & 243.3 & 243.2 & 118.5 & 194.4 & 223.4\\
 &  & $F_{\mathrm W}^{0}$ & 1578 & 1559 & 1558 & 305.5 & 890 & 1208\\
\midrule
512 & Diagonal & $F_{\mathsf{gmm}}^{0}$ & 125.4 & 149.9 & 150.4 & 101.1 & 147.2 & 137.1\\
 &  & $F_{\mathrm W}^{0}$ & 309.3 & 323.7 & 297.1 & 250.3 & 275.6 & 336.6\\
\midrule
512 & Full & $F_{\mathsf{gmm}}^{0}$ & 144.2 & 144.1 & 144.4 & 139.1 & 136.7 & 131.4\\
 &  & $F_{\mathrm W}^{0}$ & 334.4 & 335.4 & 326.6 & 316.8 & 307.8 & 260.8\\
\bottomrule
\end{tabular}
\caption{Population Fisher ratios at learned directions. Columns identify the representations used to learn the directions. Every direction, including those learned from empirical inputs, is evaluated on the same true generating GMMs. The entries are medians over the three generated datasets per setting: $F_{\mathsf{gmm}}^{0}$ uses MAW, whereas $F_{\mathrm W}^{0}$ uses ordinary projected Wasserstein distance.}
\label{tab:geograd_population_values}
\end{table}

Table~\ref{tab:geograd_population_values} reports these two ratios at the
returned directions for all five ambient dimensions. Larger values are better
within a row; the two population objectives need not rank directions alike.
Several patterns emerge. Although the generating distributions have three
components, using \(K=2\) or \(6\) does not necessarily lower
\(F_{\mathsf{gmm}}^0\): at least one of these choices exceeds \(K=3\) in
seven of the ten dimension--covariance settings. Directions learned from empirical
inputs can also be competitive on this MAW target; for example, \(m=128\)
exceeds \(K=3\) in six settings. 

GMM inputs also perform well when evaluated using ordinary population Wasserstein
distance rather than MAW. The \(K=3\) directions give higher
\(F_{\mathrm W}^0\) than \(m=64\) in nine of the ten settings, with gains
of approximately \(7\sim76\%\) in those nine settings. They also exceed \(m=128\)
in nine settings, while the total computation time for \(m=128\) is \(5.4\sim70\)
times that for \(K=3\) across the ten settings
(Table~\ref{tab:geograd_repr_time}). Thus compact GMM
representations can provide competitive, and sometimes substantially higher,
population Wasserstein ratios at considerably lower computational cost.

For empirical inputs, increasing the support size from \(m=32\) to \(128\)
improves both population ratios in eight of the ten settings. Improvements over
\(m=64\) are less consistent: \(m=128\) gives a higher ratio in six settings
for each population objective, while requiring \(2.5\sim6.3\) times as much total
computation time. Increasing support size therefore has a substantial cost and
does not guarantee a better learned projection. The objective comparisons use
medians over the three generated datasets per setting, and the runtime comparisons
use means over those same datasets. With only three replicates, these summaries do not
establish statistically significant differences.

The results are specific to the clustering-based GMM estimator and the budget
of at most nine OTAF updates after initialization. Each GMM is estimated from
all 1536 cells, whereas each empirical representation retains only 32, 64, or
128 cells; the two representations therefore need not estimate the true
distribution equally accurately. The empirical optimizer solves each transport
problem using a general-purpose LP solver. Faster solvers that exploit the
one-dimensional projected distributions could reduce its runtime and change
the timing comparison.

\subsection{Classification Studies}
\label{sec:predictive_demonstrations}

\subsubsection{Three Real Datasets}
\label{sec:real}
We examine classification of three single-cell datasets, the pulmonary-fibrosis (PF), breast-cancer (BC),
and uveal-melanoma (UM) cohorts, and a simulation in Appendix~\ref{sec:sim}. 
\begin{enumerate}
\item
PF uses the SCG3A2+ representation from \citet{PFdata}, following the cell-type
analysis of \citet{zhang2022bsde}. The supplied cohort has 19 cases and 10 controls
and retains 30 highly variable genes ($d=30$, $n=29$); cell counts range from 1 to 663 per subject.
Subjects with fewer than ten cells are modeled by a single Gaussian. For a one-cell
subject, the stored covariance is estimated from ten Gaussian perturbations with
standard deviation 0.1, following \citet{https://doi.org/10.1111/biom.13630}.
\item
BC contains 26 subjects, 10 with triple-negative cancer and 16 with
estrogen-receptor-positive (ER+) or human-epidermal-growth-factor-receptor-2-positive
(HER2+) disease \citep{Wu2021}. BC retains 100 highly variable
genes ($d=100$, $n=26$).
\item
UM contains 11 subjects, three in low-risk and eight in
high-risk gene-expression-profiling (GEP) classes \citep{Durante2020}.  UM also retains 100 highly variable genes ($d=100$, $n=11$). In such a small cohort, point estimates of performance are highly uncertain.
\end{enumerate}

For the real datasets, the supplied GMMs use two constructions. GMM-C-\(K\) clusters pooled cells, then computes
each subject's component weights, means, and covariances within the aligned
clusters. GMM-S-\(K\) clusters each subject separately and needs no component
alignment. Both use the number of components \(K\in\{3,5,7,10\}\), with fewer components
when cells are insufficient. 

For performance evaluation, the leave-one-subject-out approach is used because of small data sizes. Screened GeoGrad
comparisons use GMM-S-7 with subject pair selection and use at most ten recorded projections, including initialization.

\subsubsection{Downstream Classifiers}
\label{sec:downstream}
All the downstream classifiers take as input distances between distributions (MAW for GMMs, $W_2$ for empirical measures). When ambient dimension reduction is applied, pushforward distributions are used. Our default classifier is PMM, the distance-based pseudo-mixture classifier of
\citet{qiao2016distance}. We also compare three alternatives below.

\paragraph{PMM (default):}
Denote the density model for class $l$ by $\mathcal{M}_l$, $l=1, ..., \mathrm{m_c}$. Let \(\mathcal J_l\) be the set of indices for training subjects in class \(l\). Let $\mathcal{G}$ be the distribution of an input subject, and $\mathcal{G}_j$, $j=1, ..., n$, be the distributions of training subjects. Consider the case of MAW distance for GMMs below (replace it by $W_2$ for empirical measures). PMM's class-wise pseudo-density is
\begin{equation}
\psi(\mathcal G\mid\mathcal M_l)
=\frac{1}{|\mathcal J_l|}\sum_{j\in\mathcal J_l}
\left(\frac{1}{\sqrt{\pi \theta_b}}\right)^{2\theta_{s}}
\exp\{-\widetilde W^2(\mathcal G,\mathcal G_j)/\theta_{b}\}.
\label{eq:pseudo}
\end{equation}
The shape and scale parameters \(\theta_s,\theta_b\) are estimated from training data
as in \citet{qiao2016distance}. With the class prior estimated by empirical frequencies, the Bayes formula provides the class posterior. 
In leave-one-subject-out
evaluation, the training prior of the held-out subject's own class is reduced,
which can depress AUC when scores from different folds are pooled; below-chance
AUC therefore need not indicate a reversed projection. A sensitivity
analysis confirms a substantial effect in BC and UM
(Appendix~\ref{ap:pmm_prior_sensitivity}). We retain the stated prior-weighted
scoring rule throughout the main figures and tables.

\paragraph{Nearest-neighbor and radial alternatives:}
\begin{enumerate}
\item
The \emph{1-NN} classifier assigns the class of the nearest training subject.
For AUC, it uses the continuous score \(D_{1, \mathrm{min}}/(D_{1, \mathrm{min}}+D_{2,\mathrm{min}})\), where \(D_{l,\mathrm{min}}\)
is the nearest squared distance to class \(l=1, 2\); a small positive numerical
offset handles zero distances. Thresholding at 0.5 gives the nearest-class decision.
\item
\emph{Tuned \(k\)-NN} computes a weighted average of the binary labels
of the \(k\) nearest training subjects, with
\(k\in\{1,3,5,7\}\) when feasible. Each neighbor receives weight
proportional to \(1/D\), where \(D\) is its squared distance from
the subject being classified.
\item
The \emph{radial} classifier averages
\(\exp(-D/\Delta_{\mathrm{bw}})\) within each class and normalizes the two class averages into a
score. 
The bandwidth $\Delta_{\mathrm{bw}}$ candidates are \(\{1/4,1/2,1,2,4\}\) times the median
positive pairwise distance among training subjects, computed
using the representation and projection supplied to the classifier.
\end{enumerate}

We use outer leave-one-subject-out evaluation with inner tuning of
classifier hyperparameters. In each outer fold, one subject forms the
test set and the remaining subjects form the training set. Within this
training set, leave-one-subject-out evaluation selects hyperparameters such as
\(k\) or \(\Delta_{\mathrm{bw}}\), maximizing AUC and breaking ties by
higher balanced accuracy, then by the smaller parameter value.
The projection is learned once from the outer training set and held
fixed throughout inner tuning; only the classifier is refitted in each
inner split. The outer test subject is excluded from projection learning
and hyperparameter selection.
We compute AUC from the pooled outer-test scores and accuracy from the
corresponding class decisions. Both measures lie in \([0,1]\), with
larger values indicating better performance. PMM is used unless stated
otherwise.

\subsubsection{Baseline Classification Methods}
\label{sec:leakage_clean_classification}

The baseline methods listed below serve three purposes: assessing the effectiveness of ambient-space dimension reduction for downstream classification, comparing our proposed method with an existing distribution-based classification approach that also uses GMMs, and comparing it with vector-based methods (hereafter, vector baselines). We refer to methods that rely on distances between distributions as distribution baselines. 
\begin{enumerate}
\item
\textbf{Distribution baselines:} 
These methods differ in how they reduce the ambient dimension or compute distances between distributions. Each is paired with every downstream classifier described in Section~\ref{sec:downstream}.
\begin{enumerate}
\item Original ambient space (Orig.).
    \item Pooled-moment PCA (PCA): the projection method used in initialization (ii) of Section~\ref{sec:initialization_modes}.
    \item Subject-mean LDA with pooled-PCA completion (LDA): the projection method used in initialization (iii) of Section~\ref{sec:initialization_modes}.
    \item QOT: an external distribution-comparison method
\citep{10.1093/bib/bbae713}, evaluated using the authors' public implementation.\footnote{\url{https://github.com/PennShenLab/QOT}, revision \texttt{28cd529}.}
QOT-Full transports Gaussian components using their squared Wasserstein costs.
On the supplied GMMs, this is the same distance construction as unprojected
MAW, already reported as Orig. 
QOT-Cosine instead uses $1-\mu_i^\top\mu_j/(\|\mu_i\|\|\mu_j\|)$ as the
component cost, where $\mu_i$ and $\mu_j$ are the mean vectors of the two Gaussian components. It produces subject-level
dissimilarities without learning a projection. In the results below, QOT denotes QOT-Cosine.
Since the supplied data contain fitted GMMs, not raw cells, we apply QOT's
distance stage directly to GMM-S-7, without reproducing its cell-type-specific
fitting or preprocessing. We use the official \texttt{exact} and
\texttt{cosine} options without cost-matrix normalization.
\end{enumerate}

\item 
\textbf{Vector baselines:} We tune hyperparameters and network architectures as described below.
\begin{enumerate}
\item Random forest (RF).
    \item Logistic regression (LR).
    \item Support vector machine (SVM).
    \item MLP neural network.
\end{enumerate}
\end{enumerate}

To avoid favoring distributional methods through weak vector features, we form
five permutation-invariant, mixture-wide feature vectors from GMM-S-7. These
summarize the subject's mixture as a whole, without requiring aligned components.
Let \(\bar\mu^{(k)}\) and \(\bar\Sigma^{(k)}\) be the overall mixture mean and
covariance defined in Eq.~\eqref{eq:subject-moments-initialization}.
The three moment-based feature vectors are
\[
\text{Mean-only:}\quad \bar\mu^{(k)},
\qquad
\text{Moment-diag:}\quad
\begin{bmatrix}
\bar\mu^{(k)}\\
\operatorname{diag}(\bar\Sigma^{(k)})
\end{bmatrix},
\qquad
\text{Moment-full:}\quad
\begin{bmatrix}
\bar\mu^{(k)}\\
\operatorname{svec}(\bar\Sigma^{(k)})
\end{bmatrix}.
\]
Here \(\operatorname{diag}\) extracts the diagonal as a vector.
\emph{Symmetric vectorization}, \(\operatorname{svec}\), stacks the
upper-triangular entries columnwise, retaining diagonal entries and multiplying
off-diagonal entries by \(\sqrt{2}\). This scaling makes the vector's Euclidean
norm equal to the matrix's Frobenius norm.

The other two feature families use the characteristic function (CF). For a
frequency vector \(\mathbf t\in\mathbb R^d\), the GMM characteristic function is
\[
\chi^{(k)}(\mathbf t)=\sum_i p_i^{(k)}
\exp\!\left\{\mathrm{i}\mathbf t^\top\mu_i^{(k)}
-\tfrac12\mathbf t^\top\Sigma_i^{(k)}\mathbf t\right\},
\qquad \mathrm{i}^2=-1.
\]
CF-full stacks its real and imaginary parts at 64 fixed frequency vectors:
\[
\bigl[\operatorname{Re}\chi^{(k)}(\mathbf t_1),\ldots,
\operatorname{Re}\chi^{(k)}(\mathbf t_{64}),
\operatorname{Im}\chi^{(k)}(\mathbf t_1),\ldots,
\operatorname{Im}\chi^{(k)}(\mathbf t_{64})\bigr]^\top.
\]
CF-diag uses the same construction after replacing each component covariance
\(\Sigma_i^{(k)}\) in \(\chi^{(k)}\) by the diagonal matrix with the same
diagonal entries. This replacement acts on each component, not on the total
mixture covariance. The frequency vectors use seeded Gaussian directions,
normalized to unit length, with radii cycling through \(0.5,1,2\); the
same vectors are used for all subjects.

For Moment-diag, Moment-full, and the two CF families, these formulas are
evaluated after ambient
centering and diagonal rescaling using the pooled moments of the training subjects:
the center is the average subject mean, and each scale is the pooled standard
deviation. Mean-only uses
the overall subject mean directly. All families then undergo training-derived
feature standardization and principal-coordinate reduction.

The raw Mean-only dimension is 30 for PF and 100 for BC/UM. 
Moment-diag has 60 features for PF and 200 for BC and UM. Moment-full has 495 and 5,150 features, respectively. Both CF feature families have 128 features for every dataset. All coordinate scaling, standardization, and reduction to
at most eight principal coordinates are learned within training splits.

Three-fold inner validation tunes ridge LR, linear SVM, a 300-tree RF, and
two-layer ReLU MLPs with widths \(8\)--\(4\), \(32\)--\(16\),
\(128\)--\(64\), or \(256\)--\(128\). Selection prioritizes inner AUC, then balanced
accuracy, with remaining ties broken in favor of simpler models.
Across the five feature representations and the \(29+26+11=66\)
leave-one-out folds, there were \(5\times66=330\) selections of network
architecture. A network wider than \(8\)--\(4\) was selected
in 162 cases. Thus the MLP comparison is not
restricted to the smallest network.

\begin{table}[!p]
\centering\small
\setlength{\tabcolsep}{2pt}
\begin{tabular}{@{}cllR{.000}R{1.000}R{1.000}R{.000}R{1.000}@{\hspace{8pt}}lR{.000}R{.000}R{.000}R{.000}@{}}
\toprule
& & \multicolumn{6}{c}{Distribution-based classification} & \multicolumn{5}{c}{Vector-based classification}\\
\cmidrule(lr){3-8}\cmidrule(l){9-13}
& Data & Classifier & \multicolumn{1}{c}{OTAF} & \multicolumn{1}{c}{Orig.} & \multicolumn{1}{c}{PCA} & \multicolumn{1}{c}{LDA} & \multicolumn{1}{c@{\hspace{8pt}}}{QOT} & Features & \multicolumn{1}{c}{LR} & \multicolumn{1}{c}{SVM} & \multicolumn{1}{c}{RF} & \multicolumn{1}{c@{}}{MLP}\\
\midrule
\multirow{15}{*}{\rotatebox[origin=c]{90}{AUC}} & PF & 1-NN & .963 & .932 & .968 & .447 & .916 & Mean-only & .816 & .705 & .879 & .842\\
 &  & Tuned $k$-NN & .939 & .876 & .916 & .542 & .884 & Moment-diag & .905 & .900 & .932 & .921\\
 &  & Radial & .963 & .721 & .932 & .611 & .947 & Moment-full & .979 & .974 & .800 & .947\\
 &  & PMM & .953 & .932 & .926 & .647 & .937 & CF-diag & .637 & .779 & .900 & .816\\
 &  &  &  &  &  &  &  & CF-full & .568 & .558 & .842 & .626\\
\cmidrule(l){2-13}
 & BC & 1-NN & .700 & .713 & .688 & .694 & .700 & Mean-only & .556 & .469 & .363 & .450\\
 &  & Tuned $k$-NN & .575 & .584 & .509 & .566 & .694 & Moment-diag & .481 & .238 & .519 & .494\\
 &  & Radial & .650 & .681 & .700 & .700 & .694 & Moment-full & .500 & .606 & .613 & .431\\
 &  & PMM & .550 & .300 & .300 & .300 & .538 & CF-diag & .656 & .594 & .519 & .538\\
 &  &  &  &  &  &  &  & CF-full & .675 & .675 & .500 & .619\\
\cmidrule(l){2-13}
 & UM & 1-NN & .625 & 1.000 & 1.000 & .833 & 1.000 & Mean-only & .542 & .417 & .708 & .583\\
 &  & Tuned $k$-NN & .667 & .875 & .896 & .583 & 1.000 & Moment-diag & .958 & .917 & .833 & .958\\
 &  & Radial & .333 & .917 & 1.000 & .667 & 1.000 & Moment-full & .875 & .917 & .542 & .833\\
 &  & PMM & .875 & .250 & .125 & .708 & 1.000 & CF-diag & .875 & .833 & .625 & .583\\
 &  &  &  &  &  &  &  & CF-full & .833 & .792 & .875 & .500\\
\midrule\midrule
\multirow{15}{*}{\rotatebox[origin=c]{90}{Accuracy}} & PF & 1-NN & .793 & .759 & .862 & .586 & .862 & Mean-only & .828 & .759 & .897 & .759\\
 &  & Tuned $k$-NN & .862 & .655 & .793 & .586 & .655 & Moment-diag & .793 & .862 & .759 & .828\\
 &  & Radial & .931 & .655 & .862 & .586 & .862 & Moment-full & .966 & .966 & .759 & .862\\
 &  & PMM & .897 & .690 & .759 & .586 & .828 & CF-diag & .655 & .759 & .828 & .793\\
 &  &  &  &  &  &  &  & CF-full & .586 & .655 & .724 & .655\\
\cmidrule(l){2-13}
 & BC & 1-NN & .692 & .615 & .731 & .731 & .692 & Mean-only & .577 & .615 & .346 & .423\\
 &  & Tuned $k$-NN & .577 & .577 & .500 & .500 & .731 & Moment-diag & .538 & .423 & .538 & .462\\
 &  & Radial & .577 & .731 & .731 & .731 & .731 & Moment-full & .577 & .731 & .615 & .346\\
 &  & PMM & .731 & .654 & .615 & .615 & .731 & CF-diag & .615 & .731 & .615 & .615\\
 &  &  &  &  &  &  &  & CF-full & .654 & .731 & .538 & .615\\
\cmidrule(l){2-13}
 & UM & 1-NN & .818 & .818 & .818 & .727 & 1.000 & Mean-only & .636 & .636 & .727 & .636\\
 &  & Tuned $k$-NN & .818 & .727 & .818 & .727 & .909 & Moment-diag & .909 & .909 & .818 & .909\\
 &  & Radial & .818 & .909 & .545 & .727 & .909 & Moment-full & .818 & .818 & .636 & .636\\
 &  & PMM & .727 & .727 & .727 & .727 & .818 & CF-diag & .818 & .818 & .727 & .818\\
 &  &  &  &  &  &  &  & CF-full & .727 & .727 & .727 & .636\\
\bottomrule
\end{tabular}
\caption{Leave-one-subject-out AUC (upper block) and accuracy (lower block) on GMM-S-7.
PCA, LDA, and Screened GeoGrad use $q=1,40,1$ for PF, BC, and UM; OTAF uses confusing pairs.
Orig.\ denotes unprojected MAW and also represents QOT-Full, whose supplied-GMM distance agrees with it. QOT denotes QOT-Cosine, without projection.
Vector classifiers use mixture-wide features and at most eight training-derived
principal coordinates; MLP is tuned.
The left and right groups have independent row labels, not paired classifier--feature configurations.}
\label{tab:current_classifier_sensitivity}
\label{tab:covariance_vector_auc}
\label{tab:classification_accuracy}
\end{table}

\subsubsection{Classification Results}
\label{sec:classify_results}
Table~\ref{tab:current_classifier_sensitivity} presents the classification
results for OTAF, using Screened GeoGrad as the optimizer, and the baselines,
with AUC in the upper block and accuracy in the lower block. The left group
(distribution-based methods) varies the projection and downstream classifier;
the right group (vector baselines) varies the vector summary and classifier.
With PMM held fixed, OTAF has the highest reported AUC among the four
projection choices (OTAF, original space, PCA, and LDA) on all three cohorts.

The complete comparison also illustrates the importance of the downstream
classifier and representation. On PF, OTAF with the radial classifier achieves
an AUC of 0.963 and an accuracy of 0.931, while PCA with 1-NN achieves an AUC
of 0.968. On BC, OTAF achieves AUCs of 0.700 with 1-NN and 0.550 with PMM,
with corresponding accuracies of 0.692 and 0.731. QOT achieves higher AUC
and accuracy on BC than OTAF with tuned \(k\)-NN or the radial classifier,
and achieves an AUC of 1.000 with all four classifiers on UM. 
Strong vector classifiers can match or exceed distribution-based methods. Nevertheless,
the highest reported AUC among distribution-based methods exceeds that
among vector methods on BC (0.713 versus 0.675) and UM (1.0 versus 0.958), while being slightly lower on PF (0.968 versus 0.979).

For vector methods, Mean-only features generally perform less well than
Moment-diag or Moment-full features. Including off-diagonal covariance
entries is not uniformly beneficial: switching from Moment-diag to
Moment-full features increases the AUC of SVM on BC from 0.238 to 0.606
(paired-bootstrap 95\% interval for the AUC difference:
\((0.003,0.700)\)), but decreases the AUC of RF on PF from 0.932 to 0.800.
Richer CF features and wider networks are likewise not consistently better.

\paragraph{Representation and classifier choices in small, high-dimensional cohorts:}
\label{sec:mean_dimension}
Averaged across downstream classifiers, OTAF outperforms the other
distribution-based methods on PF in both AUC and accuracy, whereas QOT
performs best on BC and UM.
BC and UM each have 100 retained coordinates but only 26 and 11 subjects,
respectively. 
However, QOT uses cosine costs between component means and ignores component covariances, whereas the OTAF results above use GMM representations that retain covariance information. The weaker classification performance of OTAF on these two datasets therefore does not, by itself, imply that its dimension reduction is ineffective.

Two considerations motivate simpler choices of subject representation and classifier for small, high-dimensional datasets.
First, estimating a high-dimensional component covariance can be unreliable
when few cells contribute to that component. This depends on the within-subject
cell counts, whereas the small number of subjects limits supervised learning
and evaluation. Second, leave-one-subject-out evaluation changes the estimated
class priors, which can strongly affect pooled PMM scores
(Appendix~\ref{ap:pmm_prior_sensitivity}). We therefore examine 1-NN, whose
decisions and continuous AUC scores do not explicitly use estimated class
priors, on a representation that omits component covariances.

We replace each GMM-S-7 by $\sum_{i=1}^{7}p_i^{(k)}\delta_{\mu_i^{(k)}}$.
This weighted discrete distribution retains all component means and weights,
not just the overall subject mean. We use squared $W_2$ with squared
Euclidean ground costs, rather than QOT's cosine costs. Without dimension
reduction, this representation paired with 1-NN achieves AUC/accuracy
0.769/0.846 on BC and 1.000/1.000 on UM, compared with QOT+1-NN's
0.700/0.692 and 1.000/1.000 in
Table~\ref{tab:current_classifier_sensitivity}.
These results suggest that simplifying subject representations can be
beneficial in small, high-dimensional datasets.

\begin{figure}[!htb]
\centering
\includegraphics[width=\textwidth]{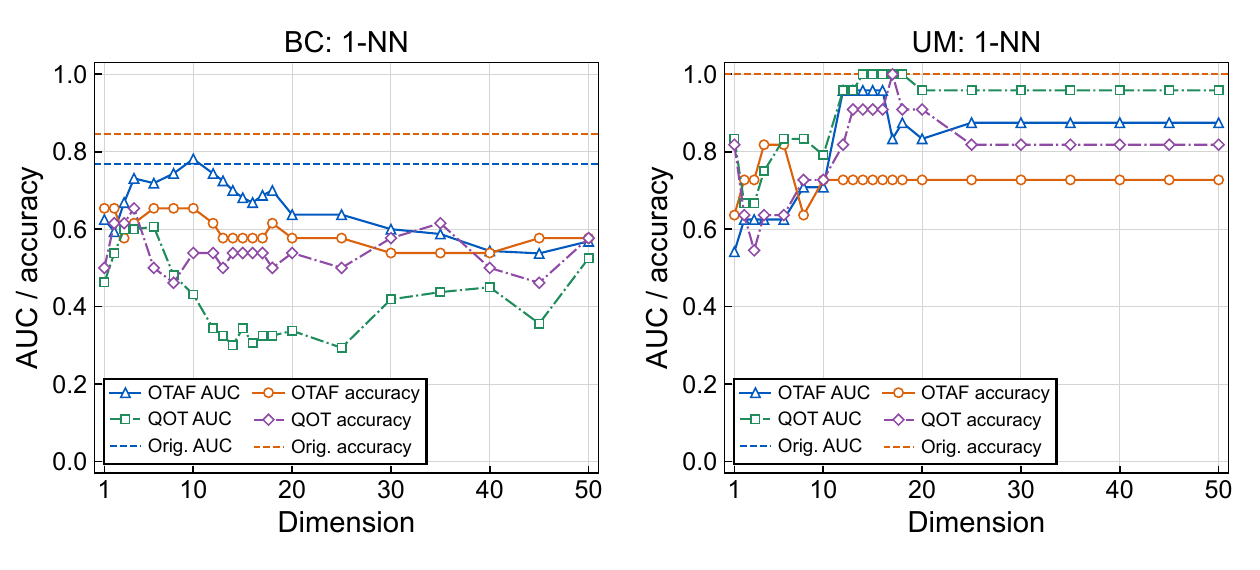}
\caption{AUC and accuracy of 1-NN using weighted component means:
BC (left) and UM (right). Solid OTAF and dash-dotted QOT curves use
squared-Euclidean and cosine component costs, respectively, at the same
Screened GeoGrad projections. Dashed Orig.\ lines are the unprojected
squared-Euclidean references.}
\label{fig:mean_dimension_1nn}
\end{figure}

Next, we examine whether OTAF can preserve classification performance in
substantially fewer dimensions using these simplified representations.
In Figure~\ref{fig:mean_dimension_1nn}, Screened GeoGrad is applied to the
weighted component means.
With squared-Euclidean transport, BC at dimension $q=10$ achieves an AUC
of 0.781, comparable to 0.769 without reduction, but accuracy falls from
0.846 to 0.654. For UM, dimensions \(q=12\sim16\) retain an AUC of 0.958,
compared with 1.000 without reduction, while accuracy falls from 1.000
to 0.727.
These results show that OTAF can retain competitive ranking performance,
as measured by AUC, in a substantially lower-dimensional space.

We also keep each fitted projection unchanged and replace only the downstream
component cost with QOT's cosine cost,
\(1-\cos(A^\top\mu_i,A^\top\mu_j)\), before applying the same 1-NN rule.
The additional curves in Figure~\ref{fig:mean_dimension_1nn} show a
dataset-dependent effect. On BC, cosine transport yields lower AUC at every
tested dimension; at \(q=10\), its AUC/accuracy is 0.431/0.538, compared
with 0.781/0.654 for squared-Euclidean transport. On UM, it achieves
AUC/accuracy 1.000/1.000 at \(q=17\), compared with 0.833/0.727 for
squared-Euclidean transport at the same dimension. These results suggest
that QOT's cosine cost is particularly suitable for UM: when used after
OTAF dimension reduction, it matches the classification performance
achieved in the original space.

\begin{figure}[!htb]
\centering
\captionsetup[subfigure]{skip=7pt}
\begin{subfigure}[t]{0.48\linewidth}
\centering
\includegraphics[width=\linewidth]{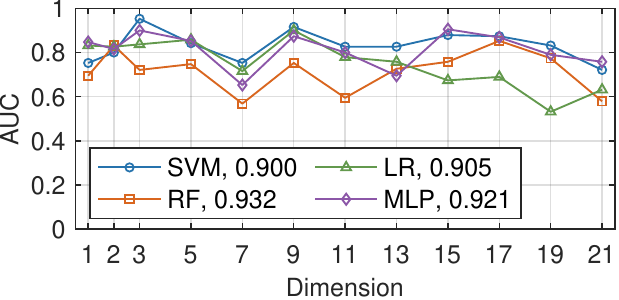}
\caption{PF: AUC}\label{fig:benchmark_ipf_auc}
\end{subfigure}\hfill
\begin{subfigure}[t]{0.48\linewidth}
\centering
\includegraphics[width=\linewidth]{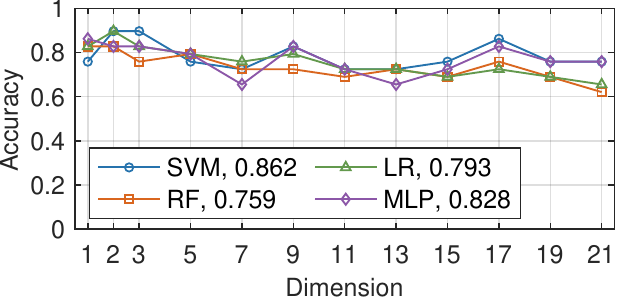}
\caption{PF: Accuracy}\label{fig:benchmark_ipf_acc}
\end{subfigure}
\par\vspace{14pt}
\begin{subfigure}[t]{0.48\linewidth}
\centering
\includegraphics[width=\linewidth]{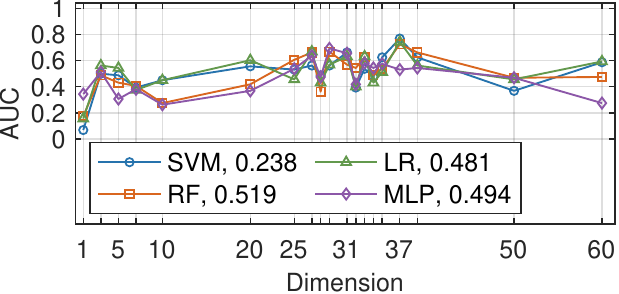}
\caption{BC: AUC}\label{fig:benchmark_bc_auc}
\end{subfigure}\hfill
\begin{subfigure}[t]{0.48\linewidth}
\centering
\includegraphics[width=\linewidth]{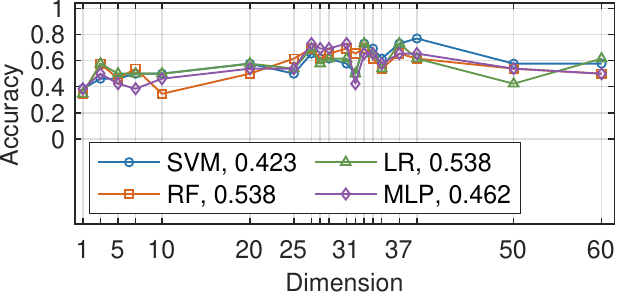}
\caption{BC: Accuracy}\label{fig:benchmark_bc_acc}
\end{subfigure}
\par\vspace{14pt}
\begin{subfigure}[t]{0.48\linewidth}
\centering
\includegraphics[width=\linewidth]{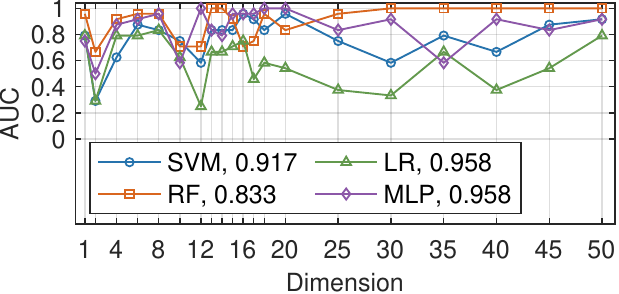}
\caption{UM: AUC}\label{fig:benchmark_um_auc}
\end{subfigure}\hfill
\begin{subfigure}[t]{0.48\linewidth}
\centering
\includegraphics[width=\linewidth]{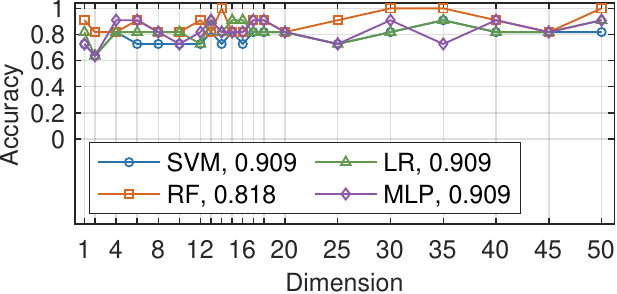}
\caption{UM: Accuracy}\label{fig:benchmark_um_acc}
\end{subfigure}
\caption{Screened GeoGrad followed by Moment-diag vector classification on
GMM-S-7: leave-one-subject-out AUC and accuracy versus projection dimension
\(q\). Each legend lists the classifier and its original-space Moment-diag
performance under the protocol in
Table~\ref{tab:current_classifier_sensitivity}.}
\label{fig:benchmark}
\end{figure}

\paragraph{OTAF combined with vector classification:}
A natural question is whether OTAF can serve as a general dimension-reduction
method, beyond its use with classifiers based on distances between distributions.
Figure~\ref{fig:benchmark} examines whether OTAF dimension reduction can improve
vector-based classification. We use GMM-S-7 and the Moment-diag features defined
above, applying the same construction to the projected mixture, whose overall
mean and covariance are \(A^\top\bar\mu^{(k)}\) and
\(A^\top\bar\Sigma^{(k)}A\). The resulting raw feature vector has \(2q\) entries;
the full covariance is projected before its diagonal is extracted.
The original-space references are the Moment-diag results in
Table~\ref{tab:current_classifier_sensitivity}.

We use the same training-tuned linear SVM, 300-tree RF, ridge LR, and two-layer
MLP protocol as in Table~\ref{tab:current_classifier_sensitivity}, including
training-derived feature preprocessing and reduction to at most eight principal
coordinates. OTAF uses the Screened GeoGrad optimizer. 
The results show that OTAF can benefit vector classifiers, although the effect
depends on the classifier and retained dimension. On BC, reducing the dimension
from 100 to \(q=37\) increases AUC for all four classifiers. SVM, RF, LR,
and MLP achieve AUCs of \(0.769\), \(0.725\), \(0.744\), and \(0.531\),
respectively, compared with \(0.238\), \(0.519\), \(0.481\), and \(0.494\)
without projection. At \(q=3\), AUC also increases for SVM, LR, and MLP,
but decreases slightly for RF, from \(0.519\) to \(0.488\). On UM, after
reduction to just one dimension, RF achieves AUC/accuracy \(0.958/0.909\),
compared with \(0.833/0.818\) in the original 100-dimensional space.
These examples illustrate that OTAF dimension reduction can improve
downstream classification even when paired with vector-based methods.

\paragraph{Additional sensitivity studies:}
Appendix~\ref{ap:real_data_sensitivity} separates five questions from the
central classification comparison. Appendix~\ref{ap:component_sensitivity}
varies the GMM construction and component count to assess sensitivity to
distribution representation; the performance on PF after applying OTAF varies relatively
little across the tested choices, without a monotone component-count effect.
Appendix~\ref{ap:dimension_sensitivity} varies the retained dimension for BC
and UM with PMM to show the consequences of compression. Appendix~\ref{ap:pair_sensitivity}
compares using only confusing pairs with using all pairs to assess whether focusing the
training objective on difficult subjects helps. Appendix~\ref{ap:representation_mismatch}
checks robustness when training and test subjects use different GMM
representations. Finally, Appendix~\ref{ap:pmm_prior_sensitivity} examines
how fold-dependent class priors affect pooled leave-one-subject-out PMM scores,
which is especially relevant for these small cohorts.

\section{Conclusion and Discussion}
\label{sec:discuss}

We have formulated supervised canonical-variate analysis for distribution-valued
observations by extending Fisher's between-to-within ratio to projected transport
distances. OTAF accommodates empirical measures and fitted GMMs within the same
dimension-reduction framework. 
Using the envelope theorem, we develop a principled gradient-ascent method for the transport-based Fisher ratio and establish a conditional first-order convergence guarantee. We also investigate spectral methods as a way to find promising starting points for gradient refinement. These developments lead to Screened GeoGrad, our proposed algorithm integrating initialization screening, a safeguarded spectral step, and gradient ascent. Simulations and three single-cell applications
show that the resulting projections can benefit classification, although performance
depends on the representation, retained dimension, and downstream classifier.

Pairwise transport is computationally demanding, but computations for different
subject pairs can be parallelized. The formulation also permits focusing on
selected, difficult-to-separate subjects, without guaranteeing improved
classification. Compact GMMs can substantially reduce runtime, while introducing
estimation and covariance-computation costs; their advantages over empirical inputs
depend on the setting.

As future work, it would be interesting to investigate selective sampling of cells and subject pairs to reduce computational cost while preserving informative distributional differences.
Another direction is to replace pairwise scatter by criteria based on Wasserstein
barycenters: within-class variation could use distances to class barycenters,
and between-class variation could use distances among those barycenters. Fast algorithms can be used to calculate the Wasserstein barycenters~\citep{ye2017fast,yang2021fast}.
Such criteria would define different objectives, whose statistical and
computational tradeoffs require investigation. Larger-cohort studies are also
needed to develop and assess validation procedures for jointly choosing the
distribution representation, retained dimension, and downstream classifier.

\acks{Jia Li's research is supported by the National Science Foundation under grant no. CCF-2205004. Lin Lin's research is supported by NCI CCSG P30CA014236 and Duke University Center for AIDS Research (CFAR) AI064518.}

\appendix
\section{Grassmann Subspaces, Principal Angles, and Distance}
\label{ap:grassmann}
Let \(\mathcal S_A,\mathcal S_B\in\operatorname{Gr}(q,d)\), and
let \(A,B\in\mathbb R^{d\times q}\) be any orthonormal matrices whose columns span
these subspaces. If
\[
\sigma_1,\ldots,\sigma_q
\]
are the singular values of \(A^\top B\), the principal angles are
\[
\theta_\ell=\arccos(\sigma_\ell),\qquad
0\leq\theta_\ell\leq\frac{\pi}{2},\quad \ell=1,\ldots,q.
\]
The angles do not depend on the particular orthonormal bases \(A\) and \(B\). Under
the canonical Grassmann metric, the geodesic distance used in this paper is
\[
d_{\mathrm{Gr}}(\mathcal S_A,\mathcal S_B)
=\left(\sum_{\ell=1}^{q}\theta_\ell^2\right)^{1/2}.
\]
It equals zero exactly when the two matrices span the same subspace. For \(q=1\), it
is the smaller angle between the two unoriented lines, so directions \(a\) and \(-a\)
have distance zero. In computation, \(A\) and \(B\) are first orthonormalized, the
singular values of \(A^\top B\) are clipped to \([0,1]\) to remove roundoff error,
and the formula above is applied. See \citet{AbsilEtAl2008} for the geometry and
optimization of Grassmann and related matrix manifolds.

\paragraph{Constructing the geodesic for backtracking.}
\label{ap:grassmann-geodesic}
For the current and proposed orthonormal projection matrices \(A\) and \(B\),
compute the singular value decomposition
\[
A^\top B=U\operatorname{diag}(\cos\theta_1,\ldots,\cos\theta_q)V^\top.
\]
Let \(a_\ell\) and \(b_\ell\) be the \(\ell\)th columns of \(AU\) and \(BV\),
respectively. These are aligned principal vectors for the two subspaces.
For each \(\theta_\ell>0\), define
\[
w_\ell=\frac{b_\ell-a_\ell\cos\theta_\ell}{\sin\theta_\ell},
\qquad
a_\ell(\alpha)=a_\ell\cos(\alpha\theta_\ell)
+w_\ell\sin(\alpha\theta_\ell),\quad 0\leq\alpha\leq1.
\]
If \(\theta_\ell=0\), the two principal vectors coincide; set
\(a_\ell(\alpha)=a_\ell\) without performing the division.
Then
\[
A(\alpha)=[a_1(\alpha),\ldots,a_q(\alpha)]
\]
has orthonormal columns, and its span traces a shortest Grassmann geodesic
between the two subspaces.
Indeed, the positive-angle vectors \(w_\ell\) are mutually orthonormal and
orthogonal to \(A\), so the construction rotates each principal vector
through the angle \(\alpha\theta_\ell\) in its corresponding two-dimensional
plane. In particular,
\[
d_{\mathrm{Gr}}\bigl(\operatorname{span}(A),\operatorname{span}(A(\alpha))\bigr)
=\alpha\left(\sum_{\ell=1}^{q}\theta_\ell^2\right)^{1/2}.
\]
Thus \(\alpha\) specifies a fraction of the total geodesic length, rather than
an entrywise linear interpolation of the projection matrices. The endpoint
matrices are \(A(0)=AU\) and \(A(1)=BV\); their spans are the required endpoints.
If a principal angle equals \(\pi/2\), a shortest path need not be unique;
the chosen singular vectors select one such path. Numerically, the implementation
treats sufficiently small angles as zero and applies a final QR factorization
to restore orthonormality against roundoff. Algorithm~\ref{alg:OTAF} uses this
construction with \(A=A_0\), \(B=A_{\mathrm{geo}}\), and
\(\alpha\in\{1,1/2,1/4,1/8,1/16\}\).

\section{Envelope Theorem and Derivation of the Exact OTAF Gradient}
\label{ap:envelope-gradient}

\paragraph{Danskin's envelope theorem.}
The following minimum-form statement is the version used in
Proposition~\ref{prop:envelope-gradient-rev3}. Let \(\mathcal U\) be an open subset
of a finite-dimensional Euclidean space, let \(\mathcal Y\) be a nonempty compact
subset of a possibly different finite-dimensional Euclidean space, and let
\(f:\mathcal U\times\mathcal Y\to\mathbb R\) be continuous. Suppose that
\(f(\,\cdot\,,y)\) is differentiable and that its derivative is continuous jointly
in \((x,y)\). Define
\[
v(x)=\min_{y\in\mathcal Y}f(x,y),
\qquad
\mathcal Y^*(x)=\underset{y\in\mathcal Y}{\operatorname{arg\,min}}\,f(x,y).
\]
Then the one-sided directional derivative satisfies
\begin{equation}
v'(x;h)
=\min_{y\in\mathcal Y^*(x)}\mathbf{d}_xf(x,y)[h].
\label{eq:appendix-danskin}
\end{equation}
If the minimizer is unique, or if all active minimizers have the same derivative
with respect to \(x\), the right-hand side is a linear function of \(h\);
\(v\) is then differentiable at \(x\), and
\[
\mathbf{d}v(x)[h]=\mathbf{d}_xf(x,y^*)[h]
\]
for any \(y^*\in\mathcal Y^*(x)\). This is Danskin's envelope theorem
\citep{Danskin1967}. A local coordinate chart gives the same statement when \(x\)
lies on a smooth manifold.

The bracket in \(\mathbf{d}v(x)[h]\) means that the differential---a linear
functional---is evaluated in direction \(h\). In the OTAF notation,
\(\mathbf{d}D_s^{(\vartheta)}(A_\tau)[H]\) is therefore the directional derivative of the
optimized pair cost when the projection moves from \(A_\tau\) in the Grassmann
tangent direction \(H\). Equivalently, if \(A(t)\) is any smooth Grassmann curve
with \(A(0)=A_\tau\) and \(A'(0)=H\), then
\[
\mathbf{d}D_s^{(\vartheta)}(A_\tau)[H]
=\left.\frac{\mathbf{d}}{\mathbf{d}t}D_s^{(\vartheta)}\{A(t)\}\right|_{t=0}.
\]

\paragraph{Verification for an OTAF pairwise transport problem.}
For a selected subject pair \(\vartheta=(k_1,k_2)\), let \(u^{(\vartheta)}\) and \(v^{(\vartheta)}\)
denote its two marginal-weight vectors, using empirical support weights or GMM
component weights as appropriate. The feasible set is
\[
\Pi_{\vartheta}=\left\{\boldsymbol\lambda\geq0:
\sum_j\lambda_{ij}=u_i^{(\vartheta)},\
\sum_i\lambda_{ij}=v_j^{(\vartheta)}\right\}.
\]
It is nonempty because the product coupling \(u^{(\vartheta)}(v^{(\vartheta)})^\top\) is
feasible. It is closed and bounded
(all entries lie in \([0,1]\)), hence compact. Moreover, the representation and its
weights are fitted before OTAF and held fixed, so projection changes only the cost
matrix and not \(\Pi_{\vartheta}\). Compactness and independence from \(A\) are therefore
automatic features of the finite transport formulation, rather than stringent
modeling assumptions. They would cease to be automatic if component weights,
empirical weights, or the representation itself were refitted after each projection.

The pair cost
\[
\varphi_s^{(\vartheta)}(A,\boldsymbol\lambda)
=\langle\boldsymbol\lambda,C_s^{(\vartheta)}(A)\rangle
\]
is continuous because it is a finite weighted sum of projected ground costs. For an
empirical representation, those ground costs are quadratic polynomials in \(A\), so
their derivatives are continuous. For a GMM representation, the same conclusion
holds wherever the projected component covariances are positive definite. This
condition is natural for full-rank covariance estimates; nearly singular fitted
covariances require regularization. At a positive-semidefinite boundary the Gaussian
\(W^2\) value remains continuous, but the smooth derivative formula used here need
not apply.

The substantive assumption is uniqueness of the optimal transport, or the weaker
condition that every optimal transport has the same derivative with respect to
\(A\). A finite transport linear program can have multiple optimizers when costs
tie. Away from such tie surfaces, uniqueness is typical under continuously varying
costs, but it is not guaranteed for a particular data set. If several optimal
couplings have different derivatives, Eq.~\eqref{eq:appendix-danskin} still gives
directional derivatives, but \(D_s^{(\vartheta)}\), and hence \(F_s\), may be nonsmooth.
The smooth-gradient convergence theorem then does not apply without a nonsmooth
extension. Finally, the assumption
\(V_{W,s}^{\mathrm{opt}}(A_\tau)>0\) is needed only to differentiate the Fisher
ratio; it fails when the selected within-class distributions have zero average
projected separation.

\paragraph{Proof of Proposition~\ref{prop:envelope-gradient-rev3}.}
Apply Eq.~\eqref{eq:appendix-danskin} in a local coordinate chart around
\(A_\tau\), with
\(\mathcal Y=\Pi_{\vartheta}\),
\(f=\varphi_s^{(\vartheta)}\), and
\(y^*=\boldsymbol\lambda_\tau^{(\vartheta),*}\). The verified assumptions and the
common-active-derivative condition give
\[
\mathbf{d}D_s^{(\vartheta)}(A_\tau)[H]
=\mathbf{d}_A\varphi_s^{(\vartheta)}
(A_\tau,\boldsymbol\lambda_\tau^{(\vartheta),*})[H].
\]
Finite averaging therefore gives the derivatives of the two
transport-reoptimized variations by differentiating their fixed-coupling versions
at \((A_\tau,\boldsymbol\Gamma_\tau)\). Since the optimized and fixed-coupling
variations also have the same values at that point, the quotient rule yields
\[
\mathbf{d} F_s(A_\tau)[H]
=\frac{
V_{W,s}^{\mathrm{opt}}(A_\tau)
\mathbf{d}_A V_{B,s}^{\mathrm{fix}}
(A_\tau;\boldsymbol\Gamma_\tau)[H]
-V_{B,s}^{\mathrm{opt}}(A_\tau)
\mathbf{d}_A V_{W,s}^{\mathrm{fix}}
(A_\tau;\boldsymbol\Gamma_\tau)[H]}
{\{V_{W,s}^{\mathrm{opt}}(A_\tau)\}^2}.
\]
This is exactly the directional derivative of
\(R_{s,\tau}\) at \(A_\tau\). Projecting its Euclidean gradient onto the
Grassmann tangent space proves the second identity in
Eq.~\eqref{eq:envelope-gradient-rev3}.

\paragraph{Ground-cost and aggregate-gradient calculation.}
For an empirical ground cost, with
\(z=x_i^{(k_1)}-x_j^{(k_2)}\),
\[
\mathbf{d}\|A^\top z\|^2[H]
=2\operatorname{tr}(H^\top zz^\top A),
\qquad
\nabla_A\|A^\top z\|^2=2zz^\top A.
\]
Fix a subject pair \(\vartheta=(k_1,k_2)\) and component indices \(i,j\).
For this derivation only, abbreviate the two components by
\[
\begin{aligned}
(\phi_1,\mu_1,\Sigma_1)
&=(\phi_i^{(k_1)},\mu_i^{(k_1)},\Sigma_i^{(k_1)}),\\
(\phi_2,\mu_2,\Sigma_2)
&=(\phi_j^{(k_2)},\mu_j^{(k_2)},\Sigma_j^{(k_2)}).
\end{aligned}
\]
Thus \(\Delta\mu=\mu_1-\mu_2=\Delta\mu_{ij}^{(\vartheta)}\).
For \(\ell\in\{1,2\}\), let \(S_\ell=A^\top\Sigma_\ell A\) be the
projected component covariance. The squared Wasserstein distance between
the two projected Gaussian components is
\[
W^2\!\left(P(\phi_1,A),P(\phi_2,A)\right)
=\|A^\top\Delta\mu\|^2+\operatorname{tr}\!\left[
S_1+S_2-2(S_1^{1/2}S_2S_1^{1/2})^{1/2}\right].
\]
Denote its covariance contribution by
\[
b(S_1,S_2)
=\operatorname{tr}\!\left[
S_1+S_2-2(S_1^{1/2}S_2S_1^{1/2})^{1/2}\right].
\]
When \(S_1\) and \(S_2\) are positive definite, define
\[
T_{12}=S_1^{-1/2}(S_1^{1/2}S_2S_1^{1/2})^{1/2}S_1^{-1/2},
\qquad T_{21}=T_{12}^{-1}.
\]
The dependence on \(A\) is suppressed in \(S_1,S_2,T_{12},T_{21}\) only
within this derivation. At \(A=A_\tau\), these abbreviations correspond to
the indexed quantities in the main text as follows:
\[
\begin{aligned}
S_1(A_\tau)&=S_{k_1,i,\tau},&
S_2(A_\tau)&=S_{k_2,j,\tau},\\
T_{12}(A_\tau)&=T_{i,j,\tau}^{(\vartheta)},&
T_{21}(A_\tau)&=T_{j,i,\tau}^{(\vartheta^r)}.
\end{aligned}
\]
To obtain the differential of \(b\), put
\(M=S_1^{1/2}S_2S_1^{1/2}\) and use
\[
\mathbf{d}\operatorname{tr}(X^{1/2})[E]
=\tfrac12\operatorname{tr}(X^{-1/2}E)
\]
for positive-definite \(X\). Differentiating with respect to \(S_2\) gives
\[
\begin{aligned}
\mathbf{d}_{S_2}b(S_1,S_2)[E_2]
=\operatorname{tr}(E_2)
-\operatorname{tr}
\!\left(S_1^{1/2}M^{-1/2}S_1^{1/2}E_2\right)=\operatorname{tr}\{(I_q-T_{21})E_2\},
\end{aligned}
\]
because
\(S_1^{1/2}M^{-1/2}S_1^{1/2}=T_{12}^{-1}=T_{21}\).
The symmetry of \(b(S_1,S_2)\) gives the analogous derivative with respect to
\(S_1\). Therefore,
\[
\mathbf{d}b(S_1,S_2)[E_1,E_2]
=\operatorname{tr}\{(I_q-T_{12})E_1\}
+\operatorname{tr}\{(I_q-T_{21})E_2\}.
\]
Using
\[
\mathbf{d} S_\ell(A)[H]
=H^\top\Sigma_\ell A+A^\top\Sigma_\ell H
\]
and adding the derivative of the mean term, we get
\[
\nabla_A W^2\!\left(P(\phi_1,A),P(\phi_2,A)\right)
=2\Delta\mu\Delta\mu^\top A
+2\Sigma_1A(I_q-T_{12})
+2\Sigma_2A(I_q-T_{21}).
\]
Evaluating at \(A=A_\tau\) and restoring the indexed notation gives
Eq.~\eqref{eq:exact-component-gradient-rev3}. Weighting these entrywise
gradients by \(\lambda_{\tau,ij}^{(\vartheta),*}\), summing over component or support-point
pairs, averaging over selected subject pairs, and applying the quotient rule
produces \(Z_\tau\) in Eq.~\eqref{eq:exact-grassmann-gradient}. Under the canonical
Grassmann metric, this Euclidean gradient must be projected onto the tangent
space. Let
\(\langle U,V\rangle_F=\operatorname{tr}(U^\top V)\) denote the Frobenius
inner product. For every Grassmann tangent direction \(H\), \(A_\tau^\top H=0\);
therefore
\[
\begin{aligned}
\langle Z_\tau,H\rangle_F
&=\langle(I_d-A_\tau A_\tau^\top)Z_\tau,H\rangle_F,\\
\langle A_\tau A_\tau^\top Z_\tau,H\rangle_F
&=\langle A_\tau^\top Z_\tau,A_\tau^\top H\rangle_F=0.
\end{aligned}
\]
Consequently, \((I_d-A_\tau A_\tau^\top)Z_\tau\) is the tangent vector that
represents the same directional derivatives along all feasible Grassmann
directions. Hence
\[
\operatorname{grad}F_s(A_\tau)
=(I_d-A_\tau A_\tau^\top)Z_\tau,
\]
completing the stated gradient calculation.

\paragraph{Numerical stabilization near singular projected covariances.}
The gradient formula above assumes that
\(S_{k_1,i,\tau}=A_\tau^\top\Sigma_i^{(k_1)}A_\tau\) and
\(S_{k_2,j,\tau}=A_\tau^\top\Sigma_j^{(k_2)}A_\tau\) are positive definite. In computing
the covariance transport matrices \(T_{i,j,\tau}^{(\vartheta)}\) and
\(T_{j,i,\tau}^{(\vartheta^r)}\), the
implementation always replaces each projected covariance by
\[
\begin{aligned}
S_{k_1,i,\tau}&\longmapsto S_{k_1,i,\tau}+\varepsilon I_q,\\
S_{k_2,j,\tau}&\longmapsto S_{k_2,j,\tau}+\varepsilon I_q,
\end{aligned}
\qquad
\varepsilon=10^{-10}
\max\{\|S_{k_1,i,\tau}\|_2,\|S_{k_2,j,\tau}\|_2\}.
\]
If both projected covariances are zero, the scale in this formula is set to one.
This relative ridge is applied even when the projected covariances are positive
definite; its numerical effect is expected to be small when they are well-conditioned.
It is used only to construct the trial direction. Each trial is evaluated using
the original transport-reoptimized objective \(F_{\mathsf{gmm}}\), without the ridge,
and the tolerance-relaxed Armijo test described in
Section~\ref{sec:gradient_otaf}. Hence the implementation permits only
tolerance-scale decreases and returns its best evaluated projection. A fixed
nonzero ridge does not produce the exact derivative used in
Theorem~\ref{thm:gradient-convergence-rev3}, even at positive-definite covariances.

\paragraph{Proof of Theorem~\ref{thm:gradient-convergence-rev3}.}
Write the QR retraction as
\(\mathcal R_A(H)=\operatorname{qf}(A+H)\) for a Grassmann tangent vector
\(H\), so \(A^\top H=0\). To verify its first-order property, let
\(A+tH=Q(t)R(t)\) be the thin QR factorization with the usual positive-diagonal
convention. Then
\[
R(t)^\top R(t)
=(A+tH)^\top(A+tH)
=I_q+t^2H^\top H.
\]
It follows that \(R(t)=I_q+O(t^2)\), and hence
\[
\mathcal R_A(tH)=Q(t)=(A+tH)R(t)^{-1}=A+tH+O(t^2).
\]
Thus \(\mathcal R_A(0)=A\) and
\(\mathbf{d}\mathcal R_A(0)[H]=H\); the QR map is a first-order retraction on
the Grassmann manifold.

By Proposition~\ref{prop:envelope-gradient-rev3}, the direction computed at
iteration \(\tau\) is the true Grassmann gradient,
\[
G_\tau=\operatorname{grad}F_s(A_\tau).
\]
The Lipschitz-gradient assumption, the smooth first-order retraction, and compactness
of \(\mathcal L_0\) imply a uniform retraction ascent bound: there exist constants
\(L_{\mathcal R}>0\) and \(\rho>0\) such that, for every
\(A\in\mathcal L_0\) and tangent vector \(H\) with \(\|H\|_F\leq\rho\),
\begin{equation}
F_s\{\mathcal R_A(H)\}
\geq F_s(A)+\langle\operatorname{grad}F_s(A),H\rangle_F
-\frac{L_{\mathcal R}}{2}\|H\|_F^2.
\label{eq:appendix-retraction-ascent}
\end{equation}
This is the standard retraction form of the Riemannian descent lemma, with signs
reversed for maximization; see \citet{AbsilEtAl2008}.

Set \(H=\alpha G_\tau\) in
Eq.~\eqref{eq:appendix-retraction-ascent}. Whenever
\(\alpha\|G_\tau\|_F\leq\rho\),
\begin{equation}
F_s\{\mathcal R_{A_\tau}(\alpha G_\tau)\}
\geq F_s(A_\tau)
+\alpha\left(1-\frac{L_{\mathcal R}\alpha}{2}\right)
\|G_\tau\|_F^2.
\label{eq:appendix-gradient-ascent}
\end{equation}
Consequently, every sufficiently small positive \(\alpha\) satisfies the Armijo
condition: it is enough that
\[
\alpha\|G_\tau\|_F\leq\rho,
\qquad
\alpha\leq\frac{2(1-\omega)}{L_{\mathcal R}}.
\]
The halving search therefore terminates after finitely many trials whenever
\(G_\tau\neq0\). Its accepted step obeys
\[
F_s(A_{\tau+1})-F_s(A_\tau)
\geq\omega\alpha_\tau\|G_\tau\|_F^2>0,
\]
which also shows inductively that every accepted iterate remains in
\(\mathcal L_0\).

It remains to show that the gradient norms converge to zero. Continuity of the
gradient and compactness of \(\mathcal L_0\) give
\[
M:=\max_{A\in\mathcal L_0}\|\operatorname{grad}F_s(A)\|_F<\infty.
\]
If \(M=0\), the result is immediate. Otherwise the initial trial step satisfies
\[
\alpha_{0,\tau}=\min\{1,\|G_\tau\|_F^{-1}\}
\geq a_0:=\min\{1,M^{-1}\}>0.
\]
Define
\[
\bar\alpha
=\min\left\{\frac{\rho}{M},
\frac{2(1-\omega)}{L_{\mathcal R}}\right\}>0.
\]
Because backtracking halves the step size, the first accepted step is bounded
uniformly below by
\[
\alpha_\tau\geq
\underline\alpha:=\frac12\min\{a_0,\bar\alpha\}>0.
\]
Since \(F_s\) is continuous on the compact set \(\mathcal L_0\), it is bounded
above there. Summing the Armijo increases yields
\[
\omega\underline\alpha
\sum_{\tau=0}^{\infty}\|G_\tau\|_F^2
\leq
\sum_{\tau=0}^{\infty}
\{F_s(A_{\tau+1})-F_s(A_\tau)\}<\infty.
\]
Therefore \(\|G_\tau\|_F\to0\). The objective sequence is nondecreasing and
bounded above, so it converges. Finally, every accumulation point
\(\bar A\) lies in \(\mathcal L_0\), and continuity of the Riemannian gradient
gives \(\operatorname{grad}F_s(\bar A)=0\). Thus every accumulation point is
first-order stationary for the transport-reoptimized objective.
\hfill\BlackBox\par\medskip

\section{Fixed-Coupling and Gaussian-Coupling Identities} \label{ap1}

\subsection{Empirical-Distribution Scatter Identities}
For this derivation, abbreviate
\(C_B=C_{B,\mathsf{emp}}(\boldsymbol\Gamma)\) and
\(C_W=C_{W,\mathsf{emp}}(\boldsymbol\Gamma)\) from
Eq.~\eqref{eq:empirical-spectral-scatter}. Put
\(z_{ij}^{(k_1,k_2)}=x_i^{(k_1)}-x_j^{(k_2)}\) and, for a fixed
coupling collection \(\boldsymbol\Gamma\), define
\begin{align}
C_B
&=\frac{1}{|\mathcal I_B|}
\sum_{(k_1,k_2)\in\mathcal I_B}\sum_{ij}
\lambda_{ij}^{(k_1,k_2)}z_{ij}^{(k_1,k_2)}
\bigl (z_{ij}^{(k_1,k_2)}\bigr)^\top,
\label{eq:CB}\\
C_W
&=\frac{1}{|\mathcal I_W|}
\sum_{(k_1,k_2)\in\mathcal I_W}\sum_{ij}
\lambda_{ij}^{(k_1,k_2)}z_{ij}^{(k_1,k_2)}
\bigl(z_{ij}^{(k_1,k_2)}\bigr)^\top.
\label{eq:CW}
\end{align}
Then
\begin{align}
V^{\mathrm{fix}}_{B,\mathsf{emp}}(A;\boldsymbol\Gamma)
&=\operatorname{tr}(A^\top C_BA),
\label{eq:V_B1}\\
V^{\mathrm{fix}}_{W,\mathsf{emp}}(A;\boldsymbol\Gamma)
&=\operatorname{tr}(A^\top C_WA).
\label{eq:V_W1}
\end{align}
Indeed, cyclicity of the trace gives
\[
\|A^\top z_{ij}^{(k_1,k_2)}\|^2
=\operatorname{tr}\!\left\{A^\top z_{ij}^{(k_1,k_2)}
\{z_{ij}^{(k_1,k_2)}\}^\top A\right\}.
\]
Substitution and linearity of the trace prove both identities.

\subsection{Quadratic Gaussian-Coupling Costs and Mixture Aggregation}
\label{ap2}

We first derive a common quadratic identity and then specialize it to the independent
and geometric couplings. Let \((Z_1,Z_2)\) be any joint Gaussian vector with means
\(\mu_1,\mu_2\), marginal covariance matrices \(\Sigma_1,\Sigma_2\), and
cross-covariance
\(U_{12}=\operatorname{Cov}(Z_1,Z_2)\).
Its block covariance is feasible precisely when
\(\left(\begin{smallmatrix}\Sigma_1&U_{12}\\
U_{12}^\top&\Sigma_2\end{smallmatrix}\right)\succeq0\). Writing
\(\Delta\mu=\mu_1-\mu_2\), we have
\[
E\{(Z_1-Z_2)(Z_1-Z_2)^\top\}
=\Delta\mu\Delta\mu^\top+
\Sigma_1+\Sigma_2-U_{12}-U_{12}^\top.
\]
Let $H(U_{12})
=\Delta\mu\Delta\mu^\top+
\Sigma_1+\Sigma_2-U_{12}-U_{12}^\top$. Then
\begin{align}
E\|A^\top(Z_1-Z_2)\|^2
&=\operatorname{tr}\!\left[A^\top H(U_{12})A\right]\, .
\label{eq:appendix-general-coupling-quadratic}
\end{align}
This proves Eq.~\eqref{eq:fixed-gaussian-coupling-cost}. Because the projected
Wasserstein cost minimizes the same expected squared difference over all projected
couplings, every feasible fixed joint coupling gives an upper bound on the exact
projected Gaussian \(W^2\) cost.

For the independent coupling, \(U_{12}=0\). At \(A=I_d\),
Eq.~\eqref{eq:appendix-general-coupling-quadratic} becomes
\[
E\|Z_1-Z_2\|^2
=\|\Delta\mu\|^2+\operatorname{tr}(\Sigma_1+\Sigma_2),
\]
which is Eq.~\eqref{eq:W2}. For a general projection \(A\), the same identity gives
\[
c_{12}^{\mathrm{ind}}(A)
=\operatorname{tr}\!\left[A^\top
\{\Delta\mu\Delta\mu^\top+\Sigma_1+\Sigma_2\}A\right],
\]
which is Eq.~\eqref{eq:W2P}.

For the geometric coupling, \(U_{12}=G_{12}=G_{12}^\top\), where
\(G_{12}\) is defined in Eq.~\eqref{eq:geometric-cross-covariance}. Hence
Eq.~\eqref{eq:appendix-general-coupling-quadratic} gives
\[
c_{12}^{\mathrm{geo}}(A)
=\operatorname{tr}\!\left[A^\top
\{\Delta\mu\Delta\mu^\top+
\Sigma_1+\Sigma_2-2G_{12}\}A\right],
\]
which is Eq.~\eqref{eq:geometric-bound-cost}. The validity of this particular joint
coupling and its common-shape exactness are proved in the next subsection.

We now aggregate the componentwise identities. For a subject pair
\(\vartheta=(k_1,k_2)\), let \(\boldsymbol\lambda^{(\vartheta)}\) be a fixed admissible
mixture-component coupling and let \(\eta\in\{\mathrm{ind},\mathrm{geo}\}\). Define
\(U_{ij}^{(\vartheta,\mathrm{ind})}=0\),
\(U_{ij}^{(\vartheta,\mathrm{geo})}=G_{ij}^{(\vartheta)}\), and
\(H_{ij}^{(\vartheta,\eta)}\) as in Eq.~\eqref{eq:spectral-component-matrix}. Linearity of
the trace gives the bound-based fixed-coupling subject-pair cost
\begin{align}
\widetilde D_\eta^{(\vartheta),\mathrm{fix}}
(A;\boldsymbol\lambda^{(\vartheta)})
&=\sum_{ij}\lambda_{ij}^{(\vartheta)}
\operatorname{tr}\!\left\{A^\top H_{ij}^{(\vartheta,\eta)}A\right\}
=\operatorname{tr}\!\left[A^\top
\left\{\sum_{ij}\lambda_{ij}^{(\vartheta)}H_{ij}^{(\vartheta,\eta)}\right\}A\right].
\label{eq:appendix-subject-pair-aggregation}
\end{align}
For \(L\in\{B,W\}\), averaging Eq.~\eqref{eq:appendix-subject-pair-aggregation}
over the selected pair set \(\mathcal I_L\) yields
\begin{align}
\widetilde V_{L,\mathsf{gmm}}^{\eta,\mathrm{fix}}
(A;\boldsymbol\Gamma)
&=\frac{1}{|\mathcal I_L|}\sum_{\vartheta\in\mathcal I_L}
\widetilde D_\eta^{(\vartheta),\mathrm{fix}}
(A;\boldsymbol\lambda^{(\vartheta)})
=\operatorname{tr}\!\left\{A^\top
C_{L,\eta}(\boldsymbol\Gamma)A\right\},
\label{eq:appendix-class-aggregation}
\end{align}
where \(C_{L,\eta}(\boldsymbol\Gamma)\) is exactly the matrix in
Eq.~\eqref{eq:spectral-scatter-matrices}. Thus both choices of coupling produce the
same quadratic aggregation structure. For \(q=1\), taking the ratio of the
between-class and within-class versions of
Eq.~\eqref{eq:appendix-class-aggregation} gives the generalized Rayleigh quotient in
Eq.~\eqref{eq:spectral-rayleigh-quotient}.

\subsection{Geometric Gaussian Coupling and Common-Shape Exactness}

\paragraph{Positive definiteness and validity of the coupling.}
Assume \(\Sigma_1,\Sigma_2\succ0\) and put
\[
M=\Sigma_1^{-1/2}\Sigma_2\Sigma_1^{-1/2}.
\]
The matrix \(M\) is positive definite because it is a congruence of
\(\Sigma_2\) by the nonsingular matrix \(\Sigma_1^{-1/2}\). Hence
\(M^{1/2}\succ0\), and Eq.~\eqref{eq:geometric-cross-covariance} can be written as
\[
G_{12}=\Sigma_1^{1/2}M^{1/2}\Sigma_1^{1/2}\succ0,
\]
where the final conclusion follows again by congruence. Moreover,
\begin{align*}
G_{12}\Sigma_1^{-1}G_{12}
=\Sigma_1^{1/2}M^{1/2}
\underbrace{\Sigma_1^{1/2}\Sigma_1^{-1}\Sigma_1^{1/2}}_{I_d}
M^{1/2}\Sigma_1^{1/2}
=\Sigma_1^{1/2}M\Sigma_1^{1/2}
=\Sigma_2.
\end{align*}

This identity provides an explicit construction of the claimed joint Gaussian
coupling. Let \(U\sim N(0,I_d)\), and define
\[
Z_1=\mu_1+\Sigma_1^{1/2}U,
\qquad
Z_2=\mu_2+G_{12}\Sigma_1^{-1/2}U.
\]
Then
\(\operatorname{Cov}(Z_1)=\Sigma_1\), 
\(\operatorname{Cov}(Z_2)=G_{12}\Sigma_1^{-1}G_{12}=\Sigma_2\),
\(\operatorname{Cov}(Z_1,Z_2)=G_{12}\).
Thus \((Z_1,Z_2)\) is a joint Gaussian vector with the required marginals and block
covariance
\begin{equation}
\Sigma_{\mathrm{joint}}=
\begin{pmatrix}
\Sigma_1&G_{12}\\G_{12}&\Sigma_2
\end{pmatrix}
=
\begin{pmatrix}
\Sigma_1^{1/2}\\G_{12}\Sigma_1^{-1/2}
\end{pmatrix}
\begin{pmatrix}
\Sigma_1^{1/2}\\G_{12}\Sigma_1^{-1/2}
\end{pmatrix}^{\!\top}\succeq0.
\label{eq:geometric-block-factorization}
\end{equation}
The factor on the right has rank \(d\), because its upper \(d\times d\) block is
nonsingular. Therefore \(\operatorname{rank}(\Sigma_{\mathrm{joint}})=d<2d\), so
\(\Sigma_{\mathrm{joint}}\) is singular. Equivalently, its Schur complement with respect to
\(\Sigma_1\) is
\(\Sigma_2-G_{12}\Sigma_1^{-1}G_{12}=0\).

\paragraph{Exactness for proportional common-shape covariances.}
Let \(\Sigma_i=s_i\Sigma_0\) and \(\Sigma_j=s_j\Sigma_0\), where
\(s_i,s_j>0\) and \(\Sigma_0\succ0\). Direct substitution into
Eq.~\eqref{eq:geometric-cross-covariance} gives
\begin{align*}
G_{ij}
&=\sqrt{s_i}\,\Sigma_0^{1/2}
\left(\frac{s_j}{s_i}I_d\right)^{1/2}
\sqrt{s_i}\,\Sigma_0^{1/2}
=\sqrt{s_is_j}\,\Sigma_0.
\end{align*}
For any \(A\in\operatorname{St}(q,d)\), put
\(S_0=A^\top\Sigma_0A\). The two projected covariance matrices are
\(S_i=s_iS_0\) and \(S_j=s_jS_0\), and therefore
\[
\left(S_i^{1/2}S_jS_i^{1/2}\right)^{1/2}
=\sqrt{s_is_j}\,S_0
=A^\top G_{ij}A.
\]
Substitution into the exact Gaussian formula in Eq.~\eqref{eq:w2phi} yields
\begin{align*}
W^2\!\left(P(\phi_i,A),P(\phi_j,A)\right)
=\|A^\top(\mu_i-\mu_j)\|^2+\operatorname{tr}\!\left[
A^\top\{\Sigma_i+\Sigma_j-2G_{ij}\}A\right]
=c_{ij}^{\mathrm{geo}}(A).
\end{align*}
Hence the geometric bound is exact for every \(A\) in this regime.

\section{Independent-Coupling Fixed-Ratio Analysis} \label{ap4}
This appendix records a detailed analysis of the independent-coupling approximation.
It addresses the fixed-coupling projection subproblem only. In particular, it does not
imply that the transport-reoptimized objective increases after the mixture-component
couplings are recomputed.

\paragraph{Gaussian component discrepancy.}
Let \(S_1=A^\top\Sigma_1A\) and \(S_2=A^\top\Sigma_2A\). Subtracting the exact
Gaussian formula in Eq.~\eqref{eq:w2phi} from the independent-coupling formula in
Eq.~\eqref{eq:W2P} gives
\begin{equation}
g_{12}(A):=c_{12}^{\mathrm{ind}}(A)
-W^2\!\left(P(\phi_1,A),P(\phi_2,A)\right)
=2\operatorname{tr}\!\left[
\left(S_1^{1/2}S_2S_1^{1/2}\right)^{1/2}\right]\geq0.
\label{eq:component_surrogate_gap}
\end{equation}
Moreover,
\begin{equation}
0\leq g_{12}(A)
\leq2\sqrt{\operatorname{tr}(S_1)\operatorname{tr}(S_2)}.
\label{eq:component_surrogate_bound}
\end{equation}
To prove the upper bound, use
\[
\operatorname{tr}\!\left[
\left(S_1^{1/2}S_2S_1^{1/2}\right)^{1/2}\right]
=\left\|S_1^{1/2}S_2^{1/2}\right\|_*
\leq\left\|S_1^{1/2}\right\|_F
\left\|S_2^{1/2}\right\|_F
\]
and \(\|S_\ell^{1/2}\|_F^2=\operatorname{tr}(S_\ell)\).

\paragraph{Effect on the fixed-coupling Fisher ratio.}
Aggregate \(g_{ij}(A)\) with the same fixed component couplings and pair averages used
in \(V_{\iota,\mathsf{gmm}}^{\mathrm{fix}}\), and denote the resulting nonnegative quantity
by \(\mathcal E_\iota(A;\boldsymbol\Gamma)\), \(\iota\in\{B,W\}\). Then
\[
\widetilde V_{\iota,\mathsf{gmm}}^{\mathrm{ind},\mathrm{fix}}
=V_{\iota,\mathsf{gmm}}^{\mathrm{fix}}+\mathcal E_\iota.
\]
When the exact variations are positive, put
\(e_\iota=\mathcal E_\iota/V_{\iota,\mathsf{gmm}}^{\mathrm{fix}}\). Direct algebra gives
\begin{equation}
\frac{\widetilde r_{\mathsf{gmm}}^{\mathrm{ind},\mathrm{fix}}}
{r_{\mathsf{gmm}}^{\mathrm{fix}}}
=\frac{1+e_B}{1+e_W},
\qquad
\widetilde r_{\mathsf{gmm}}^{\mathrm{ind},\mathrm{fix}}-r_{\mathsf{gmm}}^{\mathrm{fix}}
=r_{\mathsf{gmm}}^{\mathrm{fix}}\frac{e_B-e_W}{1+e_W}.
\label{eq:surrogate_ratio_identity}
\end{equation}
Consequently,
\begin{equation}
\left|\widetilde r_{\mathsf{gmm}}^{\mathrm{ind},\mathrm{fix}}
-r_{\mathsf{gmm}}^{\mathrm{fix}}\right|
\leq r_{\mathsf{gmm}}^{\mathrm{fix}}(e_B+e_W).
\label{eq:surrogate_ratio_bound}
\end{equation}
Thus componentwise upper bounds do not produce a ratio ordering: the sign is determined
by the difference between the relative numerator and denominator errors.

\paragraph{Fixed-coupling maximizer perturbation.}
On a compact feasible set \(\mathcal A\), suppose
\[
\sup_{A\in\mathcal A}
\left|\widetilde r_{\mathsf{gmm}}^{\mathrm{ind},\mathrm{fix}}(A;\boldsymbol\Gamma)
-r_{\mathsf{gmm}}^{\mathrm{fix}}(A;\boldsymbol\Gamma)\right|
\leq\varepsilon.
\]
Let \(A^\star\) maximize the exact fixed-coupling ratio and let \(\widehat A\)
maximize its independent-coupling approximation. Optimality of \(\widehat A\) and two
applications of the uniform bound yield
\begin{equation}
0\leq r_{\mathsf{gmm}}^{\mathrm{fix}}(A^\star;\boldsymbol\Gamma)
-r_{\mathsf{gmm}}^{\mathrm{fix}}(\widehat A;\boldsymbol\Gamma)
\leq2\varepsilon.
\label{eq:surrogate_argmax_value}
\end{equation}
If the exact fixed-coupling ratio also satisfies the local quadratic-growth condition
\[
r_{\mathsf{gmm}}^{\mathrm{fix}}(A^\star;\boldsymbol\Gamma)
-r_{\mathsf{gmm}}^{\mathrm{fix}}(A;\boldsymbol\Gamma)
\geq\frac{\mu}{2}\operatorname{dist}^2(A,A^\star),
\]
then
\begin{equation}
\operatorname{dist}(\widehat A,A^\star)
\leq2\sqrt{\varepsilon/\mu}.
\label{eq:surrogate_argmax_direction}
\end{equation}
The value conclusion needs no curvature assumption; the subspace conclusion does.

\section{Finite-Representation Perturbation Proofs} \label{ap5}
We prove Propositions~\ref{prop:empirical_estimation_chain}
and~\ref{prop:gmm_estimation_chain} from Section~
\ref{sec:representation_error_propagation}, then state and derive the sufficient
sample-size requirements summarized there.

\subsection{Objective and Subspace Perturbation Bounds}

\paragraph{Empirical pair costs.}
For orthonormal \(A\), pushforward by \(x\mapsto A^\top x\) is
\(1\)-Lipschitz in \(W_2\). By the reverse triangle inequality, the pairwise
distance error is at most
\(W_2(\widehat{\mathcal P}_{k_1},\mathcal P_{k_1})
+W_2(\widehat{\mathcal P}_{k_2},\mathcal P_{k_2})\). Thus
\[
\left|W_2(A_\#\widehat{\mathcal P}_{k_1},A_\#\widehat{\mathcal P}_{k_2})
-W_2(A_\#\mathcal P_{k_1},A_\#\mathcal P_{k_2})\right|
\leq2\varepsilon_{\mathsf{emp}}.
\]
The population distance is at most \(D_{\mathsf{emp}}^{\mathsf{up}}\), and the empirical distance is
therefore at most \(D_{\mathsf{emp}}^{\mathsf{up}}+2\varepsilon_{\mathsf{emp}}\). Applying
\(|x^2-y^2|=|x-y|(x+y)\) gives, uniformly in \(A\),
\begin{equation}
\left|W^2(A_\#\widehat{\mathcal P}_{k_1},A_\#\widehat{\mathcal P}_{k_2})
-W^2(A_\#\mathcal P_{k_1},A_\#\mathcal P_{k_2})\right|
\leq\zeta_{\mathsf{emp}}.
\label{eq:empirical_pair_cost_error}
\end{equation}
Here \(\zeta_{\mathsf{emp}}
=4\varepsilon_{\mathsf{emp}}(D_{\mathsf{emp}}^{\mathsf{up}}+\varepsilon_{\mathsf{emp}})\).

\paragraph{GMM component costs and MAW values.}
Fix a subject pair \(\vartheta=(k_1,k_2)\) and a projection \(A\).
Write \(C^0\) and \(\widehat C\) for the oracle and fitted projected
component-cost matrices, with components matched as in
Proposition~\ref{prop:gmm_estimation_chain}. The preceding argument gives
\[
0\leq C^0_{ij}\leq D_{\mathsf{gmm}}^2,
\qquad
\max_{ij}|\widehat C_{ij}-C^0_{ij}|
\leq4\varepsilon_{\mathsf{gmm}}(D_{\mathsf{gmm}}+\varepsilon_{\mathsf{gmm}}).
\]
Let \(\Pi^0\) and \(\widehat\Pi\) be the coupling sets with marginals
\((p^{(k_1)},p^{(k_2)})\) and
\((\widehat p^{(k_1)},\widehat p^{(k_2)})\), respectively. Here \(p^{(k_1)}=(p_i^{(k_1)})_i\) denotes the marginal vector of the component weights for subject \(k_1\); the other marginal vectors are defined analogously.
For any \(\boldsymbol\lambda\in\Pi^0\), define
\[
\lambda^-_{ij}=\lambda_{ij}
\min\left\{1,\frac{\widehat p_i^{(k_1)}}{p_i^{(k_1)}}\right\}
\min\left\{1,\frac{\widehat p_j^{(k_2)}}{p_j^{(k_2)}}\right\}.
\]
If a denominator is zero, take its factor to be one; the corresponding row
or column of \(\boldsymbol\lambda\) is zero. The marginal deficits
\[
u=\widehat p^{(k_1)}-\boldsymbol\lambda^-\mathbf1,
\qquad
v=\widehat p^{(k_2)}-(\boldsymbol\lambda^-)^\top\mathbf1
\]
are nonnegative and have the same total mass
\(t=1-\sum_{ij}\lambda^-_{ij}\). Moreover,
\[
t\leq\frac12\|\widehat p^{(k_1)}-p^{(k_1)}\|_1
+\frac12\|\widehat p^{(k_2)}-p^{(k_2)}\|_1
\leq\varepsilon_w.
\]
For \(t>0\), set
\(\widehat{\boldsymbol\lambda}=\boldsymbol\lambda^-+uv^\top/t\);
for \(t=0\), set
\(\widehat{\boldsymbol\lambda}=\boldsymbol\lambda^-\).
Then \(\widehat{\boldsymbol\lambda}\in\widehat\Pi\).
Both \(\boldsymbol\lambda-\boldsymbol\lambda^-\) and
\(\widehat{\boldsymbol\lambda}-\boldsymbol\lambda^-\) are nonnegative
matrices of total mass \(t\), so
\[
\bigl|\langle\widehat{\boldsymbol\lambda}-\boldsymbol\lambda,C^0\rangle\bigr|
\leq D_{\mathsf{gmm}}^2t\leq D_{\mathsf{gmm}}^2\varepsilon_w.
\]
Applying this construction in both directions, with the oracle and fitted
weights interchanged, and taking minima gives
\[
\left|\min_{\boldsymbol\lambda\in\widehat\Pi}
\langle\boldsymbol\lambda,C^0\rangle
-\min_{\boldsymbol\lambda\in\Pi^0}
\langle\boldsymbol\lambda,C^0\rangle\right|
\leq D_{\mathsf{gmm}}^2\varepsilon_w.
\]
Changing the cost matrix while keeping the fitted marginals fixed gives
\[
\left|\min_{\boldsymbol\lambda\in\widehat\Pi}
\langle\boldsymbol\lambda,\widehat C\rangle
-\min_{\boldsymbol\lambda\in\widehat\Pi}
\langle\boldsymbol\lambda,C^0\rangle\right|
\leq\max_{ij}|\widehat C_{ij}-C^0_{ij}|.
\]
Combining the last two inequalities proves the MAW-value bound:
\begin{align*}
\left|\min_{\boldsymbol\lambda\in\widehat\Pi}
\langle\boldsymbol\lambda,\widehat C\rangle
-\min_{\boldsymbol\lambda\in\Pi^0}
\langle\boldsymbol\lambda,C^0\rangle\right|
\leq4\varepsilon_{\mathsf{gmm}}(D_{\mathsf{gmm}}+\varepsilon_{\mathsf{gmm}})
+D_{\mathsf{gmm}}^2\varepsilon_w
=\zeta_{\mathsf{gmm}},
\end{align*}
as defined in Eq.~\eqref{eq:gmm_maw_error_constant}.

\paragraph{Variations and ratios.}
For either representation \(s\in\{\mathsf{emp},\mathsf{gmm}\}\), let \(\zeta_s\) be its
uniform selected-pair squared-cost bound. Because each variation is a normalized
nonnegative weighted average over a fixed pair set,
\[
|\widehat V_{\iota,s}(A)-V_{\iota,s}^{0}(A)|\leq\zeta_s,
\qquad \iota\in\{\mathrm{B},\mathrm{W}\}.
\]
Assume \(V_{W,s}^{0}(A)\geq\nu_s>\zeta_s\) and
\(V_{B,s}^{0}(A)\leq\zeta^{\mathrm{up}}_{B,s}\), uniformly in \(A\).
Then \(\widehat V_{W,s}(A)\geq\nu_s-\zeta_s\), and direct ratio algebra gives
\[
|\widehat r_s(A)-r_s^0(A)|
\leq
\frac{\zeta_s}{\nu_s-\zeta_s}
+\frac{\zeta^{\mathrm{up}}_{B,s}\zeta_s}{\nu_s(\nu_s-\zeta_s)}.
\]
The uniform ratio-error constant and bound are therefore
\[
\varepsilon_{r,s}=
\frac{\zeta_s(\nu_s+\zeta^{\mathrm{up}}_{B,s})}
{\nu_s(\nu_s-\zeta_s)},
\qquad
\sup_A|\widehat r_s(A)-r_s^0(A)|\leq\varepsilon_{r,s}.
\]
This proves Eqs.~\eqref{eq:empirical_ratio_estimation_bound}
and~\eqref{eq:gmm_ratio_estimation_bound}.

\paragraph{Global maximizers and subspaces.}
\label{ap:statistical-subspaces}
Let \(A_s^0\) maximize \(r_s^0\), and let \(\widehat A_s\) maximize
\(\widehat r_s\). Adding and subtracting the estimated ratios at these
two projections gives
\begin{align*}
r_s^0(A_s^0)-r_s^0(\widehat A_s)
={}&\bigl[r_s^0(A_s^0)-\widehat r_s(A_s^0)\bigr]
+\bigl[\widehat r_s(A_s^0)-\widehat r_s(\widehat A_s)\bigr]
+\bigl[\widehat r_s(\widehat A_s)-r_s^0(\widehat A_s)\bigr]\\
\leq{}&\varepsilon_{r,s}+0+\varepsilon_{r,s}
=2\varepsilon_{r,s}.
\end{align*}
The first and third terms are each at most \(\varepsilon_{r,s}\) by the
uniform ratio bound; the middle term is nonpositive because
\(\widehat A_s\) maximizes \(\widehat r_s\).
Combining this inequality with the stated quadratic-growth condition proves
Eqs.~\eqref{eq:empirical_argmax_subspace_bound}
and~\eqref{eq:gmm_argmax_subspace_bound}. No stability of a particular optimal
transport plan is needed for these value and maximizer conclusions.

The local quadratic-growth condition means that moving a Grassmann distance
\(\varepsilon_{\mathrm{sub}}\) from the population optimum lowers the ratio by at
least \(\varrho_s\varepsilon_{\mathrm{sub}}^2/2\). A larger \(\varrho_s\) therefore
indicates a more sharply identified optimum. Such a condition holds locally if the
objective is twice Riemannian differentiable and has a negative-definite Hessian
on the Grassmann tangent space at an isolated maximizer. It need not hold for an
objective with a flat ridge or multiple distinct maximizing subspaces. Moreover,
the estimated maximizing subspace must lie in the stated neighborhood; that
localization is an additional premise, not a consequence of the ratio bound alone.
The pair-cost, ratio-error, and objective-regret bounds do not need this premise.
For an output that does not globally maximize \(\widehat r_s\), its gap to that
maximum must be added to the \(2\varepsilon_{r,s}\) objective-regret bound before
applying quadratic growth.

\subsection{Empirical Sample-Size Calculations}
\label{ap:empirical-sample-rates}
We use Proposition~\ref{prop:empirical_estimation_chain} to determine a sufficient
number \(m\) of retained observations per subject for a specified accuracy.
Let \(0<\varepsilon_{\mathrm{tar}}<1\) be the desired uniform \emph{ratio-error}
tolerance, rather than a bound on distribution error. The objective is to choose
\(m\) so that the ratio computed from the sampled observations differs from its
population counterpart by at most \(\varepsilon_{\mathrm{tar}}\) for every
orthonormal projection, with a prescribed probability.
Fix the number of subjects and a desired confidence level
\(1-\varepsilon_{\mathrm{fail}}\), where
\(0<\varepsilon_{\mathrm{fail}}<1\) is the allowed probability of failing to meet
the accuracy target. Assume a finite moment of order greater than four for each subject.
For fixed dimension \(d\), assume that the distance bound
\(D_{\mathsf{emp}}^{\mathsf{up}}\) and between-class variation bound
\(\zeta^{\mathrm{up}}_{B,\mathsf{emp}}\) are bounded above, and that the
within-class lower bound \(\nu_{\mathsf{emp}}\) is bounded below by a positive
constant, independently of \(m\) and \(\varepsilon_{\mathrm{tar}}\).
On the event that all subject-distribution errors are at most
\(\varepsilon_{\mathsf{emp}}\), also require
\(\zeta_{\mathsf{emp}}\leq\nu_{\mathsf{emp}}/2\).

The bounds of \citet{fournier2015rate}, applied to squared \(W_2\), imply
\begin{equation}
\begin{aligned}
\max_k W_2(\widehat{\mathcal P}_k,\mathcal P_k)
&=O_{\mathbb P}\{a_{\mathsf{emp}}(m,d)\},\\
a_{\mathsf{emp}}(m,d)&=
\begin{cases}
m^{-1/4},&d<4,\\
m^{-1/4}\{\log(1+m)\}^{1/2},&d=4,\\
m^{-1/d},&d>4.
\end{cases}
\end{aligned}
\label{eq:empirical_w2_rate}
\end{equation}
Here \(O_{\mathbb P}(a_m)\) means that, for any fixed confidence level, the random
error is at most a sufficiently large constant times \(a_m\) with at least that
probability for all sufficiently large \(m\). The constant may depend on the
confidence level, dimension, moments, and number of subjects; it is independent of
\(m\). Equivalently, the error divided by \(a_m\) is bounded in probability.
Thus one can choose \(\varepsilon_{\mathsf{emp}}\) proportional to
\(a_{\mathsf{emp}}(m,d)\) on an event of the desired probability.

To obtain a ratio-error bound from this distribution-error bound, first note
that \(\varepsilon_{\mathsf{emp}}\leq1\) implies
\(
\zeta_{\mathsf{emp}}
=4\varepsilon_{\mathsf{emp}}
(D_{\mathsf{emp}}^{\mathsf{up}}+\varepsilon_{\mathsf{emp}})
\leq4(D_{\mathsf{emp}}^{\mathsf{up}}+1)\varepsilon_{\mathsf{emp}}
\).
Also, \(\zeta_{\mathsf{emp}}\leq\nu_{\mathsf{emp}}/2\) gives
\(\nu_{\mathsf{emp}}-\zeta_{\mathsf{emp}}\geq\nu_{\mathsf{emp}}/2\).
Substituting these two inequalities into the formula for
\(\varepsilon_{r,\mathsf{emp}}\) in
Proposition~\ref{prop:empirical_estimation_chain} yields
\begin{equation}
\begin{aligned}
\varepsilon_{r,\mathsf{emp}}
&=\frac{\zeta_{\mathsf{emp}}
(\nu_{\mathsf{emp}}+\zeta^{\mathrm{up}}_{B,\mathsf{emp}})}
{\nu_{\mathsf{emp}}(\nu_{\mathsf{emp}}-\zeta_{\mathsf{emp}})}\\
&\leq\frac{2(\nu_{\mathsf{emp}}+\zeta^{\mathrm{up}}_{B,\mathsf{emp}})}
{\nu_{\mathsf{emp}}^2}\,\zeta_{\mathsf{emp}}\\
&\leq
\frac{8(D_{\mathsf{emp}}^{\mathsf{up}}+1)
(\nu_{\mathsf{emp}}+\zeta^{\mathrm{up}}_{B,\mathsf{emp}})}
{\nu_{\mathsf{emp}}^2}\,\varepsilon_{\mathsf{emp}}.
\end{aligned}
\label{eq:empirical_distribution_ratio_conversion}
\end{equation}
The coefficient multiplying \(\varepsilon_{\mathsf{emp}}\) is bounded above
under the assumptions stated at the start of this subsection.
Consequently, the uniform ratio error has order
\(O_{\mathbb P}\{a_{\mathsf{emp}}(m,d)\}\).
Choosing \(a_{\mathsf{emp}}(m,d)\) to be a sufficiently small multiple of
\(\varepsilon_{\mathrm{tar}}\) and solving for \(m\) gives, for sufficiently small
\(\varepsilon_{\mathrm{tar}}\), the sufficient condition
\begin{equation}
m\geq c_{\mathrm{conf}}
\begin{cases}
\varepsilon_{\mathrm{tar}}^{-4},&d<4,\\
\varepsilon_{\mathrm{tar}}^{-4}
\{\log(1/\varepsilon_{\mathrm{tar}})\}^{2},&d=4,\\
\varepsilon_{\mathrm{tar}}^{-d},&d>4.
\end{cases}
\label{eq:empirical_support_for_ratio_accuracy}
\end{equation}
Here \(c_{\mathrm{conf}}>0\) is a sufficiently large constant depending on the
chosen confidence level, the number of subjects, moment bounds, dimension, and the
conditioning constants just listed, but not on \(\varepsilon_{\mathrm{tar}}\).
With this choice, the sample-size condition ensures
\[
\mathbb P\!\left\{
\sup_A|\widehat r_{\mathsf{emp}}(A)-r_{\mathsf{emp}}^0(A)|
\leq\varepsilon_{\mathrm{tar}}\right\}
\geq1-\varepsilon_{\mathrm{fail}},
\]
where the probability is over the observations sampled for all subjects.
For example, the high-dimensional case solves
\(m^{-1/d}\leq\text{constant}\times\varepsilon_{\mathrm{tar}}\);
the borderline case inverts
\(m^{-1/4}\sqrt{\log(1+m)}\), producing the squared logarithmic factor.
Intrinsic dimension and multiscale structure can improve this ambient-dimensional
benchmark~\citep{weed2019sharp}.

For a desired subspace distance \(\varepsilon_{\mathrm{sub}}\), the condition
\(
\varepsilon_{r,\mathsf{emp}}\leq
\varrho_{\mathsf{emp}}\varepsilon_{\mathrm{sub}}^2/4
\)
implies the requested accuracy under the quadratic-growth and neighborhood
assumptions of Proposition~\ref{prop:empirical_estimation_chain}.
For curvature \(\varrho_{\mathsf{emp}}\) bounded away from zero and fixed distance
and conditioning bounds, inserting this ratio tolerance into
Eq.~\eqref{eq:empirical_support_for_ratio_accuracy} gives sufficient support-size orders
\(\varepsilon_{\mathrm{sub}}^{-8}\) for \(d<4\),
\(\varepsilon_{\mathrm{sub}}^{-8}\{\log(1/\varepsilon_{\mathrm{sub}})\}^2\)
for \(d=4\), and \(\varepsilon_{\mathrm{sub}}^{-2d}\) for \(d>4\).
These are conservative sufficient orders, not lower bounds. Furthermore, the
displayed dimension exponents do not account for any additional dimension dependence
in the moment or conditioning constants; those must also be controlled in a
comparison across dimensions.

\subsection{GMM Sample-Size Calculations}
\label{ap:gmm-sample-rates}

Let \(N_{\mathrm{cell}}\) be the number of observed cells used to fit each subject's
GMM, taken equal across subjects for this calculation. Retain the uniform
ratio-error target \(\varepsilon_{\mathrm{tar}}\) and confidence level
\(1-\varepsilon_{\mathrm{fail}}\) defined above.
Consider a correctly specified, well-separated spherical mixture whose component
weights are of order \(1/K\) and whose component standard deviations are bounded
above by a fixed constant. Assume the initialization and accuracy conditions of
\citet[Theorem~7]{kwon2020em}. For ratio stability, require
\(V_{W,\mathsf{gmm}}^0(A)\geq\nu_{\mathsf{gmm}}>0\), uniformly in \(A\),
and \(\zeta_{\mathsf{gmm}}\leq\nu_{\mathsf{gmm}}/2\) on the estimation event.
The component-distance bound \(D_{\mathsf{gmm}}\) and between-class variation
bound \(\zeta^{\mathrm{up}}_{B,\mathsf{gmm}}\) must be bounded above, and
\(\nu_{\mathsf{gmm}}\) must be bounded below by a positive constant,
independently of the target accuracy; these bounds must also remain uniform
when comparing different dimensions.

Under these assumptions, we prove that a sufficient fitting sample size per
subject has order
\begin{equation}
N_{\mathrm{cell}}=\widetilde O(Kd/\varepsilon_{\mathrm{tar}}^2).
\label{eq:gmm_sample_size_for_ratio_accuracy}
\end{equation}
This describes a sufficient fitting budget, not an upper limit on how many cells
may be used. The notation \(\widetilde O\) omits logarithmic factors; the
multiplicative constants depend on the chosen confidence level and the stated bounds.

More precisely, a sufficiently large sample size of this order ensures
\[
\mathbb P\!\left\{
\sup_A|\widehat r_{\mathsf{gmm}}(A)-r_{\mathsf{gmm}}^0(A)|
\leq\varepsilon_{\mathrm{tar}}\right\}
\geq1-\varepsilon_{\mathrm{fail}}.
\]

\begin{proof}
We prove Eq.~\eqref{eq:gmm_sample_size_for_ratio_accuracy} by relating the
parameter-accuracy tolerance \(\varepsilon_{\mathrm{fit}}\) in the cited
EM theorem to the desired ratio accuracy \(\varepsilon_{\mathrm{tar}}\).
Let \(p_{\min}=\min_{k,i}p_i^{(k)}\). Under the theorem's initialization,
separation, and sample-splitting conditions, a sufficient sample size has order
\[
\widetilde O\!\left(\frac{d}{p_{\min}\varepsilon_{\mathrm{fit}}^2}\right)
=\widetilde O\!\left(\frac{Kd}{\varepsilon_{\mathrm{fit}}^2}\right).
\]
The confidence level can be chosen so that the parameter bounds hold for all
subjects simultaneously with probability at least
\(1-\varepsilon_{\mathrm{fail}}\).
Write \(\Sigma_i^{(k)}=(\sigma_i^{(k)})^2 I_d\).
After matching components, the guarantee bounds mean error by
\(\sigma_i^{(k)}\varepsilon_{\mathrm{fit}}\), relative weight error by
\(\varepsilon_{\mathrm{fit}}\), and variance error by
\((\sigma_i^{(k)})^2\varepsilon_{\mathrm{fit}}/\sqrt d\).
Because
\[
|\widehat\sigma_i^{(k)}-\sigma_i^{(k)}|
=\frac{|(\widehat\sigma_i^{(k)})^2-(\sigma_i^{(k)})^2|}
{\widehat\sigma_i^{(k)}+\sigma_i^{(k)}},
\]
the standard-deviation error is at most
\(\sigma_i^{(k)}\varepsilon_{\mathrm{fit}}/\sqrt d\).
For spherical components, the exact formula then gives
\[
W^2(\widehat\phi_i^{(k)},\phi_i^{(k)})
=\|\widehat\mu_i^{(k)}-\mu_i^{(k)}\|^2
+d(\widehat\sigma_i^{(k)}-\sigma_i^{(k)})^2
\leq2(\sigma_i^{(k)})^2\varepsilon_{\mathrm{fit}}^2.
\]
Summing the relative weight errors also gives
\(\|\widehat p^{(k)}-p^{(k)}\|_1\leq\varepsilon_{\mathrm{fit}}\).
Let \(\bar\sigma\) be a fixed upper bound on all component standard deviations.
Then the preceding inequalities allow us to take
\(
\varepsilon_{\mathsf{gmm}}=\sqrt2\,\bar\sigma\,\varepsilon_{\mathrm{fit}}\),
\(\varepsilon_w=\varepsilon_{\mathrm{fit}}
\)
as the component-distance and weight-error bounds, respectively. Consequently,
\(
\varepsilon_{\mathsf{gmm}}+\varepsilon_w
=(\sqrt2\,\bar\sigma+1)\varepsilon_{\mathrm{fit}}
=O(\varepsilon_{\mathrm{fit}})
\).
With \(D_{\mathsf{gmm}}\) bounded, substituting these bounds into
Eq.~\eqref{eq:gmm_maw_error_constant} also gives
\(\zeta_{\mathsf{gmm}}=O(\varepsilon_{\mathrm{fit}})\).
As in Appendix~\ref{ap:empirical-sample-rates}, the denominator bound
\(\nu_{\mathsf{gmm}}-\zeta_{\mathsf{gmm}}\geq\nu_{\mathsf{gmm}}/2\)
then gives
\[
\sup_A|\widehat r_{\mathsf{gmm}}(A)-r_{\mathsf{gmm}}^0(A)|
\leq\varepsilon_{r,\mathsf{gmm}}
\leq\frac{2(\nu_{\mathsf{gmm}}+\zeta^{\mathrm{up}}_{B,\mathsf{gmm}})}
{\nu_{\mathsf{gmm}}^2}\,\zeta_{\mathsf{gmm}}
=O(\varepsilon_{\mathrm{fit}}).
\]
Thus there is a constant \(c_r>0\), independent of
\(\varepsilon_{\mathrm{fit}}\), such that the ratio error is at most
\(c_r\varepsilon_{\mathrm{fit}}\) on the same event.
Choosing \(\varepsilon_{\mathrm{fit}}\leq\varepsilon_{\mathrm{tar}}/c_r\)
ensures the desired ratio accuracy. For sufficiently small
\(\varepsilon_{\mathrm{tar}}\), taking
\(\varepsilon_{\mathrm{fit}}=\varepsilon_{\mathrm{tar}}/c_r\) also
satisfies \(\zeta_{\mathsf{gmm}}\leq\nu_{\mathsf{gmm}}/2\) and the
theorem's accuracy conditions. Substituting this choice into the sample-size
bound above proves Eq.~\eqref{eq:gmm_sample_size_for_ratio_accuracy}
and its stated probability guarantee.
\end{proof}

Equivalently, the same calculation gives the ratio-error rate
\begin{equation}
\sup_A|\widehat r_{\mathsf{gmm}}(A)-r_{\mathsf{gmm}}^0(A)|
=\widetilde O_{\mathbb P}\!\left(\sqrt{Kd/N_{\mathrm{cell}}}\right).
\label{eq:spherical_gmm_ratio_rate}
\end{equation}
The tilde suppresses logarithmic factors; \(O_{\mathbb P}\) retains the
fixed-confidence meaning explained above. If component weights are more
imbalanced, the dependence on \(1/p_{\min}\) must be retained.
Under local quadratic growth and localization of the fitted maximizer, a target
subspace error \(\varepsilon_{\mathrm{sub}}\) is certified by
Eq.~\eqref{eq:gmm_argmax_subspace_bound} when the ratio error is at most
\(\varrho_{\mathsf{gmm}}\varepsilon_{\mathrm{sub}}^2/4\).
Substituting this ratio tolerance into
Eq.~\eqref{eq:gmm_sample_size_for_ratio_accuracy} gives the sufficient order
\(\widetilde O(Kd/\varepsilon_{\mathrm{sub}}^4)\) when
\(\varrho_{\mathsf{gmm}}\) is bounded away from zero.
This subspace conclusion additionally requires the fitted maximizing subspace to
lie in the neighborhood specified in the proposition.

For full covariances, suppose a separated, identifiable estimator provides
\[
\varepsilon_{\mathsf{gmm}}+\varepsilon_w
=\widetilde O_{\mathbb P}\!\left(\sqrt{Kd^2/N_{\mathrm{cell}}}\right).
\]
The same ratio perturbation bound then gives a sufficient ratio-accuracy sample
order \(\widetilde O(Kd^2/\varepsilon_{\mathrm{tar}}^2)\), and, under the additional
subspace conditions, \(\widetilde O(Kd^2/\varepsilon_{\mathrm{sub}}^4)\).
The density-learning theorem of \citet{ashtiani2018nearly} concerns total variation;
it does not by itself establish this matched-component \(W_2\) premise.
This full-covariance calculation therefore remains conditional on an appropriate
parameter-estimation guarantee. Neither benchmark applies automatically to poorly
separated or misspecified mixtures.

\section{Simulation Designs and Supplementary Representation Analyses}
\label{ap:optimization_simulations}

This appendix specifies three known-GMM benchmarks: the 1,440-dataset optimizer
comparison (Section~\ref{sec:two_dimensional_optimizer_study}), the 1,920-dataset
initialization comparison (Section~\ref{sec:initialization_diagnostics_short}),
and the 192-dataset higher-dimensional comparison
(Section~\ref{sec:highdim_optimizer_study}). It also gives the simulation and
population-evaluation details for
Section~\ref{sec:representation_diagnostics_rev3}. In the three optimization and
initialization benchmarks, the mixture parameters are
generated directly: no cells are sampled and no GMM is fitted. Each dataset has
six subjects, three in each of two classes. Write \(y_k=2Y_k-3\), so that
\(y_k=-1\) for subjects \(k=1,2,3\) and \(y_k=1\) for \(k=4,5,6\).
All nine unordered between-class pairs and six unordered within-class pairs
enter the corresponding normalized variations. The separate fitted-GMM
classification simulation used for the short-budget comparison is described in
Appendix~\ref{sec:sim}.

\subsection{Two-Dimensional GMMs}
\label{ap:optimization_gmm_2d}

Each subject has two Gaussian components. Let \(e_1,e_2\) be the coordinate
vectors in \(\mathbb R^2\), and denote the five design parameters by
\(b_{\mathrm{mean}}\) (class-mean offset), \(c_{\mathrm{cov}}\) (covariance scale),
\(\kappa\) (anisotropy), \(\theta_{\mathrm{cls}}\) (class-dependent orientation
shift), and \(w_{\mathrm{imb}}\) (mixture-weight imbalance).
For each subject $k$, draw a vector \(u_k\sim N(0,I_2)\) and a scalar
\(z_k^{\mathrm{wt}}\sim N(0,1)\). For each component \(i=1,2\), draw
\(v_{ki}\sim N(0,I_2)\) and scalars
\(z_{ki}^{\mathrm{ang}},z_{ki}^{\mathrm{sc}}\sim N(0,1)\).
These random variables are mutually independent within a dataset. Set
\(\mu_i^{(k)}
=y_k b_{\mathrm{mean}}e_1+0.75(-1)^i e_2+0.12u_k+0.04v_{ki}\),
and
\(p_1^{(k)}=\min\!\left\{0.95,\max\!\left\{0.05,
0.5+y_k w_{\mathrm{imb}}+0.02z_k^{\mathrm{wt}}\right\}\right\}\),
\(p_2^{(k)}=1-p_1^{(k)}\).
The component covariances are
\[
\begin{aligned}
\theta_{ki}&=(-1)^i\pi/6+y_k\theta_{\mathrm{cls}}/2
                 +0.06z_{ki}^{\mathrm{ang}},\\
\Sigma_i^{(k)}&=c_{\mathrm{cov}}\exp(0.12z_{ki}^{\mathrm{sc}})
R(\theta_{ki})
\begin{pmatrix}\kappa&0\\0&\kappa^{-1}\end{pmatrix}
R(\theta_{ki})^\top,
\qquad
R(\theta)=\begin{pmatrix}\cos\theta&-\sin\theta\\
\sin\theta&\cos\theta\end{pmatrix}.
\end{aligned}
\]
Thus the class-mean parameter gives component offsets
\(\pm b_{\mathrm{mean}}e_1\); corresponding unperturbed component centers
in the two classes differ by \(2b_{\mathrm{mean}}e_1\).
The covariance eigenvalue ratio is \(\kappa^2\).
The generated subject distribution is
\(\mathcal G_k^0=\sum_{i=1}^2p_i^{(k)}N(\mu_i^{(k)},\Sigma_i^{(k)})\).
The implementation symmetrizes each covariance to remove floating-point asymmetry.

For the 1,440-dataset study, the level sets are
\(b_{\mathrm{mean}}\in\{0.35,1,2\}\),
$c_{\mathrm{cov}}\in\{0.05,0.5,2\}$, $\kappa\in\{1,9\}$, 
$\theta_{\mathrm{cls}}\in\{0,\pi/4\}$,
$w_{\mathrm{imb}}\in\{0,0.25\}$.
Each of the 72 parameter settings has 20 random replicates, giving
1,440 datasets. For the initialization study, the class-mean offset
set is expanded to \(\{0.1,0.35,1,2\}\), giving 96 settings, each
evaluated with 20 new random replicates, for a total of 1,920 datasets.
Within each study, corresponding replicates share the same Gaussian
perturbations across parameter settings.
Reusing a seed across settings reuses the Gaussian
perturbations; these are factorial comparisons, not 1,440 or 1,920 independently
seeded datasets. Within each dataset, subjects are generated in increasing
\(k\). The draws for a subject occur in the order \(u_k\),
\(z_k^{\mathrm{wt}}\), followed for each component in increasing \(i\) by
\(v_{ki},z_{ki}^{\mathrm{ang}},z_{ki}^{\mathrm{sc}}\).

In the initialization study, the ambient-coupling, pooled-moment PCA, and
subject-mean LDA directions are constructed directly from the generated GMM
parameters. For ambient-coupling initialization, the within-class matrix is
\(C_{W,\mathrm{ind}}(\boldsymbol\Gamma_{\mathrm{amb}})\) from
Eq.~\eqref{eq:spectral-scatter-matrices}, where
\(\boldsymbol\Gamma_{\mathrm{amb}}\) contains the optimal component couplings
computed in the original two-dimensional space.
For subject-mean LDA, we represent each subject by its mixture mean and
average these vectors within each class. The between-class matrix is the
outer product of the difference between the two class averages. The
within-class matrix is the sum of the outer products of the class-centered
subject mean vectors, divided by \(6-2=4\).

In this benchmark, each of these two generalized
eigenproblems adds to every diagonal entry \(0.001\) times the mean diagonal
entry of its own within-class matrix. These matrices construct the initial
directions before the optimization updates begin.
The eight random candidates are normalized independent standard-Gaussian vectors
in \(\mathbb R^2\). They are generated separately from the data, using seed
\(310000+100j+r\), where \(j\) is the one-based parameter-setting index and
\(r=1,\ldots,20\) is the replicate index. Settings are ordered with
\(b_{\mathrm{mean}},c_{\mathrm{cov}},\kappa,\theta_{\mathrm{cls}},w_{\mathrm{imb}}\)
varying from slowest to fastest in their listed order.
Directions are normalized and sign-canonicalized; screening selects the first
maximum in the order ambient-coupling, PCA, LDA, and the eight random candidates.
The gradient-only paths from all 11 starts are stored to evaluate both
initial-value screening and the more expensive terminal-value portfolio.
The gradient-only and Screened GeoGrad optimizers both use at most 50
outer updates. The reference for this study is grid-only, as specified below.

\subsection{Higher-Dimensional GMMs}
\label{ap:optimization_gmm_highd}

The higher-dimensional generator uses three components per subject and a
different mean and covariance construction. Let \(d\in\{8,32,64\}\),
\(q\in\{1,3\}\), and \(b_{\mathrm{mean}}\in\{1,2\}\). Here \(q\) sets both
the retained dimension and the dimension of the generating mean signal; changing
\(q\) therefore changes the generated dataset, not merely its projection.
Write \(E_q=(e_1,\ldots,e_q)\in\mathbb R^{d\times q}\), and let \(H_q\)
contain the first \(q\) rows of
\[
H=\begin{pmatrix}
-0.85&0.15&0.95\\
0.55&-0.75&0.20\\
-0.35&0.65&-0.10
\end{pmatrix}.
\]
For independent standard-Gaussian vectors \(u_k,v_{ki}\in\mathbb R^q\), set
\[
\mu_i^{(k)}=E_q(H_q)_{\cdot i}
+y_k b_{\mathrm{mean}}\bigl(e_1+0.25\mathbf1_{\{q=3\}}e_2\bigr)
+0.08E_qu_k+0.03E_qv_{ki}.
\]
All mean coordinates beyond \(q\) are zero. For an independent
\(z_k\sim N(0,I_3)\), define the weight vector by
\[
\widetilde p^{(k)}=\max\!\left\{
\begin{pmatrix}0.23\\0.34\\0.43\end{pmatrix}
+y_k\begin{pmatrix}0.035\\-0.015\\-0.020\end{pmatrix}
+0.012z_k,\ 0.04\mathbf1_3\right\},\qquad
p_i^{(k)}=\frac{\widetilde p_i^{(k)}}{\sum_{j=1}^3\widetilde p_j^{(k)}},
\]
where the maximum is taken componentwise.

For the covariance construction, let
\(
\upsilon_\ell=0.2\,9^{(\ell-1)/(d-1)}\),
\(\upsilon_{i\ell}=\upsilon_{\,1+\operatorname{mod}(\ell-1+2(i-1),d)}\), $\ell=1,\ldots,d$.
The diagonal-covariance setting uses
\(T_i=\operatorname{diag}(\upsilon_{i1},\ldots,\upsilon_{id})\).
For full covariances, draw a separate \(d\times d\) matrix with independent
standard-Gaussian entries for each component, take the orthogonal factor \(R_i\)
of its QR factorization, and set
\(T_i=R_i\operatorname{diag}(\upsilon_{i1},\ldots,\upsilon_{id})R_i^\top\).
With independent \(z_{ki}^{\mathrm{sc}}\sim N(0,1)\), define
\(
\Sigma_i^{(k)}=\exp(0.035z_{ki}^{\mathrm{sc}})T_i
\).
The three component covariance shapes are thus shared across subjects, with
independent subject--component scale perturbations; their eigenvalue ratio is nine.
The subject distribution is
\(\mathcal G_k^0=\sum_{i=1}^3p_i^{(k)}N(\mu_i^{(k)},\Sigma_i^{(k)})\).

Full-covariance rotations are
drawn first. Subjects then follow class \(-1\) before class \(+1\), with three
subjects per class. For each subject, draws occur in the order subject-mean
jitter, the \(q\times3\) component-mean jitter matrix, the three weight
perturbations, and the three covariance-scale perturbations.
Seeds 1 and 2 supply calibration datasets and seeds \(3\sim10\) supply evaluation datasets
in each of the \(3\times2\times2\times2=24\) parameter settings.
Seeds are reused across settings, so the 192 evaluation datasets are not
192 independently seeded draws. In particular, changing only
\(b_{\mathrm{mean}}\) retains the same random perturbations and changes only the
class shift. Diagonal and full-covariance settings consume different numbers of
random draws, so their subject-level perturbations need not agree.

To compare the optimization updates without also comparing initialization
strategies, all optimizers use the same random initial projection \(A_0\) on
each dataset. We sample the entries of a \(d\times q\) matrix independently
from \(N(0,1)\), compute its thin QR factorization, and take the \(q\)
orthonormal columns of its \(Q\) factor as \(A_0\). This matrix is generated
separately from the data, using seed \(970000+100d+10q+r\), where \(r\) is
the data-generation seed. 

\subsection{Numerical References and Update Counts}
\label{ap:optimization_references}

For the 1,440-dataset comparison, the reference first evaluates the
transport-reoptimized ratio at the 360 angles
\(0^\circ,0.5^\circ,\ldots,179.5^\circ\).
An angle is a grid-local maximum when its value is no smaller than the values
at its two cyclic neighbors. Every such local maximum at \(\theta_j\) is refined by
minimizing the negative ratio over
\([\theta_j-0.5^\circ,\theta_j+0.5^\circ]\), using MATLAB's
\texttt{fminbnd} with angular tolerance \(10^{-10}\) degrees.
The largest value among all grid and refined candidates defines
\(F_{\mathrm{ref}}^*\). 
For the 1,920-dataset initialization study,
the reference is the largest value on the same 360-point grid without local
refinement. 

For Table~\ref{tab:optimizer-oracle-rev3}, one attempted update means one outer
iteration in which a projection update is attempted. It includes all inner
backtracking trials and, for double-bound, both proposal constructions and both
backtracking searches. If an initial geometric proposal is rejected and gradient
ascent is attempted immediately at the unchanged projection, these operations
count as one outer update. A gradient-norm check that terminates the method before
any proposal is attempted does not itself add an attempt.
The reported per-update statistic is the median of each dataset's total optimizer
time divided by its number of attempted updates, with divisor one if that count
is zero. 
The total optimizer time includes the initial exact evaluation and geometric-cache
setup when needed, but excludes dataset generation, reference searches,
post-run diagnostic checks, and downstream classification.

\subsection{Calibration of the Higher-Dimensional Iteration Budgets}
\label{ap:optimization_budgets}

Calibration is separate for each \((d,q)\). For each such setting, the two
types of covariance, two class-mean offsets, and seeds 1 and 2 provide eight pilot
datasets, which are excluded from evaluation. A reference scheme used to set
the calibration target considers the gradient candidate and both spectral candidates at each
update and accepts the admissible candidate with the largest exact objective.
The common time target is the median total runtime of this scheme with a
ten-update cap on the eight pilot datasets, after code-path warm-up.

The double-bound cap is the integer from 1 to 100 whose predicted runtime is
closest to that target. The prediction adds the median setup-plus-initialization
time to the candidate cap times the median pilot time per outer iteration.
For gradient ascent, complete pilot trajectories are timed at candidate caps
\(\{10,15,20,25,30,35,40\}\), and the cap with median total time closest to the
same target is chosen. These complete-trajectory timings replace the initial
linear extrapolation for gradient, which underestimated the cost of longer runs.
Only pilot data enter cap selection; the frozen caps are then applied to seeds
\(3\sim10\). The resulting caps are
\[
\begin{array}{cc|rr}
d&q&\text{Gradient cap}&\text{Double-bound cap}\\\hline
8&1&25&34\\
8&3&25&23\\
32&1&15&19\\
32&3&25&21\\
64&1&20&19\\
64&3&25&22
\end{array}
\]
The budgets are calibrated targets, not enforced equal wall-clock times.
All 192 double-bound trajectories stop before their caps: 95 because both
proposals are rejected and 97 because the accepted subspace change falls below
\(10^{-3}\) radians. The reported \(14.82\%\) shortfall averages the paired
relative differences using the gradient objective as denominator; it is not the
relative difference between two pooled mean objectives. 

\subsection{Data Generation and Representation Construction}
\label{ap:representation_simulation}

The representation comparison in Section~\ref{sec:representation_diagnostics_rev3}
uses the same GMM-generating family as
Appendix~\ref{ap:optimization_gmm_highd}, with a two-coordinate mean signal.
There are six subjects, three per class, and three components per subject.
Let \(E_2=(e_1,e_2)\), and let \(H_2\) contain the first two rows of the
template matrix \(H\) defined there. For \(y_k\in\{-1,+1\}\), the means are
\(
\mu_i^{(k)}=E_2(H_2)_{\cdot i}
 +y_k b(e_1+0.25e_2)+0.08E_2u_k+0.03E_2v_{ki}\),
$i=1,2,3$,
where \(u_k,v_{ki}\) are independent \(N(0,I_2)\) vectors. All other mean
coordinates are zero. Mixture weights, diagonal or rotated full covariance
shapes, and subject--component covariance-scale perturbations are generated
exactly as in Appendix~\ref{ap:optimization_gmm_highd}.
The generating mean-signal dimension is two; it is distinct from the retained
dimension \(q=1\) used in the representation comparison. The earlier
higher-dimensional optimizer comparison instead sets both dimensions to one
or three. 

For the comparison retained in Tables~\ref{tab:geograd_repr_time}
and~\ref{tab:geograd_population_values}, \(b=1\),
\(d\in\{16,64,128,256,512\}\), and the three generating seeds are
\(100101\sim100103\). These seeds are reused across dimension--covariance settings;
the settings are not 30 independently seeded draws. For each subject, we sample
1536 component labels using its mixture weights, then sample a Gaussian cell
conditional on each label. Cell sampling for subject \(k\) uses seed
\(
\operatorname{mod}\bigl((r+710000)1000003+7919k,\,2^{32}-1\bigr),
\)
where \(r\) is the generating seed. The first 32, 64, or 128 cells form the nested,
equally weighted empirical representations. Each GMM fit uses all 1536 cells.

GMM construction uses MATLAB \texttt{kmeans} with \texttt{Start='plus'},
five restarts, at most 200 iterations, and \texttt{EmptyAction='singleton'}.
The best clustering assigns each cell to one cluster. For a cluster containing
\(n_i\) cells, its mixture weight is \(n_i/1536\), its mean is the cell average,
and its unregularized covariance \(S_i\) is the average centered outer product,
with divisor \(n_i\), not \(n_i-1\). For diagonal fits, the off-diagonal entries
of \(S_i\) are discarded. We then use
\(
\widehat\Sigma_i=S_i+\max\{10^{-6},10^{-3}\operatorname{tr}(S_i)/d\}I_d
\).
The covariance structure is supplied from the generating setting rather than
selected from the sample. This is a hard-clustering, moment-based GMM estimate;
there is no subsequent EM refinement. For GMM choice \(j=1,2,3\), corresponding to
\(K=2,3,6\), its subject-\(k\) clustering seed is
\(
\operatorname{mod}\bigl((r+20000d+10N_{\mathrm{cell}}+j)10007+1009k,
\,2^{32}-1\bigr)\).
The random initialization candidates use seed
\(r+10000d+N_{\mathrm{cell}}\), shared by all six representations in a setting.
All methods use the same nine between-class and six within-class pairs,
Screened GeoGrad with \(q=1\), and at most ten recorded projections including
initialization. Matching seeds and iteration limits does not match the
estimation errors or the achieved runtimes.

Representation-construction time includes clustering and moment estimation for
GMMs, or extracting empirical supports, as well as arranging their parameters
or support coordinates and weights into the optimizer's input arrays.
Arranging these arrays is data formatting, not an additional estimation stage.
Optimizer time covers the complete call, including the
ambient-coupling initialization, candidate screening, geometric-covariance
cache, geometric step, gradient updates, and transport-reoptimized objective
checks. The order of the six representations is randomized within each setting.
Synthetic-data generation, subsequent population evaluation, independent
validation, file operations, and classification are outside the timed intervals.

\subsection{Population Objectives at Learned Directions}
\label{ap:population_direction_evaluation}

We evaluate every learned direction \(\widehat a\) on the same generating
distributions \(\mathcal G_k^0\). For \(F_{\mathsf{gmm}}^0(\widehat a)\),
project their true component means and covariances, solve the component-coupling
problem for each subject pair, and form the between-to-within ratio.
For \(F_{\mathrm W}^0(\widehat a)\), defined in
Eq.~\eqref{eq:representation_population_ratio}, the pairwise distances instead
allow unrestricted transport between the full projected mixtures.
This evaluation does not depend on whether \(\widehat a\) was obtained using
a fitted GMM or empirical supports. In particular, no GMM is fitted to the
output of an empirical-input optimizer.

Projection to one dimension makes the evaluation of \(F_{\mathrm W}^0(\widehat a)\) feasible through
quantile functions. The distribution function of subject \(k\)'s projected
generating mixture is
\[
\mathsf H_{k,a}(z)=\sum_{i=1}^m p_i^{(k)}
\Phi\!\left(\frac{z-a^\top\mu_i^{(k)}}
{\sqrt{a^\top\Sigma_i^{(k)}a}}\right),
\]
where $m=3$ is the number of components in our simulation and \(\Phi\) is the standard-normal distribution function. Its inverse
\(\mathsf H_{k,a}^{-1}\) is the quantile function. The one-dimensional
Wasserstein identity gives
\[
W^2(a_{\#}\mathcal G_k^0,a_{\#}\mathcal G_\ell^0)
=\int_0^1
\bigl[\mathsf H_{k,a}^{-1}(u)-\mathsf H_{\ell,a}^{-1}(u)\bigr]^2\,du.
\]
Thus we can evaluate the distance between the full projected mixtures without
solving a continuous transport problem in the original high-dimensional space.

We approximate this integral by dividing \((0,1)\) into
\(n_{\mathrm{quad}}=8192\) equal intervals and evaluating at their midpoints,
\(u_j=(j-1/2)/n_{\mathrm{quad}}\), $j=1,\ldots,n_{\mathrm{quad}}$.
Using the same probabilities for both subjects gives
\[
W^2(a_{\#}\mathcal G_k^0,a_{\#}\mathcal G_\ell^0)
\approx\frac{1}{n_{\mathrm{quad}}}
\sum_{j=1}^{n_{\mathrm{quad}}}
\bigl[\mathsf H_{k,a}^{-1}(u_j)-\mathsf H_{\ell,a}^{-1}(u_j)\bigr]^2.
\]

To find \(\mathsf H_{k,a}^{-1}(u_j)\), we numerically solve
\(\mathsf H_{k,a}(z)=u_j\) for \(z\) using the bisection method. 
At each learned direction \(a=\widehat a\), we substitute these approximate
squared distances into the between-class and within-class averages in
Eq.~\eqref{eq:representation_population_ratio}. Dividing the between-class
average by the within-class average gives the reported numerical value of
\(F_{\mathrm W}^0(\widehat a)\). Population evaluations
cover all five dimensions \(d\in\{16,64,128,256,512\}\). Comparing 4096 and
8192 quadrature points changes the population Wasserstein ratio by at most
0.026\% across the directions reported in
Table~\ref{tab:geograd_population_values}.

\section{Simulation of Classification Performance}
\label{sec:sim}

\begin{figure}[!htbp]
	\centering
    \begin{subfigure}[b]{0.48\textwidth}
         \centering
         \includegraphics[height=0.90\textwidth]{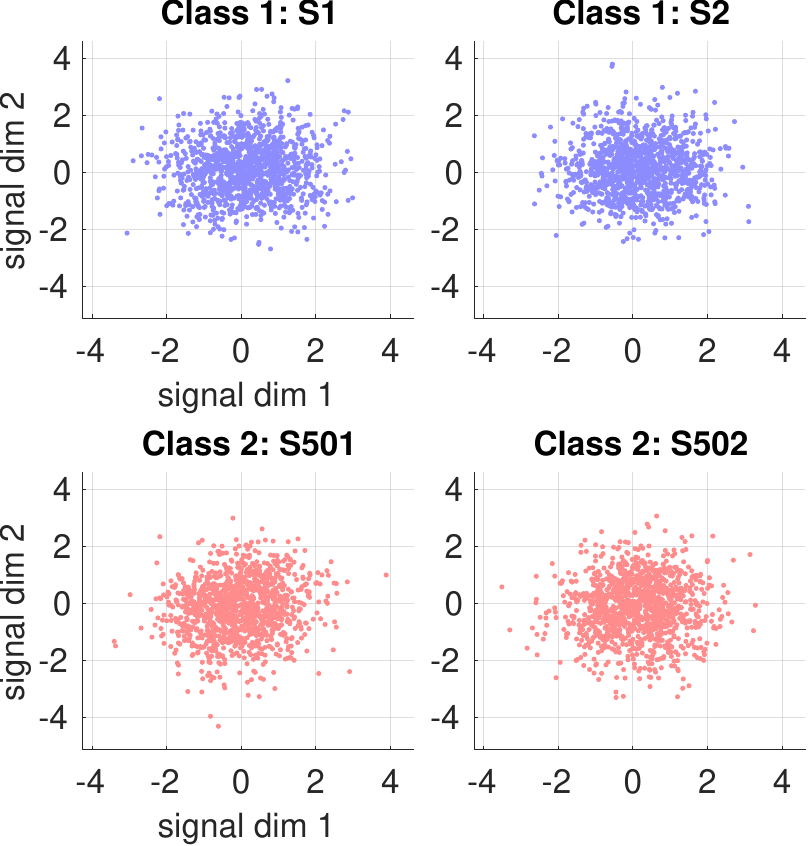}\\
         (a)
    \end{subfigure}
	\begin{subfigure}[b]{0.48\textwidth}
         \centering
         \includegraphics[height=0.90\textwidth]{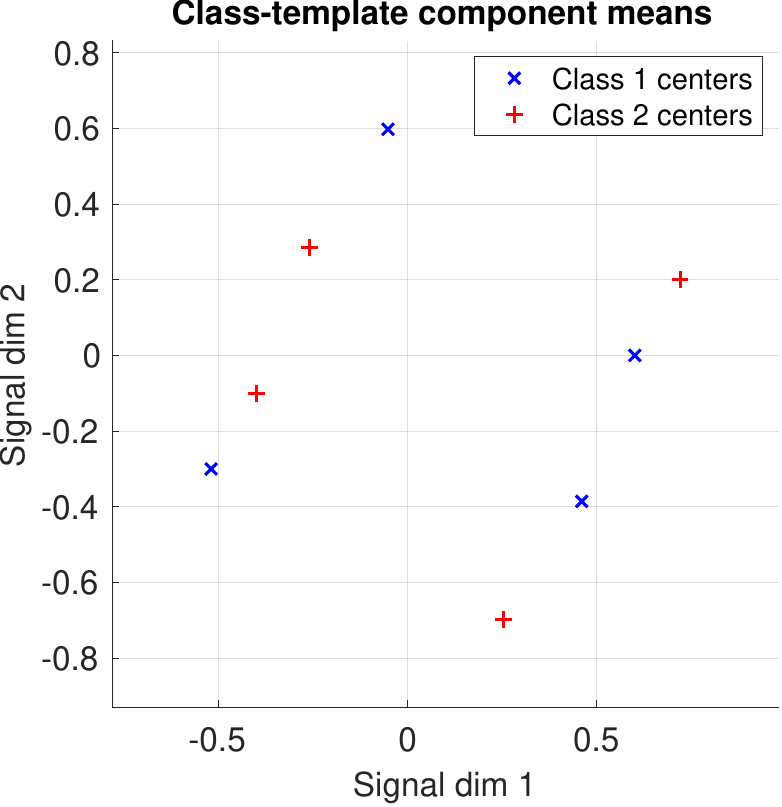}\\ (b)
    \end{subfigure}
	\caption{Illustration of the simulated data before
dataset-level rotation. (a) Scatter plots of the two signal coordinates for
example subjects. The top row shows two subjects from Class 1, and the bottom
row shows two subjects from Class 2. (b) Signal-coordinate component means
of the two class templates. Subject-specific Gaussian perturbations of these
means produce the within-class variation.}
	\label{fig:simulate}
\end{figure}

Complementing the representation and optimization diagnostics in
Section~\ref{sec:representation_diagnostics_rev3}, this simulation examines
classification and support-size-dependent runtime in moderate ambient dimension.
Each subject contains 1,200 observations in \(d=8\). The first two unrotated
coordinates contain the class signal; the remaining six have the same
distribution in both classes. For subject \(k\), class \(c\), and component
\(\ell\in\{1,2,3,4\}\),
\[
\Pr(Z_j^{(k)}=\ell)=p_\ell^{(k)},\qquad
x_j^{(k)}\mid Z_j^{(k)}=\ell\sim N_8\!\left(
\begin{bmatrix}m_{c\ell}+\eta_{k\ell}\\0_6\end{bmatrix},\Sigma_\ell\right),
\]
where \((p_1^{(k)},p_2^{(k)},p_3^{(k)},p_4^{(k)})=(0.4,0.3,0.2,0.1)\)
for every subject \(k\),
\(\eta_{k\ell}\stackrel{\mathrm{ind}}{\sim}N_2(0,0.1^2I_2)\), and
\[
\begin{split}
m_{1\ell}&=0.6(\cos\theta_\ell,\sin\theta_\ell)^\top,
\quad \theta=(0^\circ,95^\circ,210^\circ,320^\circ),\\
m_{2\ell}&=-m_{1\ell}+(0.2,-0.1)^\top,\\
\Sigma_\ell&=\operatorname{diag}(v_\ell,v_\ell,1.1,1.1,1.1,1.1,1.1,1.1),
\quad v=(0.8,1.0,0.6,0.7).
\end{split}
\]
Figure~\ref{fig:simulate} illustrates the signal coordinates. We first generate
500 subjects per class, each with 1,200 cells, and store this collection as a
\emph{finite pool}. We then form 20 datasets by uniformly selecting 30 Class-1
and 20 Class-2 subjects without replacement within each dataset. This
subject-level subsampling selects whole subjects together with their recorded
cells; it does not select the support points of their empirical representations.
Reusing the pool provides repeated subsets of an available cohort rather than
20 independently generated cohorts. A single random rotation is
shared by all subjects within each dataset, hiding the common signal subspace
without destroying it.

Separately, each empirical representation is formed by uniformly selecting
\(m=50\), 100, or 200 of a subject's 1,200 cells without replacement and assigning
each retained cell mass \(1/m\). GMMs are fitted to all 1,200 cells. This mimics
practice, where both representations are constructed from a finite observed
sample: subsampling reduces transport computation without acquiring new cells.
The original cells are independent draws from the subject's GMM. Because selection
does not depend on their values, the subsample has the same distribution as
\(m\) fresh draws when averaging over the original cell generation.
Conditional on the recorded cells, however, it is sampling
without replacement from that fixed collection. Drawing fresh cells directly
from the GMM would also be valid; using the observed sample is a design choice,
not a requirement of OTAF.

\begin{figure}[!htb]
	\centering
    \begin{subfigure}[b]{0.48\textwidth}
         \centering
         \includegraphics[width=0.98\textwidth]{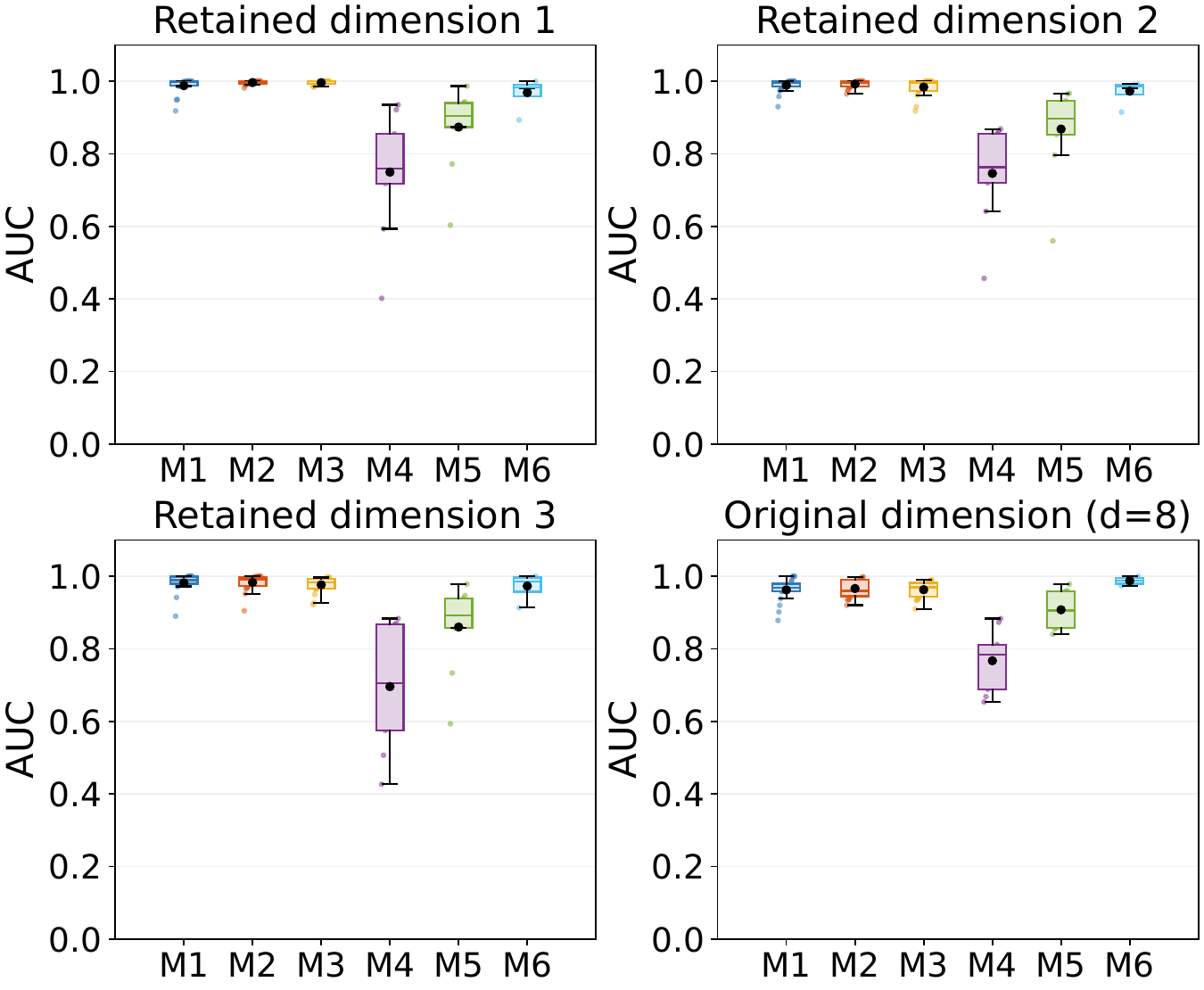}\\
         (a) AUC
    \end{subfigure}
	\begin{subfigure}[b]{0.48\textwidth}
         \centering
         \includegraphics[width=0.98\textwidth]{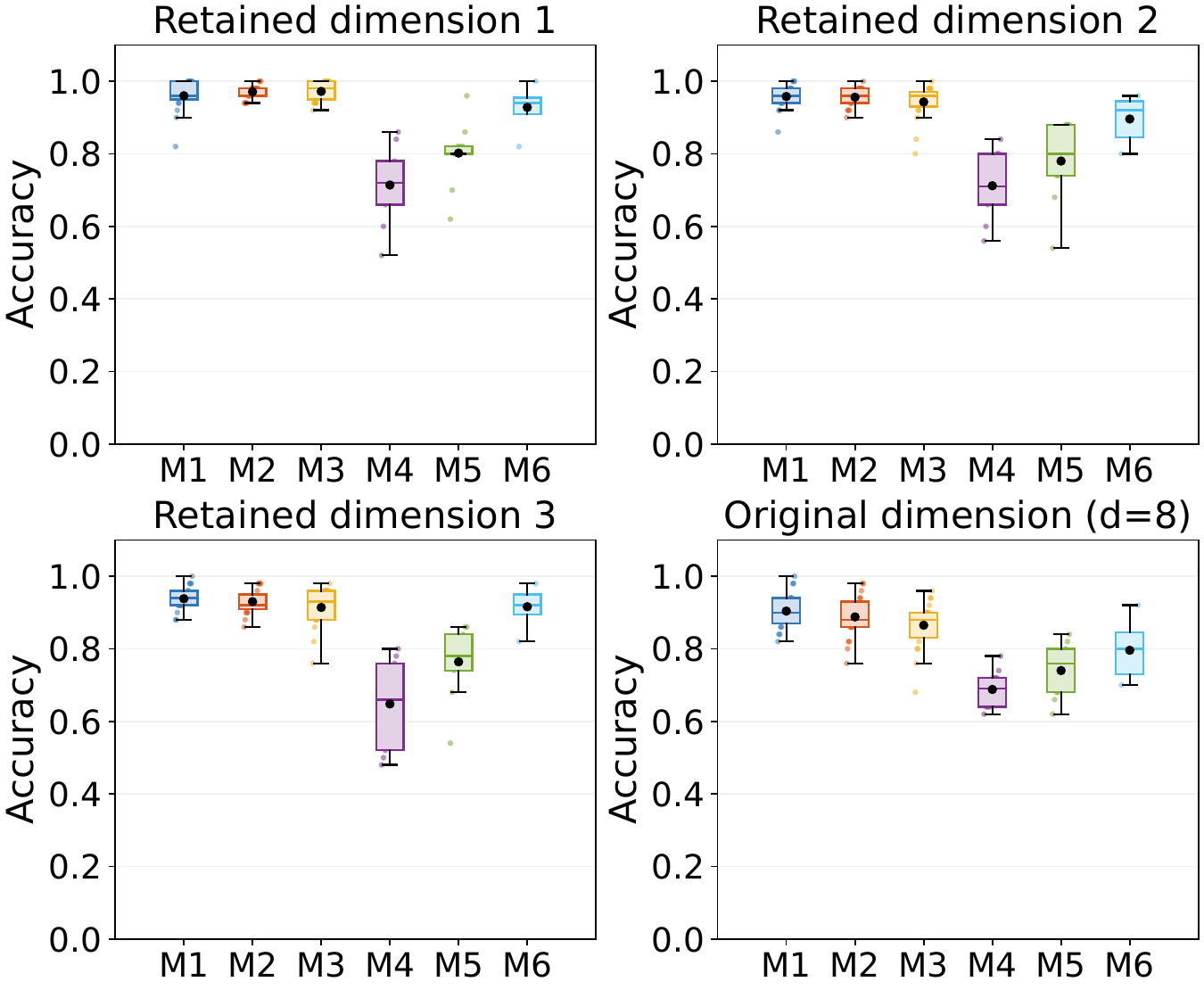}\\ (b) Classification accuracy
    \end{subfigure}
	\caption{Classification performance for the two
distributional representations: (a) AUC and (b) accuracy. Within each
subfigure, panels show retained dimensions 1, 2, and 3 and the original
eight-dimensional space. M1--M3 use fitted GMMs with \(K=3,4,6\); M4--M6 use
empirical representations with \(m=50,100,200\) support points. All
reduced-dimension results use Screened GeoGrad; all panels use PMM.
Each box summarizes 20 datasets for M1--M3, 10 for M4--M5, and five for M6;
its black dot marks the mean.}
	\label{fig:simulate_aucacc}
\end{figure}

The fitted representations are full-covariance GMMs with \(K=3,4,6\), using
five starts, covariance regularization \(10^{-5}\), and at most 1,000 EM
iterations, requiring convergence. Within each dataset, one fixed ten-fold
assignment, with five subjects per fold, is shared across representations and
retained dimensions. Figure~\ref{fig:simulate_aucacc} compares original-space
PMM with Screened GeoGrad followed by PMM at \(q=1,2,3\) for both GMM and
empirical inputs. Each reduced-space fit uses the confusing-pair rule, screens
11 initial projections, and permits one safeguarded geometric step followed by
gradient refinement, with at most ten recorded projections including
initialization. For point masses, the geometric and independent Gaussian bounds
coincide with the exact quadratic fixed-coupling cost.

GMM results use all 20 datasets. To limit computation, empirical inputs use
the first 10 datasets for \(m=50,100\) and the first five for \(m=200\),
chosen before observing their results. Within each subject, the smaller
empirical supports are nested within the larger ones. AUC is computed from
the 50 pooled held-out scores in each dataset, and accuracy from the
corresponding predictions. Comparisons of reduced and original dimensions
use the same datasets and folds; comparisons across representation sizes
must also account for the unequal numbers of datasets.

For \(K=3,4,6\), one-dimensional Screened GeoGrad gives mean AUC
\(0.988,0.996,0.996\) and accuracy \(0.960,0.971,0.972\), compared with
original-space AUC \(0.962\sim0.966\) and accuracy \(0.865\sim0.904\).
The corresponding mean AUC gains are \(0.025,0.030,0.033\).
Dimensions two and three also improve mean performance for every \(K\), but
less consistently across individual datasets. For empirical inputs, mean AUC
is lower after projection at every support size, whereas mean accuracy usually
increases. For example, at \(m=200\), reducing to \(q=1\) changes mean
AUC/accuracy from \(0.987/0.796\) to \(0.969/0.928\) over the five evaluated
datasets. Thus the benefit of dimension reduction depends on the representation,
retained dimension, and performance measure.

\begin{figure}[!htbp]
\centering
\includegraphics[width=0.65\textwidth]{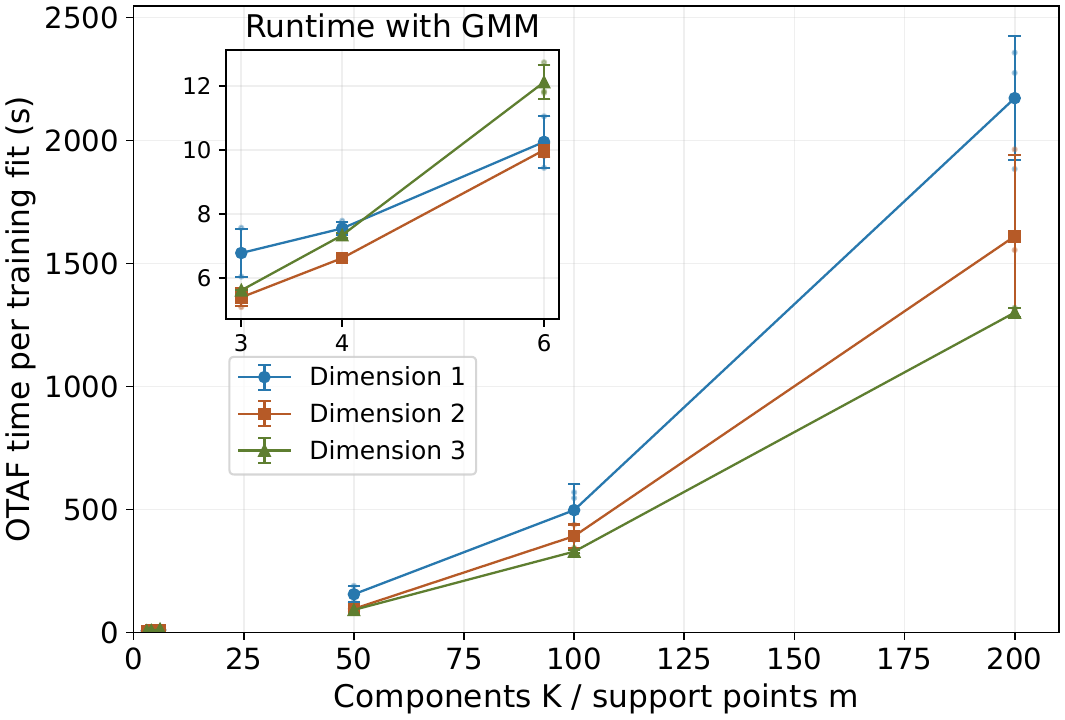}
\caption{Screened GeoGrad runtime versus representation size (Table~\ref{tab:runtime}).
Small points are dataset times; larger symbols and whiskers are means and sample
standard deviations. The inset expands the GMM range. Lines connect observed
means, not fitted scaling laws. The series retain dimensions 1, 2, and 3.}
\label{fig:time}
\end{figure}

\paragraph{Support-size scaling.}
Table~\ref{tab:runtime} and Figure~\ref{fig:time} compare Screened GeoGrad on fitted GMMs with \(K=3,4,6\) and empirical representations
with \(m=50,100,200\) support points per subject. We use datasets \(1\sim3\) of the
preceding simulation and their first training fold, containing 45 subjects.
The empirical supports are sampled uniformly without replacement from each
subject's 1,200 cells, with smaller supports nested within larger ones.
The confusing-pair rule in Section~\ref{sec:objective_hierarchy} is applied separately to each
representation, so the selected subject pairs need not be identical.
Each Screened GeoGrad fit screens 11 initializations and permits at most ten recorded
projections, including initialization. For point masses, both Gaussian-coupling
bounds coincide with the exact quadratic fixed-coupling cost. Backtracking
counts can differ even under the same iteration limit.

For Figure~\ref{fig:time}, the timed interval is the complete call to Screened GeoGrad, starting from already constructed subject representations. The following steps are included:
selection of confusing subject pairs; the full-dimensional
transport calculations needed for the ambient-coupling initialization; construction and
transport-reoptimized evaluation of the 11 candidate subspaces; precomputation of
reusable covariance quantities for the geometric step;
the safeguarded geometric step; gradient updates, transport recomputation, and
backtracking objective evaluations.
The runtime excludes the following: GMM fitting or empirical subsampling and assembly of the
optimizer's input arrays;
PMM fitting and prediction (and hence AUC/accuracy calculation); synthetic-data
generation; independent correctness checks; and file input/output.

Thus Figure~\ref{fig:time} reports optimizer time, not the time for an entire
classification pipeline. Representation-construction times are given separately
below. In contrast, Section~\ref{sec:representation_diagnostics_rev3} reports construction,
Screened GeoGrad optimization, and their sum, with a fixed all-pair choice rather than
confusing-pair selection. Both representations
use general transport LPs even at \(q=1\).

\begin{table}[t]
\centering\small
\setlength{\tabcolsep}{3pt}
\begin{tabular}{cR{6.8 (0.8)}R{7.5 (0.2)}R{10.2 (0.8)}R{154.9 (32.8)}R{497.1 (105.9)}R{2172.2 (252.8)}}
\toprule
\multirow{2}{*}{Dimension} & \multicolumn{3}{c}{GMM components \(K\)}
& \multicolumn{3}{c}{Empirical support points \(m\)}\\
 & \multicolumn{1}{c}{3} & \multicolumn{1}{c}{4} & \multicolumn{1}{c}{6}
 & \multicolumn{1}{c}{50} & \multicolumn{1}{c}{100} & \multicolumn{1}{c}{200}\\
\midrule
\csname @@input\endcsname revision_inputs/table_rows.tex
\bottomrule
\end{tabular}
\caption{Screened GeoGrad runtime in seconds per 45-subject training fit: mean (sample standard
deviation) over the first three datasets described above, at \(d=8\). The optimizer and iteration limit are shared across representations,
but support sizes are not calibrated to equal statistical accuracy.}
\label{tab:runtime}
\end{table}

Representation construction was measured separately. Fitting all 50 subject GMMs in a dataset took, on average, \(10.6\), \(14.7\), \(24.0\) seconds for \(K=3,4,6\), respectively. The corresponding sample standard deviations were \(1.1\), \(0.3\), \(0.6\) seconds. Forming the 50 empirical representations by subsampling took \(0.005\)--\(0.013\) seconds on average across support sizes.

\section{Gaussian-Cost and Transport-Solver Timing}
\label{ap:gaussian_kernel_timing}
The computational question changes in high dimension. A GMM has few transport
supports, but each support contains a covariance matrix; calculating Gaussian
Wasserstein costs can therefore be expensive even when the component-transport
LP is small. We isolate this effect by holding the number of subjects and
components fixed while increasing the ambient dimension.

We use four subjects with three dense-full-covariance components each, all six
subject pairs, and \(d\in\{8,16,32,64,128,256,512\}\), with five seeds.
For each generated dataset we separately time construction of all six Gaussian
component-cost matrices and solution of the six corresponding transport LPs.
We then project onto the first three coordinates, \(A=[e_1,e_2,e_3]\), and
repeat these calculations. The projected dimension is always \(q=3\), so the
projected cost calculations remain small as \(d\) increases.
This is a computational diagnostic at a prescribed projection, not an optimizer
comparison. The current Gaussian-cost and transport routines are used, but
GMM fitting, initialization screening, geometric proposals, gradient updates,
classification, validation, and file operations are not included. These are
standalone routine timings, including their covariance factorizations, not a
profile of the cache-aware stages of a complete Screened GeoGrad run.

\begin{table}[t]
\centering
\small
\setlength{\tabcolsep}{4pt}
\begin{tabular}{R{512}R{0.00193}R{0.00432}R{0.000417}R{0.000911}R{0.00327}}
\toprule
\multicolumn{1}{c}{$d$} & \multicolumn{1}{c}{\makecell{Ambient\\Gaussian costs}} & \multicolumn{1}{c}{\makecell{Ambient\\transport LPs}} & \multicolumn{1}{c}{Projection} & \multicolumn{1}{c}{\makecell{Projected\\Gaussian costs}} & \multicolumn{1}{c}{\makecell{Projected\\transport LPs}}\\
\midrule
8 & 0.00193 & 0.00432 & 0.000417 & 0.000911 & 0.00327\\
16 & 0.00342 & 0.00327 & 0.000384 & 0.000785 & 0.00263\\
32 & 0.00981 & 0.00273 & 0.000137 & 0.000575 & 0.00221\\
64 & 0.0389 & 0.00282 & 0.000194 & 0.000531 & 0.00217\\
128 & 0.157 & 0.00390 & 0.000385 & 0.000559 & 0.00239\\
256 & 1.06 & 0.00476 & 0.00113 & 0.000588 & 0.00258\\
512 & 6.81 & 0.00531 & 0.00428 & 0.000637 & 0.00258\\
\bottomrule
\end{tabular}
\caption{High-dimensional Gaussian and transport routines at $n=4$, $K=3$, and $q=3$. Entries are median seconds over five seeds, with all six subject pairs. These are separate routine measurements: even their sum is not a complete Screened GeoGrad runtime. }
\label{tab:high_dimension_kernels}
\end{table}

Table~\ref{tab:high_dimension_kernels} shows a change in the dominant operation.
At $d=8$, median ambient Gaussian-cost construction took
0.00193 seconds, less than the 0.00432 seconds for its transport LPs.
At $d=128$, the corresponding times were 0.157 and 0.00390 seconds.
At $d=512$, Gaussian-cost construction took 6.81 seconds, whereas the
original-dimensional LPs took 0.00531 seconds. After projection to $q=3$,
Gaussian costs and LPs took only 0.000637 and 0.00258 seconds, respectively.
Original-dimensional Gaussian-cost construction accounted for a median
99.8\% of the sum of the five measured stages at $d=512$.
This percentage shows that, with dense
high-dimensional covariances, Gaussian cost
construction can dominate even though the component-transport problems are small.

The implication is particularly relevant to high-dimensional applications:
the inequality \(K\ll m\) alone does not establish that a fitted GMM is cheaper
than an empirical representation. Covariance structure and dimension affect
cost construction, while mixture fitting adds another expense; caching and
specialized covariance routines can also change these costs.

\section{Additional Real-Data Sensitivity Analyses}
\label{ap:real_data_sensitivity}

\subsection{GMM Representation and Component Count}
\label{ap:component_sensitivity}

\begin{figure}[!htb]
	\centering
    \begin{subfigure}[b]{0.48\textwidth}
         \centering
         \includegraphics[width=0.95\textwidth]{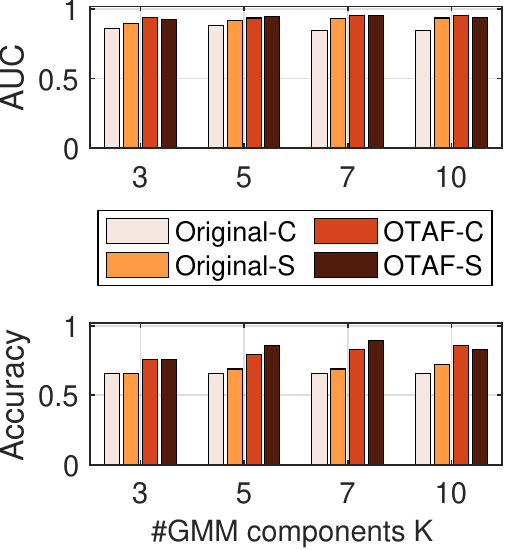}\\
         (a)
    \end{subfigure}
	\begin{subfigure}[b]{0.48\textwidth}
         \centering
         \includegraphics[width=0.95\textwidth]{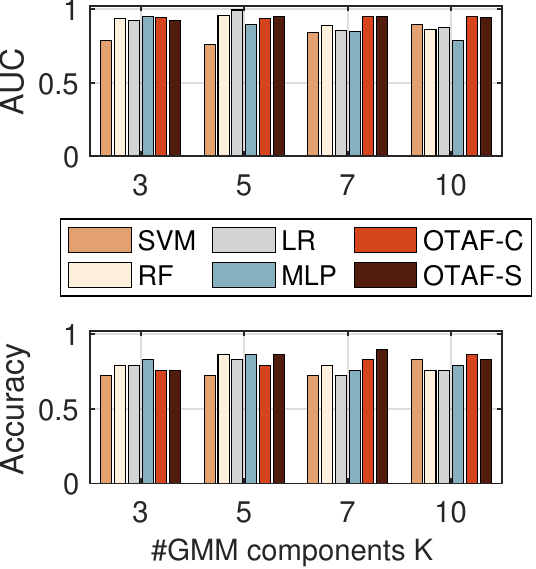}\\ (b)
    \end{subfigure}
\caption{PF component-count sensitivity using Screened GeoGrad. (a) PMM on
jointly clustered (C) and separately fitted (S) GMMs over
	\(K\in\{3,5,7,10\}\), in the original 30 dimensions and after OTAF
	reduction to one dimension. (b) Vector baselines using original-space
cluster-concatenated moments from GMM-C-\(K\), with SVM, RF, LR, and a
\(256\)--\(128\) MLP,
together with OTAF-C and OTAF-S.
OTAF-C and OTAF-S use training-fold confusing pairs and PMM after
one-dimensional reduction. }
	\label{fig:1}
\end{figure}

Figure~\ref{fig:1} examines sensitivity to component count on PF using Screened GeoGrad. OTAF-C and OTAF-S denote OTAF applied to GMM-C-$K$ and GMM-S-$K$, respectively. The retained dimension is one and the downstream classifier is PMM. Across the eight matched representations, AUC ranges from 0.926 to 0.953
after reduction, versus 0.847 to 0.937 in the original space. AUC and accuracy
increase in every comparison with matched component counts, but performance is not monotone in the component count. 

The vector classifiers in Figure~\ref{fig:1}(b) use the \emph{cluster-concatenated moments}:
component proportions, means, and marginal variances in a common aligned-cluster
order. Here these features are formed in the original 30-dimensional space from
GMM-C-\(K\), for each \(K\in\{3,5,7,10\}\); they are not computed from
GMM-S-\(K\) or from the one-dimensional OTAF projections. This comparison thus
allows vector classifiers to use the aligned cluster structure while the number
of components varies. 

\subsection{Reduced-Dimension Sensitivity with PMM}
\label{ap:dimension_sensitivity}
\begin{figure}[!htb]
  \centering
  \begin{subfigure}[t]{0.45\textwidth}
    \centering
    \includegraphics[width=\linewidth]{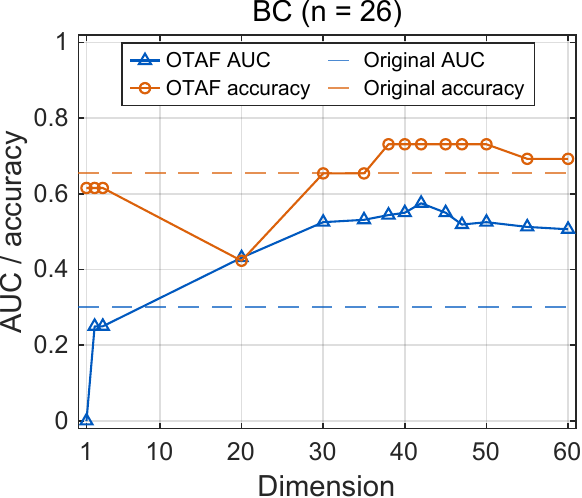}
    \subcaption{BC}\label{fig:3a}
  \end{subfigure}\hfill
  \begin{subfigure}[t]{0.45\textwidth}
    \centering
    \includegraphics[width=\linewidth]{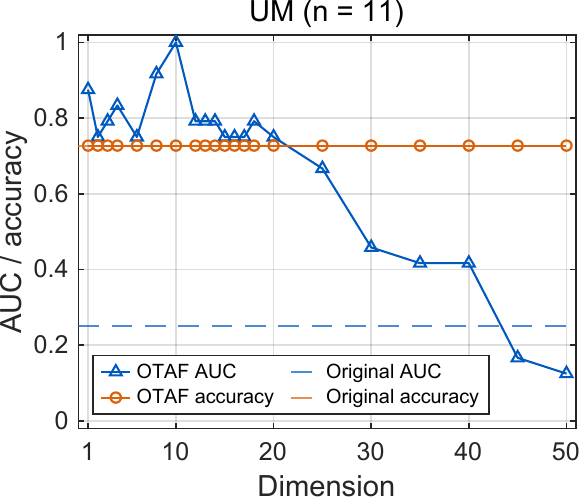}
    \subcaption{UM}\label{fig:3c}
  \end{subfigure}
  \caption{Screened GeoGrad dimension sensitivity for (a) BC and
  (b) UM, both using GMM-S-7 in ambient dimension 100, training-selected
  confusing pairs, and PMM. Solid curves show leave-one-subject-out AUC and
  accuracy after reduction; dashed lines show original-space PMM.
  For UM, the two accuracy curves coincide at $8/11$.
  BC uses $q\in\{1,2,3,20,30,35,38,40,42,45,47,50,55,60\}$;
  UM uses $q\in\{1,2,3,4,6,8,10,12,13,14,15,16,17,18,20,25,30,35,40,45,50\}$.}
  \label{fig:3}
\end{figure}

For BC and UM, Figure~\ref{fig:3} shows the dependence on reduced
dimension using Screened GeoGrad followed by PMM on GMM-S-7. The same original-space PMM result
provides a reference across dimensions.
Across this grid, BC AUC ranges from 0.000 to 0.575 and accuracy from 0.423 to 0.731;
UM AUC ranges from 0.125 to 1.000 and accuracy remains 0.727.
The PMM prior sensitivity examined in Appendix~\ref{ap:pmm_prior_sensitivity}
also affects these dimension curves;
for example, BC's dimension-one AUC changes from 0.000 to 0.350 when the
known training-prior odds are removed from the stored scores. Thus the very
low values should not be attributed to dimension reduction alone.
The BC point at dimension 40 (AUC 0.550, accuracy 0.731) and UM point
at dimension one (AUC 0.875, accuracy 0.727) agree with the OTAF+PMM
entries in Table~\ref{tab:current_classifier_sensitivity}.
These curves describe dimension sensitivity; selecting their best point
would not give an unbiased estimate of performance for a dimension-selection
procedure. In particular, UM has only 11 subjects, so its AUC can change
substantially when only a few held-out scores change order.
Section~\ref{sec:mean_dimension} and Figure~\ref{fig:mean_dimension_1nn}
provide a complementary dimension study with 1-NN after discarding component covariances.

\subsection{Sensitivity to Subject-Pair Selection}
\label{ap:pair_sensitivity}
The main comparisons use confusing-pair OTAF. Table~\ref{tab:pair_rule_sensitivity}
compares that rule with using all pairs, keeping Screened GeoGrad,
the supplied GMM-S-7 inputs, the retained dimension, and the downstream
classifier protocol unchanged between pair rules. Confusing pairs are selected within the
outer training fold; classifier tuning follows the conditional training-only
procedure in Section~\ref{sec:downstream}.
The two pair rules are fixed experimental configurations, not choices made
by selecting the best held-out result.

The effect depends on the cohort and classifier. On PF, confusing pairs give
higher AUC for all four classifiers; radial-classifier AUC changes from
0.858 with all pairs to 0.963 with confusing pairs. On BC, all pairs give
higher radial AUC (0.788 versus 0.650), whereas confusing pairs give higher
PMM AUC (0.550 versus 0.300). On UM, confusing-pair PMM gives AUC 0.875
and accuracy 0.727, compared with 0.583 and 0.636 for all pairs.
These small-cohort results do not establish a uniformly preferable pair rule.

\begin{table}[!htbp]
\centering\small
\setlength{\tabcolsep}{5pt}
\begin{tabular}{llR{.000}R{.000}R{.000}R{.000}}
\toprule
& & \multicolumn{2}{c}{AUC} & \multicolumn{2}{c}{Accuracy}\\
\cmidrule(lr){3-4}\cmidrule(lr){5-6}
Data & Classifier & \multicolumn{1}{c}{Confusing pairs} & \multicolumn{1}{c}{All pairs} & \multicolumn{1}{c}{Confusing pairs} & \multicolumn{1}{c}{All pairs}\\
\midrule
PF & 1-NN & .963 & .868 & .793 & .862\\
PF & Tuned $k$-NN & .939 & .884 & .862 & .862\\
PF & Radial & .963 & .858 & .931 & .862\\
PF & PMM & .953 & .863 & .897 & .828\\
\midrule
BC & 1-NN & .700 & .644 & .692 & .692\\
BC & Tuned $k$-NN & .575 & .675 & .577 & .615\\
BC & Radial & .650 & .788 & .577 & .692\\
BC & PMM & .550 & .300 & .731 & .615\\
\midrule
UM & 1-NN & .625 & .583 & .818 & .727\\
UM & Tuned $k$-NN & .667 & .604 & .818 & .727\\
UM & Radial & .333 & .333 & .818 & .727\\
UM & PMM & .875 & .583 & .727 & .636\\
\bottomrule
\end{tabular}
\caption{Subject-pair selection sensitivity of Screened GeoGrad on GMM-S-7. Both pair rules use the same fixed dimensions ($q=1$ for PF and UM, $q=40$ for BC) and leave-one-subject-out classifier protocols. Confusing pairs are selected using only the outer-training subjects. Neither pair rule is chosen using held-out outcomes.}
\label{tab:pair_rule_sensitivity}
\end{table}

\subsection{Training--Test Representation Mismatch on PF}
\label{ap:representation_mismatch}
\begin{figure}[!htb]
	\centering
    \begin{subfigure}[b]{0.48\textwidth}
         \centering
         \includegraphics[width=0.95\textwidth]{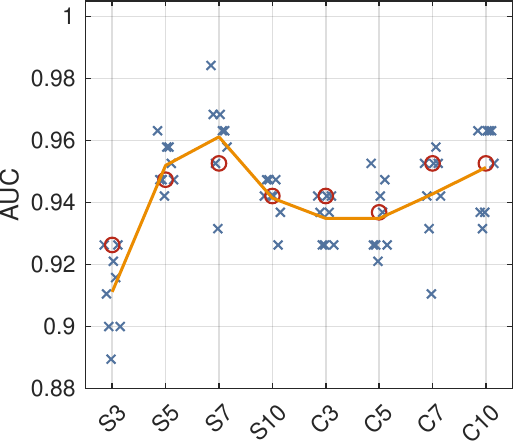}\\
         (a)
    \end{subfigure}
	\begin{subfigure}[b]{0.48\textwidth}
         \centering
         \includegraphics[width=0.95\textwidth]{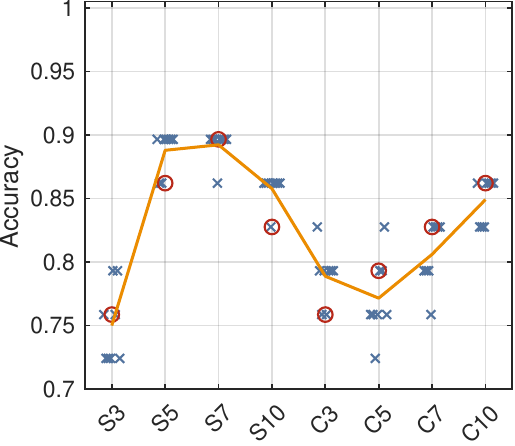}\\ (b)
    \end{subfigure}
	\caption{Screened GeoGrad on PF under representation-mismatch
	sensitivity at \(q=1\): (a) AUC and (b) accuracy. Each of the eight
	training representations (GMM-S-\(K\) followed by
	GMM-C-\(K\), \(K\in\{3,5,7,10\}\)) supplies a
	fold-contained OTAF map and PMM fit. The held-out subject is then
	represented under each of the eight test setups. Blue-gray crosses show all
	test setups, red circles identify the matched training--test setup, and
	the orange line is the mean across test setups for each training
	representation. The OTAF map and PMM fit are fold-contained conditional on the
	fixed supplied GMM representations; GMM construction itself is not repeated
	inside the folds.}
	\label{fig:2}
\end{figure}

Figure~\ref{fig:2} examines sensitivity to representing a held-out subject
differently from the training subjects. It reuses the 232 Screened GeoGrad projections fitted for the component-count study in Figure~\ref{fig:1}.
Each of eight training representations (GMM-S-\(K\) and GMM-C-\(K\),
\(K\in\{3,5,7,10\}\)) supplies 29 leave-one-subject-out projections at
\(q=1\) and their PMM fits. Each fitted map and classifier is then reused
with each of the eight representations of the held-out subject, giving
64 training--test configurations.

Mean AUC is 0.944 for the eight matched setups and 0.941 for the 56
mismatched setups; individual AUCs across all 64 setups range from 0.889
to 0.984. The corresponding mean accuracies are 0.823 and 0.826.
Transport permits unequal component counts without requiring aligned component
labels. Nevertheless, this sensitivity study neither selects an optimal
component count nor establishes uniform insensitivity to representation choice.
The analysis is conditional on the supplied GMMs: GMM construction itself
is not repeated inside the folds.

\subsection{Sensitivity of Leave-One-Subject-Out PMM AUC to Class Priors}
\label{ap:pmm_prior_sensitivity}
The PMM scores used in the real-data comparisons include the class proportions
in each training fold. These proportions differ systematically across
leave-one-subject-out folds: removing a positive subject lowers the positive
class prior, whereas removing a negative subject raises it. If the two
class pseudo-densities were equal for every held-out subject, the resulting
pooled scores would rank every positive below every negative, giving AUC zero
despite the absence of discriminatory information. Thus a pooled
leave-one-subject-out AUC below one half need not imply that the distance or projection
reverses the biological ordering.

Let
\(s_k\) be the stored positive-class PMM posterior score and
\(\widehat\pi_{1,-k}\) the positive-class proportion in its training fold.
Then the posterior score under equal priors is
\(\displaystyle
u_k=\frac{s_k/\widehat\pi_{1,-k}}
{s_k/\widehat\pi_{1,-k}+(1-s_k)/(1-\widehat\pi_{1,-k})} .
\) 
We can then assess the prior's effect without refitting any projection or classifier. 
This is the normalized ratio of the two fitted class pseudo-densities; it
changes only the scoring convention, not the learned representations,
distances, or fitted PMM scale. Table~\ref{tab:pmm_prior_sensitivity} compares
the corresponding AUCs on the same held-out subjects.
The below-chance BC and UM baselines are strongly affected by the prior
factor. 

For QOT-Cosine, removing the PMM prior factor changes PF, BC, and UM AUCs
from 0.937, 0.538, and 1.000
to 0.937, 0.675, and 1.000, respectively.

\begin{table}[htbp]
\centering\small
\setlength{\tabcolsep}{5pt}
\begin{tabular}{lR{.000}R{.000}R{.000}R{.000}}
\toprule
Data & \multicolumn{1}{c}{Original} & \multicolumn{1}{c}{PCA} & \multicolumn{1}{c}{LDA} & \multicolumn{1}{c}{Screened GeoGrad}\\
\midrule
PF, reported & .932 & .926 & .647 & .953\\
PF, prior removed & .942 & .932 & .653 & .953\\
\midrule
BC, reported & .300 & .300 & .300 & .550\\
BC, prior removed & .506 & .531 & .538 & .700\\
\midrule
UM, reported & .250 & .125 & .708 & .875\\
UM, prior removed & .750 & .958 & .792 & .917\\
\bottomrule
\end{tabular}
\caption{PMM AUC sensitivity. All methods use
GMM-S-7; projected methods use $q=1,40,1$ for PF, BC, and UM, respectively. ``Reported'' denotes the training-prior
scores used in Table~\ref{tab:current_classifier_sensitivity}. ``Prior removed'' uses $u_k$ above.}
\label{tab:pmm_prior_sensitivity}
\end{table}

\bibliography{revision_inputs/reference_rev103}

\end{document}